\documentclass{article}

\usepackage[preprint]{neurips_data_2024}

\usepackage[utf8]{inputenc} % allow utf-8 input
\usepackage[T1]{fontenc}    % use 8-bit T1 fonts
\usepackage[pagebackref,breaklinks,colorlinks,bookmarks=False]{hyperref}
\usepackage{url}            % simple URL typesetting
\usepackage{booktabs}       % professional-quality tables
\usepackage{amsfonts}       % blackboard math symbols
\usepackage{nicefrac}       % compact symbols for 1/2, etc.
\usepackage{microtype}      % microtypography
\usepackage{xcolor}         % colors

\usepackage[labelformat=simple]{subcaption}

\usepackage{wrapfig,lipsum,booktabs}
\usepackage[inline]{enumitem}
\usepackage{caption}
\usepackage{amsmath}
\usepackage{graphicx}
\usepackage{placeins}
\usepackage{multicol}
\usepackage{multirow}
\usepackage{siunitx}
\usepackage[linesnumbered,ruled,vlined]{algorithm2e}
\usepackage{algorithmic}
\usepackage{diagbox}
\usepackage{rotating}
\usepackage{tcolorbox}
\usepackage{xspace}
\usepackage{fontawesome5}

\makeatletter
\newcommand{\settitle}{\@maketitle}
\makeatother

\newcommand{\spc}{\!\ \!\ }

\definecolor{Green}{RGB}{18, 159, 87}
\definecolor{dark-blue}{RGB}{0,101,240}
\definecolor{Blue}{RGB}{30,118,181}
\definecolor{light-blue}{RGB}{0,170,255}
\definecolor{grass}{RGB}{93,161,48}

\hypersetup{
    urlcolor=dark-blue,
    citecolor=light-blue,
    linkcolor=dark-blue
}
\setcitestyle{numbers,square,comma}

\newcommand{\Q}[1]{\noindent\textbf{Q}\textbf{{{#1} }\\}}
\newcommand{\A}[1]{\noindent-- {#1}}

\def\TITLEVAL{RClicks: Realistic Click Simulation\\for Benchmarking Interactive Segmentation}

\title{\TITLEVAL}

\newcommand{\weburl}{\url{https://github.com/emb-ai/rclicks}\xspace}
\author{%
\textbf{Anton Antonov}\textsuperscript{$1*$\spc \faEnvelope[regular]} \quad
\textbf{Andrey Moskalenko}\textsuperscript{$1,2*$} \quad
\textbf{Denis Shepelev}\textsuperscript{$1*$} \\ 
\textbf{Alexander Krapukhin}\textsuperscript{$1$} \quad
\textbf{Konstantin Soshin}\textsuperscript{$1$} \quad
\textbf{Anton Konushin}\textsuperscript{$1,2$} \quad
\textbf{Vlad Shakhuro}\textsuperscript{$1,2$$\dagger$} \\
\textsuperscript{$1$}AIRI, Moscow, Russia\quad
\textsuperscript{$2$}Lomonosov Moscow State University\\
 \texttt{\{lastname\}@airi.net}\\
 \textsuperscript{$*$}Equal contribution\quad
 \textsuperscript{$\dagger$}Project leader\quad
 \textsuperscript{\faEnvelope[regular]}Corresponding author\\
 \faGithub\;\weburl
}

\begin{document}
\setcitestyle{numbers}

\maketitle

\begin{abstract}
    The emergence of Segment Anything (SAM) sparked research interest in the field of interactive segmentation, especially in the context of image editing tasks and speeding up data annotation. Unlike common semantic segmentation, interactive segmentation methods allow users to directly influence their output through prompts (e.g. clicks). However, click patterns in real-world interactive segmentation scenarios remain largely unexplored. Most methods rely on the assumption that users would click in the center of the largest erroneous area. Nevertheless, recent studies show that this is not always the case. Thus, methods may have poor performance in real-world deployment despite high metrics in a baseline benchmark. To accurately simulate real-user clicks, we conducted a large crowdsourcing study of click patterns in an interactive segmentation scenario and collected 475K real-user clicks. Drawing on ideas from saliency tasks, we develop a clickability model that enables sampling clicks, which closely resemble actual user inputs. Using our model and dataset, we propose RClicks benchmark for a comprehensive comparison of existing interactive segmentation methods on realistic clicks. Specifically, we evaluate not only the average quality of methods, but also the robustness w.r.t. click patterns. According to our benchmark, in real-world usage interactive segmentation models may perform worse than it has been reported in the baseline benchmark, and most of the methods are not robust. We believe that RClicks is a significant step towards creating interactive segmentation methods that provide the best user experience in real-world cases.
\end{abstract}

\section{Introduction}
\label{intro_ref}

The task of interactive segmentation involves providing additional hints or prompts to the method, allowing it to produce more precise annotations compared to conventional semantic segmentation.
The most famous member of interactive segmentation methods is Segment Anything (SAM) \citep{kirillov2023segment, ravi2024sam}.
Nowadays, SAM-like methods are applied in various fields, including the thin object segmentation~\citep{ke2023segment, xie2024pasam}, medical segmentation~\citep{hu2023sam, zhou2023sam, biswas2023polypsam, zhang2023segment, SAM4MIS, mazurowski2023segment}, 3D segmentation~\citep{yang2023sam3d, xu2023sampro3d}, tracking~\citep{cheng2023segment} and video~\citep{ravi2024sam}.

Typically, interactions occur in several rounds, where in each round the user corrects the prediction errors of the previous one.
Evaluation of such methods requires user inputs.
However, collecting many real-user inputs for multiple rounds is impractical since such a dataset needs to be rebuilt for every method and every interaction round due to its iterative nature.
Thus, researchers often resort to a simple strategy %heuristics
to simulate user inputs.
According to this strategy, a single click for each interaction round is generated as follows: 
\begin{enumerate*}[label=(\arabic*)]
\item select the largest error region in the previous interaction round, and
\item click in the furthest point from the boundaries of this region (center point).    
\end{enumerate*}
Hereinafter, we refer to the above click sampling strategy as a \textit{baseline strategy}, following~\citep{moskalenko2024tetris}.
However, relying solely on this approach may result in overfitting and degraded performance in real-world usage scenarios.

\begin{figure}
    \centering
    \begin{subfigure}[t]{0.21\textwidth}
        \centering
        \includegraphics[height=0.23\textheight]{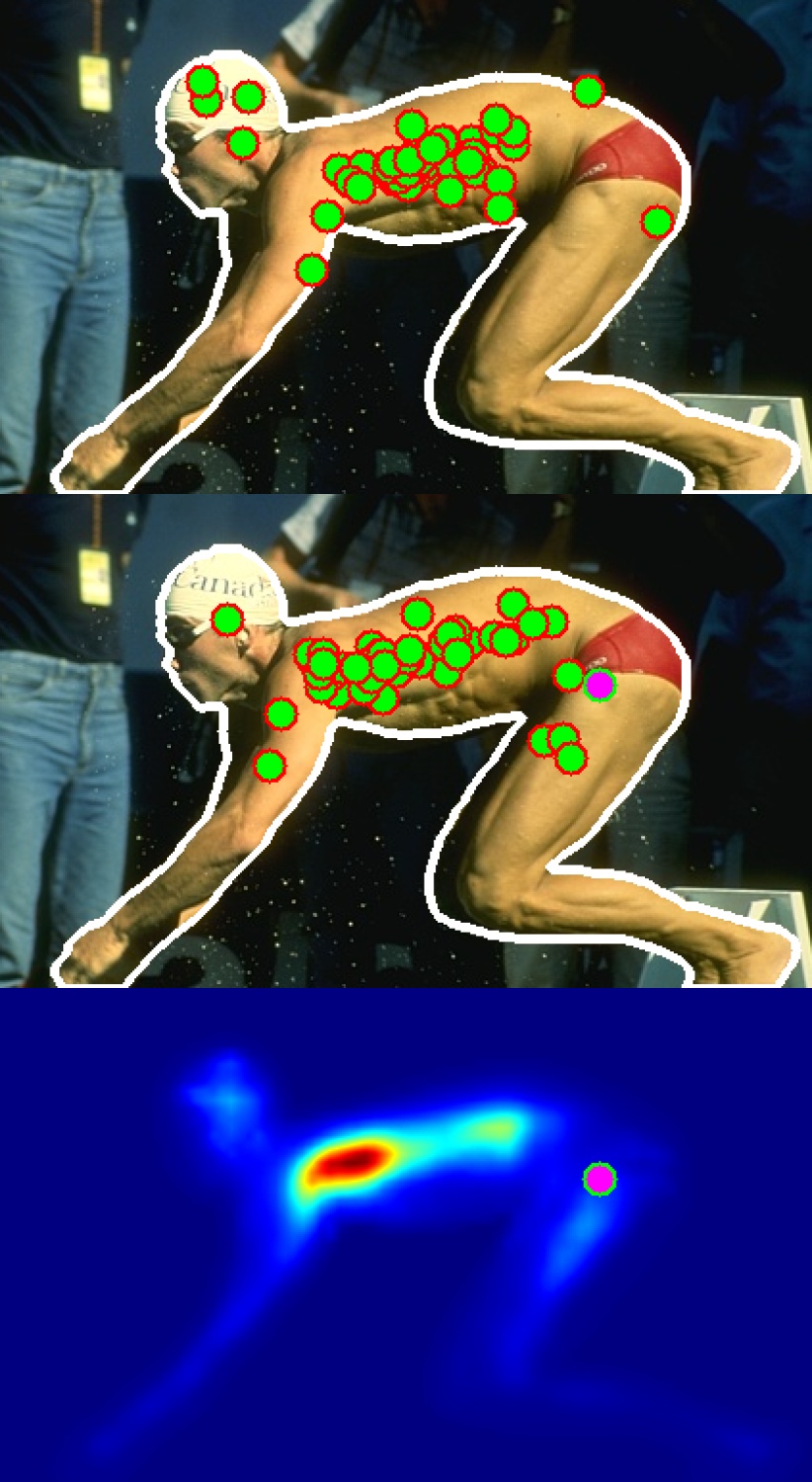}
        \caption{GrabCut~\cite{grabcut}}
        \label{fig:teaser:grabcut}
    \end{subfigure}
    \hspace*{\fill}
    \begin{subfigure}[t]{0.18\textwidth}
        \centering
        \includegraphics[height=0.23\textheight]{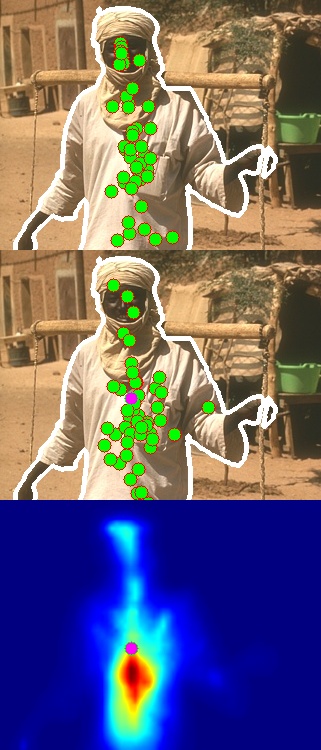}
        \caption{Berkeley~\cite{berkeley-intro}}
        \label{fig:teaser:berkeley}
    \end{subfigure}
    \hspace*{\fill}
    \begin{subfigure}[t]{0.21\textwidth}
        \centering
        \includegraphics[height=0.23\textheight]{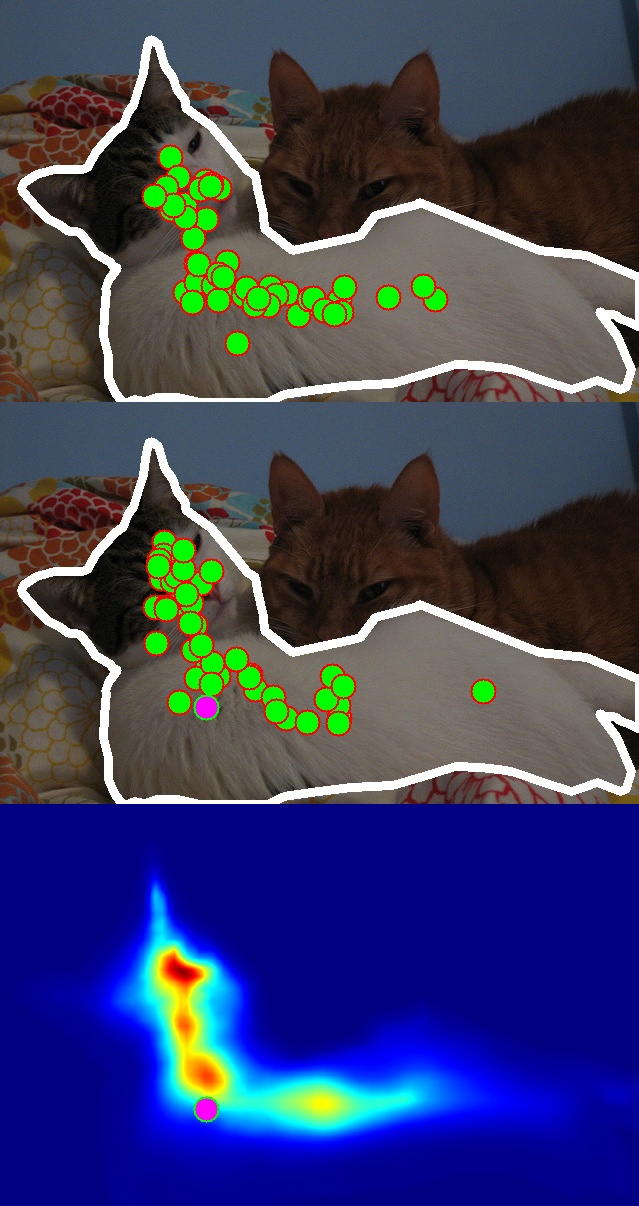}
        \caption{COCO-MVal~\cite{coco}}
        \label{fig:teaser:coco}
    \end{subfigure}
    \hspace*{\fill}
    \begin{subfigure}[t]{0.18\textwidth}
        \centering
        \includegraphics[height=0.23\textheight]{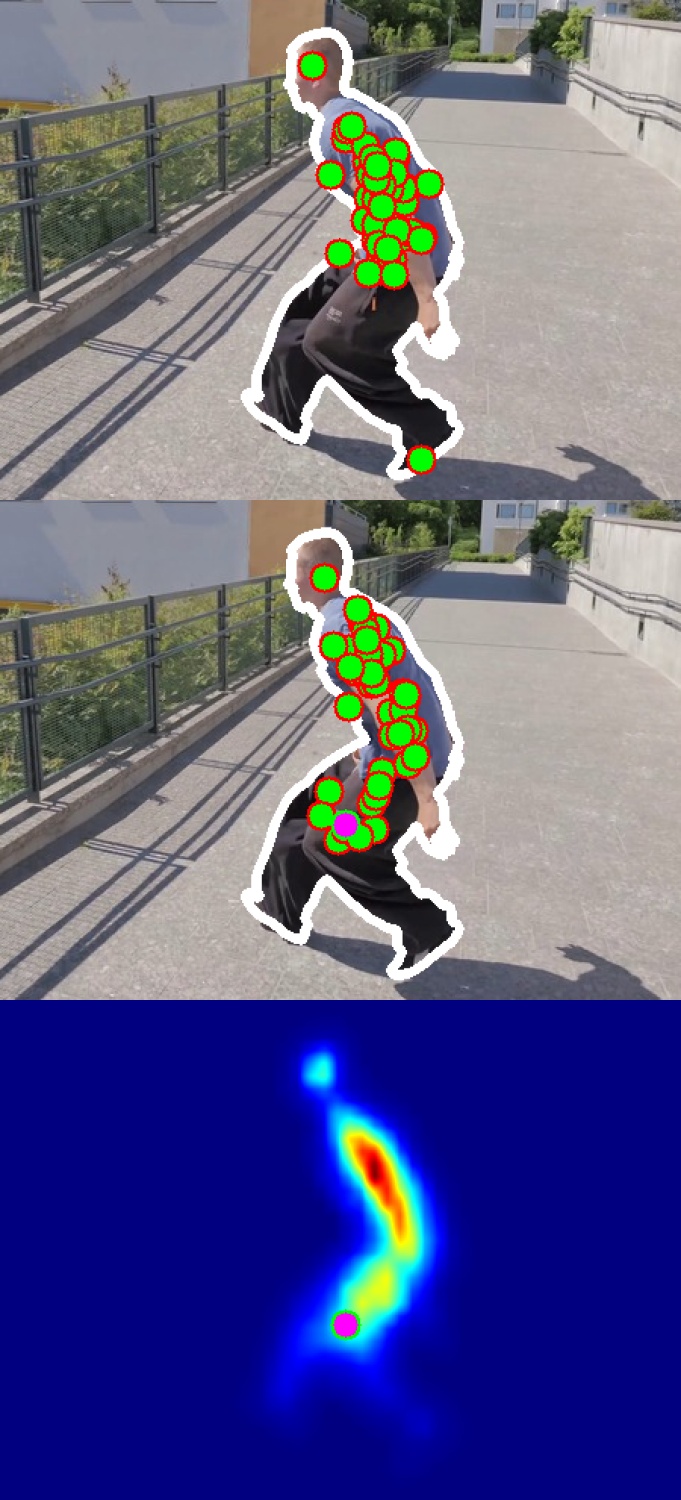}
        \caption{DAVIS~\cite{davis}}
        \label{fig:teaser:davis}
    \end{subfigure}
    \hspace*{\fill}
    \begin{subfigure}[t]{0.17\textwidth}
        \centering
        \includegraphics[height=0.23\textheight]{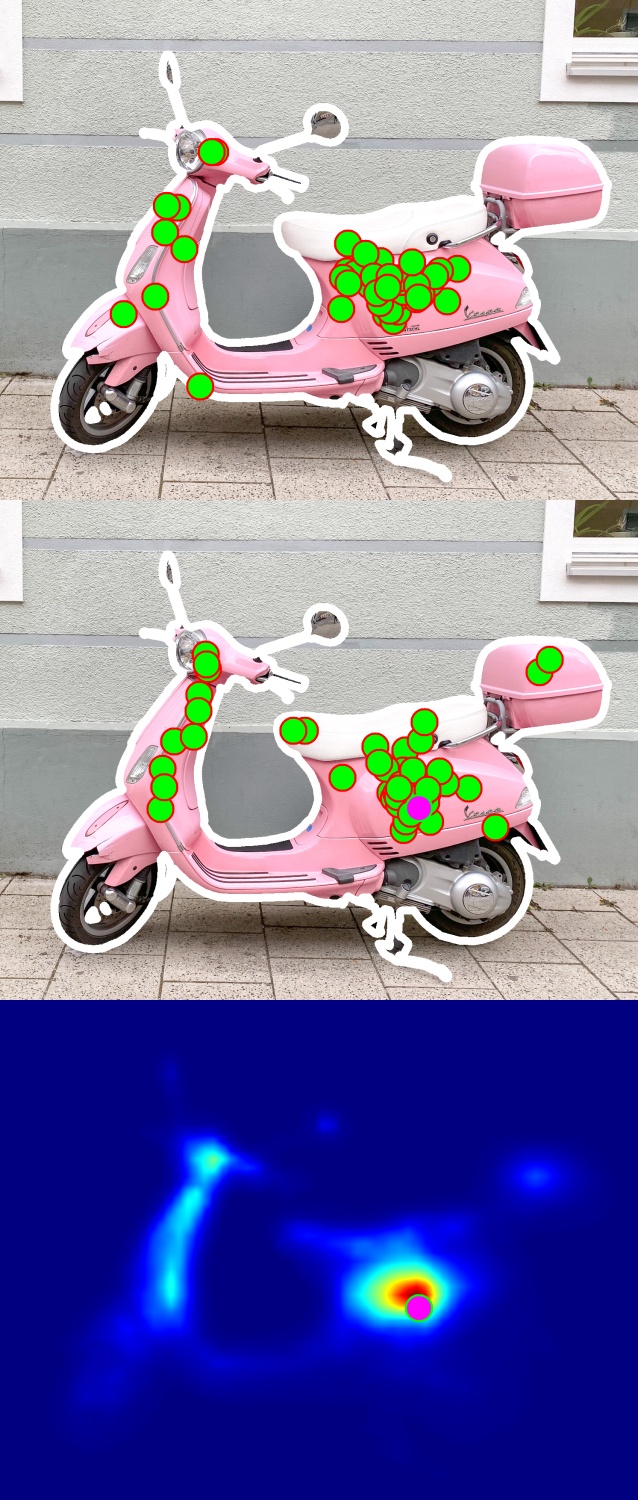}
        \caption{TETRIS~\cite{moskalenko2024tetris}}
        \label{fig:teaser:tetris}
    \end{subfigure}
    \hspace*{\fill}
    \caption{Examples of real and predicted users' clicks of interactive segmentation task. 
    The upper row depicts real-users clicks (green) for a given target object (white contour); the middle and bottom rows visualize, correspondingly, clicks and their distribution predicted by our clickability model.
    Purple points in the middle and bottom rows represent clicks generated by the \textit{baseline strategy}~\cite{ritm}.
    Mostly baseline click is close to a mode of users' distribution (see \subref{fig:teaser:berkeley} and \subref{fig:teaser:tetris}), however, in some cases it may be far from the mode (e.g.~\subref{fig:teaser:grabcut}, \subref{fig:teaser:davis}) or may not represent all modes of the distribution (e.g.~\subref{fig:teaser:coco}, \subref{fig:teaser:tetris}).
    }
    \label{fig:teaser}
\vspace{-1em}
\end{figure}

Our goal is to create a highly realistic simulator of user clicks to enable a more accurate evaluation of interactive segmentation methods.
We begin our research with a simple observation: when a user clicks, their gaze is focused on the area where they click. 
In turn, the task of predicting saliency is well-studied, with benchmark data collected using specialized devices such as eye trackers.
Saliency prediction models generate spatial attention heatmaps, from which fixation points of viewers can be sampled and utilized.
However, saliency models assume a free-viewing task, which differs from interactive segmentation, where the user should segment a specific area.
In other words, clicks should be sampled from a spatial distribution, that is conditioned not only on an image but also on a target segmentation area.

Drawing from best practices in interactive segmentation and saliency prediction tasks, we collect a dataset of task-specific user clicks, and propose a model that facilitates sampling of the most realistic click positions in interactive segmentation.
Overall, our main contributions are as follows: \begin{itemize}
\item We curate a large multiple-round interaction dataset in the interactive segmentation task (see samples from the first round in Figure~\ref{fig:teaser}).
To achieve this, we introduce a click collection methodology and conduct an ablation to address presentation bias, involving users on both PCs and mobile devices.
\item We introduce a novel click sampling strategy based on a \textit{clickability model} that can sample more realistic clicks than the \textit{baseline strategy} and estimate click probabilities.
\item We present RClicks~-- a novel benchmark, that leverages the \textit{clickability model} to estimate the real-world performance of interactive methods.
We conduct extensive comparisons and benchmark state-of-the-art methods using both the \textit{baseline strategy} clicks, and realistic clicks simulated by the \textit{clickability model}.
This comparison reveals that benchmarks employing the \textit{baseline strategy} may overestimate methods' real-world performance.
Moreover, we conclude that current segmentation methods are unable to achieve both optimal performance and robustness simultaneously on all datasets.
\item We utilize the collected first-round real-user clicks to evaluate the performance of segmentation methods.
Furthermore, we propose a methodology to estimate the real-world segmentation difficulty for state-of-the-art methods for each instance in a dataset.

\end{itemize}

We believe that the proposed methodology enhances 
comprehension of real-users actions and will facilitate the development of interactive methods that are more applicable in real-world cases.

\section{Related Work}
\label{gen_inst}

\subsection{User Input Types in Interactive Segmentation}

Various types of user inputs have been explored in the literature.
In \citep{grabcut, ferrari2019scribbles} an initial selection is obtained using bounding boxes, and then refined with strokes.
In \citep{gueziri2017latency} object selection is done with strokes.
\citep{popenova2023contour} considers contours for selecting small objects, minor parts of an object, or a group of objects of the same type.
\citep{cheng2021flexible} proposes to use trimap, scribblemap or clickmap as an input.
Segment Anything, or SAM~\citep{kirillov2023segment}, processes multiple types of user prompts, including a point, a box, a mask, or a text.

Clicks-based approach selects objects of interest according to multiple user clicks (either positive or negative), and was first introduced in~\citep{deep-object-selection} and investigated in~\citep{ritm, liu2022simpleclick, sun2023cfricl, zhou2023interactivegpcis, kirillov2023segment, ke2023segment, zhang2023faster}.
We focus solely on the click-based approach, since it is well-explored and has an established evaluation procedure in the field.

\subsection{Benchmarking Interactive Segmentation}

GrabCut~\citep{grabcut} is the first dataset proposed for interactive segmentation task.
Then \citep{berkeley-intro} adapted Berkeley~\citep{berkeley} segmentation dataset to evaluate interactive segmentation methods, but it required manual testing.
However, manual testing is a time-consuming and resource-intensive process.
Interactive segmentation expects multiple rounds of interactions, when each interaction depends on previous ones, and it is infeasible to apply manual procedure for larger scales.
For these reasons, in practice, benchmarks generate user interactions automatically based on previous interactions.

In \citep{deep-object-selection} authors proposed an automatic clicks generation strategy for evaluation on PASCAL~VOC~2012~\citep{pascal-voc-2012} and COCO~\citep{coco} segmentation datasets.
The subsequent work~\citep{latent-diversity} used DAVIS~\citep{davis} and SBD~\citep{sbd} datasets for interactive segmentation, applying the same \textit{baseline strategy}.

Most of the existing click-based methods~\citep{ritm, liu2022simpleclick, sun2023cfricl, zhou2023interactivegpcis, kirillov2023segment, ke2023segment, zhang2023faster} use the \textit{baseline strategy}.
However, it has not been validated in real-world usage scenarios until recently.
\citep{moskalenko2024tetris} introduced TETRIS benchmark and revealed that real users do not always click in the center of an area with the largest error, as assumed in the \textit{baseline strategy}.
Using the adversarial attacks, the paper demonstrated that methods have a tendency to overfit to the \textit{baseline strategy}.
Specifically, when the baseline clicks are used, the segmentation quality may be high, but even a slight change in the click position can result in a significant drop in quality.
Therefore, the \textit{baseline strategy} may not accurately estimate the quality of the methods in real usage.
We believe that to estimate the actual quality, each click should be generated in accordance with \textit{human perception}.

\subsection{Saliency Prediction}

The task of saliency prediction aims to model human perception by predicting probability maps~\citep{TranSalNet, kroner2020contextual} of user engagement in a free-view observation for a given media content.
Reference data for this task usually comes from a specialized device -- an eye tracker -- which records eye \textit{fixations}~\citep{judd2009learning, judd2012benchmark, CAT2000}.
Subsequently, fixations from multiple viewers are aggregated into a probability distribution through Gaussian at each fixation point, with sigma corresponding to the retinal angle of a human's field of view~\citep{judd2009learning}.
Since scaling expensive eye tracker experiments is too complex, several researchers~\citep{jiang2015salicon, tavakoli2017saliency} proposed to use mouse movements as a proxy for saliency when training saliency models.
However, saliency fixations cannot be directly used in the interactive segmentation task because saliency observers engage in free-viewing, while in our task, the user's goal is to make a click to highlight a specific object or a part of it.
Thus, for the interactive segmentation problem, real-user clicks should be collected.

\section{Users' Clicks Dataset}

We propose a novel dataset of real-users clicks for interactive segmentation.
Our dataset is based on the existing image segmentation datasets.
In total, we collected 475\,544 user inputs for GrabCut~\cite{grabcut}, Berkeley~\cite{berkeley}, DAVIS~\cite{davis}, COCO-MVal~\cite{lin2014microsoft}, TETRIS~\cite{moskalenko2024tetris}.
To gather users' clicks, we developed a specialized presentation tool.
Specifically, in each task, we asked users to click on the target objects by displaying images and corresponding segmentation masks.
We considered several display modes to instruct users what objects should be selected.
As interface can cause bias in clicks distribution, we conducted a user study to select the option that best mimics natural user object selections.

This section is organized as follows.
First, we present task display modes~\eqref{sec:preview}.
Then we choose one that eliminates bias associated with user viewing mode behavior~\eqref{sec:preview:ablation}.
Finally, we describe clicks collection in the first and the subsequent rounds of interactions~\eqref{sec:collected_clicks}.

\subsection{Collection Procedure}
\label{sec:preview}

\begin{wrapfigure}{r}{0.7\textwidth}
\vspace{-1.5em}
    \centering
    \includegraphics[width=0.7\textwidth]{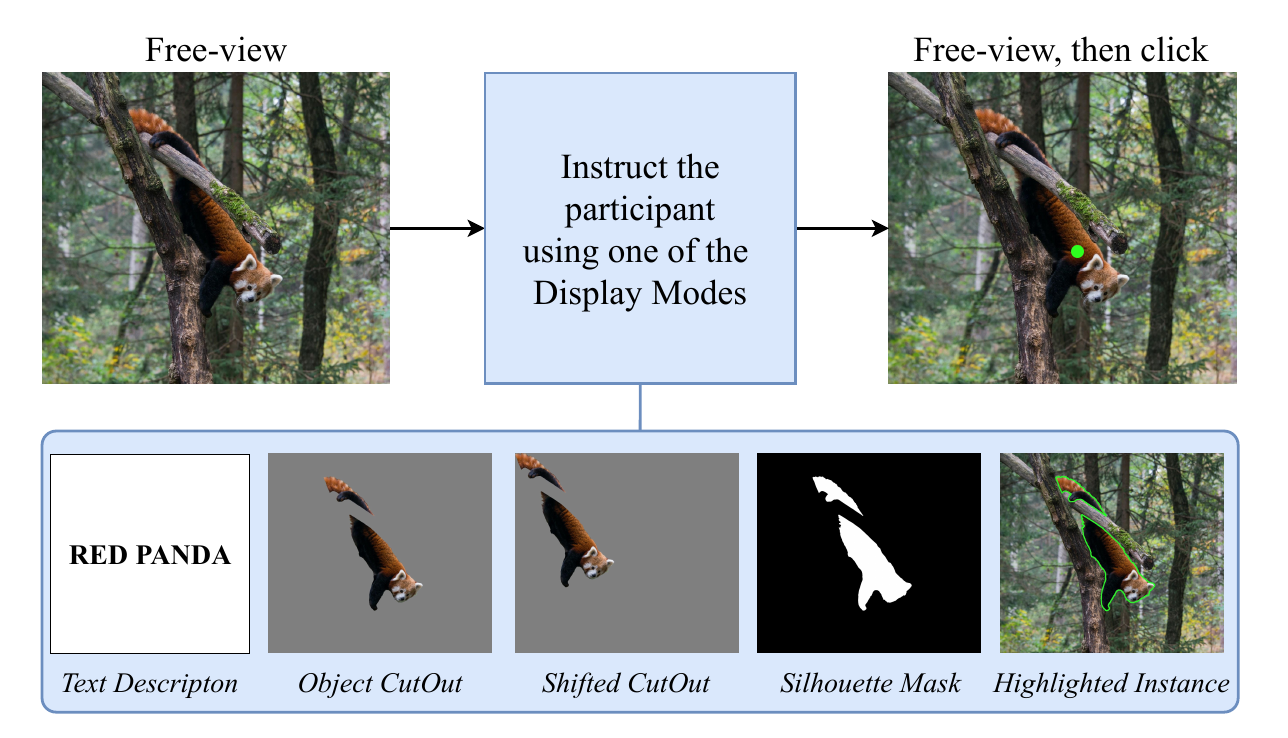}
    \caption{Illustration of the tested display modes to reduce presentation bias.
    The best result was obtained with the \textit{Object CutOut} mode, where an object is presented on a gray background without shifts.
    }
    \label{fig:presentation_strategies}
\vspace{-1em}
\end{wrapfigure}

When collecting user clicks, we executed the following procedure  (see Figure~\ref{fig:presentation_strategies}): \begin{enumerate*}[label=(\arabic*)]
\item \label{diplay:show_im1} Show the entire image for 1.5 seconds. \item \label{diplay:modes} Show segmentation target using one of the Display Modes. \item \label{diplay:show_im2} The entire image is shown again for 1.5 seconds, during which clicking is not allowed. \item The user makes a click.
\end{enumerate*}
Steps~\ref{diplay:show_im1} and~\ref{diplay:show_im2} simulate the user behavior during real-world interactive segmentation, when individuals initially view the image and then interact with it by clicking.
Step~\ref{diplay:modes} visualizes the object that should be selected by the user.

We considered the following task displaying modes.
\begin{enumerate*}[label=(\alph*)]
\item \textit{Text Description} mode shows the textual description of the target object for 2.5 seconds.
\item  \textit{Object CutOut} visualizes the target instance in its original position on a gray background for 2 seconds.
\item  During \textit{Shifted CutOut}, the target instance is shown for 2 seconds on a gray background, then shifted to the top-left corner, which aims to motivate the assessor to independently locate the instance on the image as its position is shifted.
\item  \textit{Silhouette Mask} shows a black-and-white mask of the target instance in the original position for 2 seconds.
\item  \textit{Highlighted Instance} displays the original image with the background where the target instance is highlighted with a green border.
Rationale for our choice of these displaying modes and time periods can be found in Appendix~\ref{sec:data_collection_procedure}.
\end{enumerate*}

\subsection{Selecting Unbiased Task Display Mode}
\label{sec:preview:ablation}

\begin{figure}[h]
    \centering
    \begin{subfigure}[t]{0.24\textwidth}
        \centering
        \includegraphics[width=0.658\textwidth,angle=90]{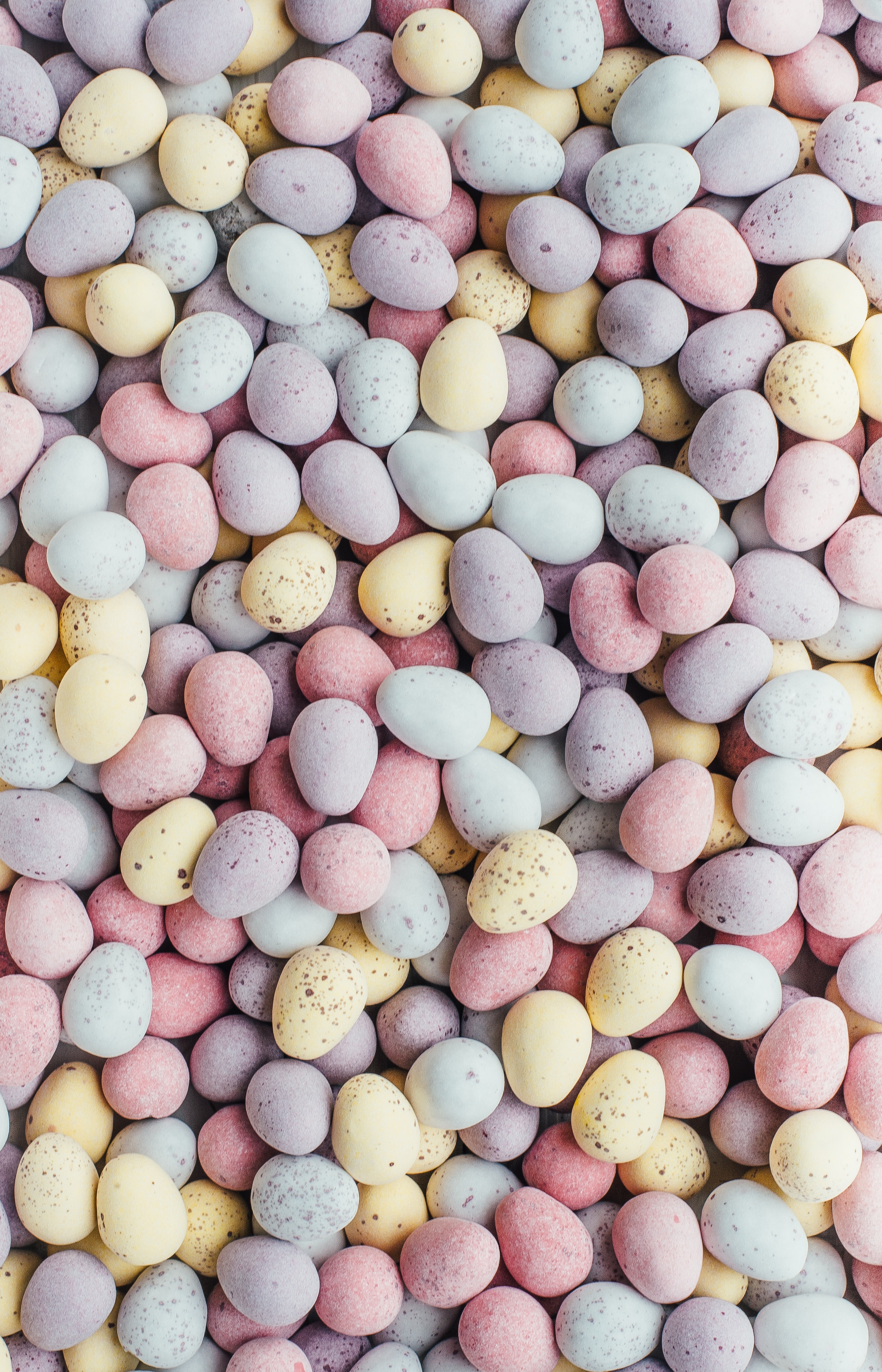}
        \caption{Original image}
        \label{fig:hard_clicks_descriptions:1}
    \end{subfigure}
    \hspace*{\fill}
    \begin{subfigure}[t]{0.24\textwidth}
        \centering
        \includegraphics[width=0.658\textwidth,angle=90]{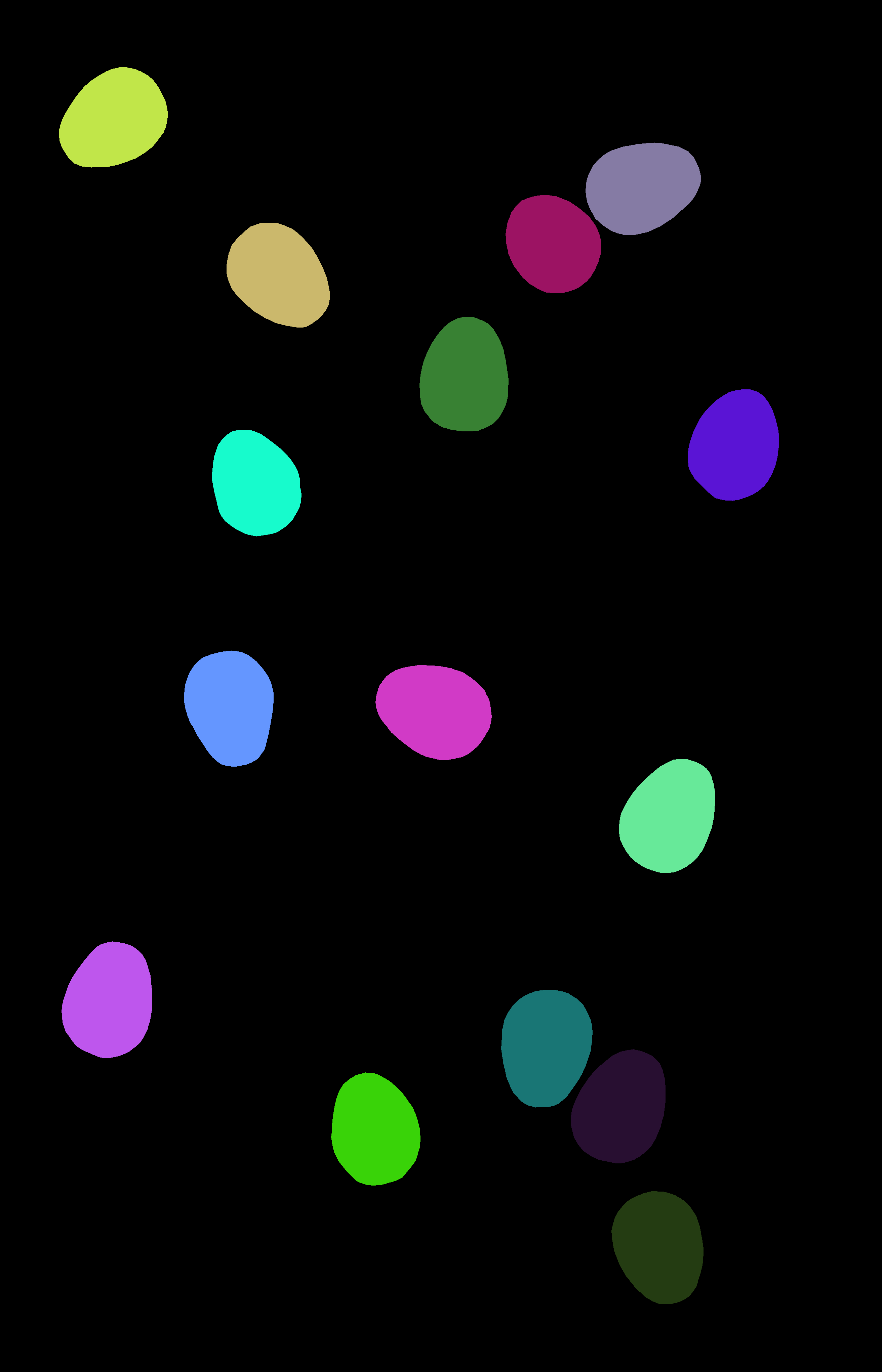}
        \caption{Instances masks}
        \label{fig:hard_clicks_descriptions:2}
    \end{subfigure}
    \hspace*{\fill}
    \begin{subfigure}[t]{0.24\textwidth}
        \centering
        \includegraphics[width=\textwidth]{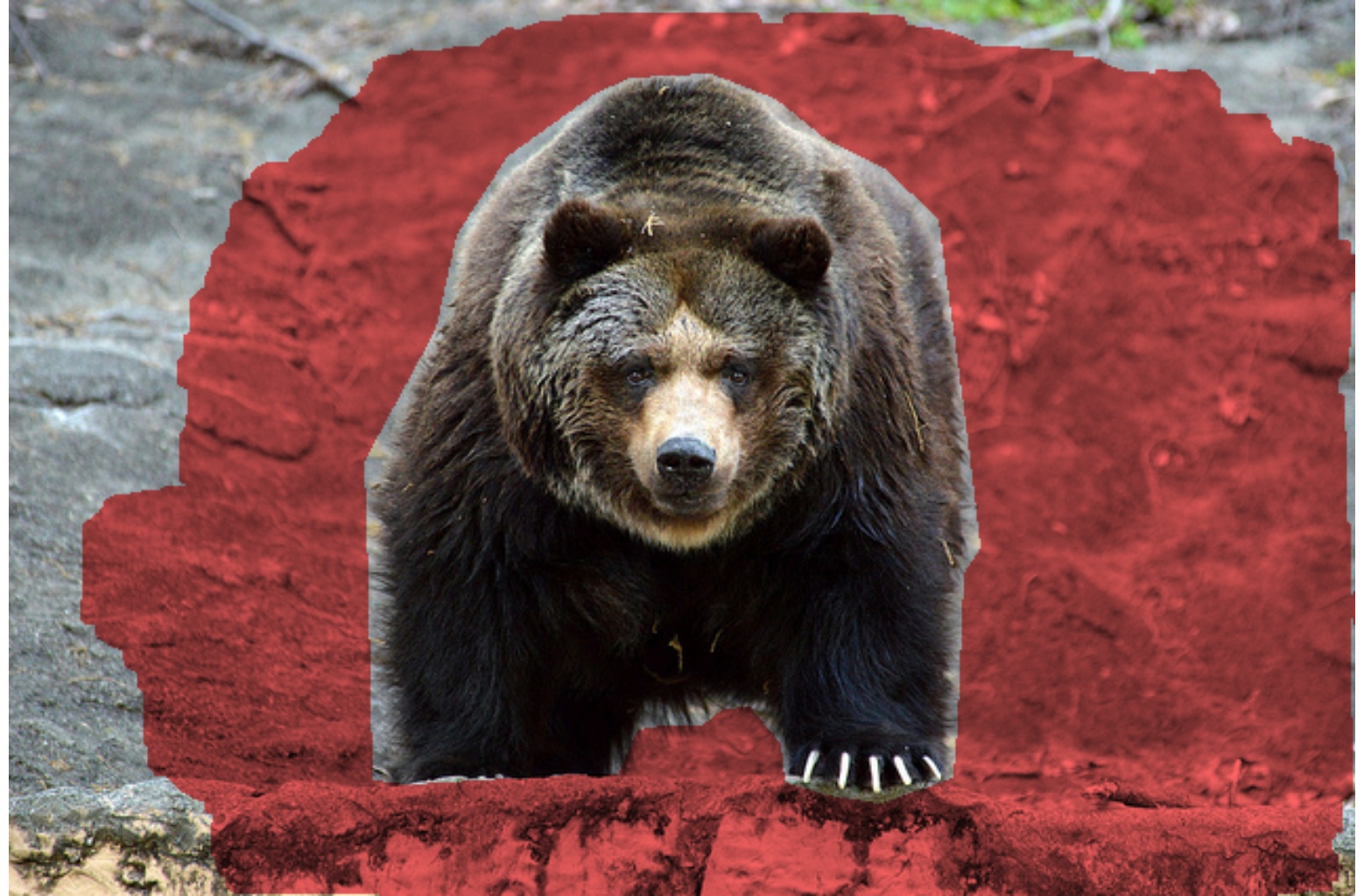}
        \caption{False-positive error (red mask)}
        \label{fig:hard_clicks_descriptions:fp}
    \end{subfigure}
    \hspace*{\fill}
    \begin{subfigure}[t]{0.24\textwidth}
        \centering
        \includegraphics[width=\textwidth]{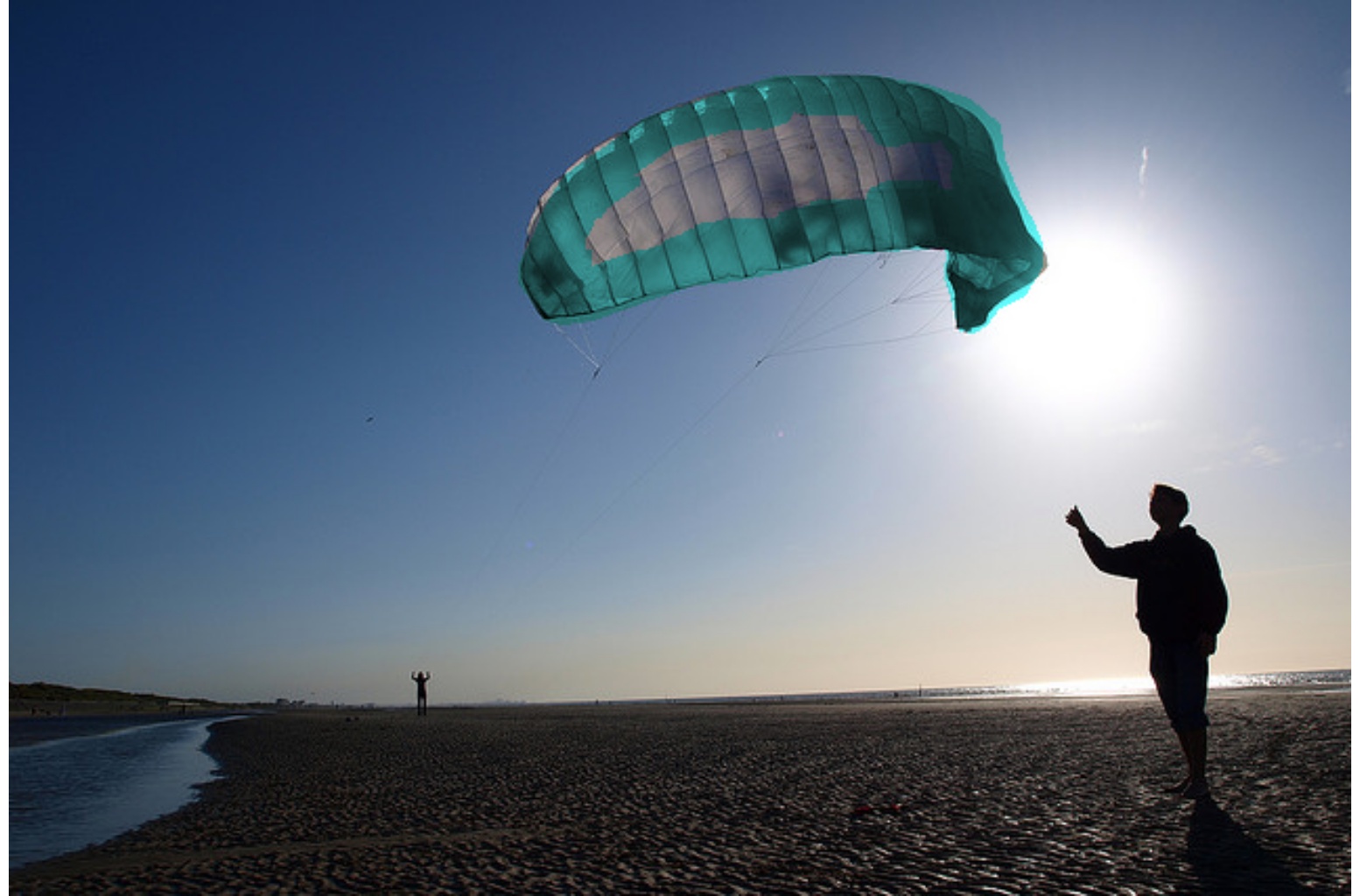}
        \caption{False-negative error (teal mask)}
        \label{fig:hard_clicks_descriptions:fn}
    \end{subfigure}
    \caption{Examples of situations where instructing participants with text descriptions may be challenging or ambiguous: selection of a certain instance in the first round (\subref{fig:hard_clicks_descriptions:1}-\subref{fig:hard_clicks_descriptions:2}); and selecting or unselecting a certain error area in the subsequent round (\subref{fig:hard_clicks_descriptions:fp}-\subref{fig:hard_clicks_descriptions:fn}).
    }
    \label{fig:hard_clicks_descriptions}
\end{figure}

\textit{Text Description} is considered to be unbiased because users do not see the target segmentation mask, as in real-world interactive segmentation.
However, textual descriptions may be ambiguous for certain
types of instances or areas (see Figure~\ref{fig:hard_clicks_descriptions}).
In other display modes, the mask is presented, which could potentially distort the distribution of user clicks.
To choose mask-based display mode with minimal bias, we compare all modes with \textit{Text Description} mode.

Therefore, we conducted an ablation study on 100 randomly selected images from TETRIS dataset~\cite{moskalenko2024tetris}.
We used images and segmentation masks from TETRIS, and additionally manually annotated textual descriptions.
In this study, we compare the clicks obtained via considered display modes with clicks collected through \textit{Text Description}.

Clicks gathering was done on Toloka AI~\footnote{\url{https://toloka.ai/}} crowdsourcing platform.
Each participant was given a batch of 10 unique images, in each image they were required to make one click.
Each participant received on average 3 batches of images.
Participants did not receive the same image more than once.
For every display mode, different people were involved in labeling.
A participant's clicks are considered to be valid, if at least 7 out of the 10 clicks in a batch were within the object mask.
We also considered a click to be valid whether it was within the object mask or not farther than one click radius from the border.
Otherwise, clicks were disregarded as invalid.
To select an unbiased presentation strategy, for each image, we collected 25 clicks from participants using computers and 25 clicks from those using mobile devices.
After filtering, in total we obtained 47\,725 valid clicks.

To compare quantitatively display modes with unbiased \textit{Text Description} mode, we utilized the following sample-based metrics: \begin{enumerate*}[label=(\alph*)]
    \item PL$_1$ -- mean of all pairwise L$_1$-distances between click coordinates, normalized by object width and height. 
    \item WD -- Wasserstein distance~\cite{peyre2020computational} between click coordinates, normalized by object width and height.
    \item KS -- Kolmogorov-Smirnov test in 2D case~\cite{christian2020re,  peacock1983two, fasano1987multidimensional}.
    We conclude that clicks are not significantly different if a p-value is greater than 0.05.
    The indicator function is used as a metric, which equals 1 when clicks are not significantly different.
\end{enumerate*}

\begin{wraptable}{r}{0.35\linewidth}
\vspace{-1.4em}
\caption{Comparison of display modes with \textit{Text Description}.}
\label{tab:gui_comp}
\fontsize{8pt}{11pt}\selectfont
\tabcolsep=1.8pt
\centering
\begin{tabular}{cccc}
\toprule
Display mode
& PL$_1$ $\downarrow$ & KS $\uparrow$ & WD $\downarrow$ \\ \hline 
\textit{Object CutOut}        & \textbf{0.242} & \textbf{0.58} & \textbf{0.042} \\ 
\textit{Shifted CutOut}       & 0.246 & 0.56 & 0.046 \\ 
\textit{Silhouette Mask}      & 0.246 & 0.41 & 0.048 \\ 
\textit{Highlighted Instance} & 0.258 & 0.37 & 0.051 \\ 
\bottomrule
\end{tabular}
\vspace{-1em}
\end{wraptable}
The average values w.r.t. images of the listed metrics for each display mode are presented in Table~\ref{tab:gui_comp}.
The best results were obtained with the \textit{Object CutOut} method.
Note that we did not use probability map based metrics used for evaluating saliency prediction, because the actual model of clicks distribution is unknown, and we did not want to limit it in the ablation stage. 
Here, we ablated display modes only for the first interaction rounds.
We cannot examine our display modes in the subsequent rounds, as, in addition to the reference instance, the user needs to know which error should be corrected and where it is located, which is not possible to describe textually.
Thus, we assume that the best display mode for the first round is also best for the subsequent ones.

In the following, we used \textit{Object CutOut} mode to collect clicks for the remaining datasets.

\subsection{Collected Interactions}
\label{sec:collected_clicks}

\begin{wraptable}{r}{0.35\textwidth}
\vspace{-1.4em}
\caption{The number of collected clicks for each dataset in interaction rounds.
}
\label{tab:collected_clicks}
\fontsize{8pt}{11pt}\selectfont
\tabcolsep=1.8pt
\centering
\begin{tabular}{cccc}
\toprule
Dataset & First \# & Subseq. \# & Sum \# \\ 
\midrule
GrabCut  & 2\,395   & 3\,427   & 5\,822   \\
Berkeley & 4\,859   & 6\,937   & 11\,796  \\
DAVIS    & 16\,975  & 23\,687  & 40\,662  \\
COCO-MV     & 38\,097  & 53\,926  & 92\,023  \\
TETRIS   & 123\,023 & 202\,218 & 325\,241 \\ \hline
All      & 185\,349 & 290\,195 & 475\,544 \\
\bottomrule
\end{tabular}
\vspace{-1em}
\end{wraptable}
We annotated each instance in all common interactive segmentation benchmark datasets -- DAVIS~\cite{davis}, GrabCut~\cite{grabcut}, COCO-MVal~\cite{coco}, Berkeley~\cite{berkeley}, TETRIS~\cite{moskalenko2024tetris} using PC and mobile clicks.
Collected clicks were validated similarly to the ablation stage.
When annotating subsequent interaction rounds, the user should click in the area of the segmentation error.
To obtain error masks for the subsequent rounds, we applied state-of-the-art interactive segmentation methods -- SAM~\cite{kirillov2023segment}, SimpleClick~\cite{liu2022simpleclick}, and RITM~\cite{ritm} -- to all images and all clicks corresponding to those images from the first round.
Then, for each image, we selected the mask with the highest quality up to a threshold of 0.95~IoU.
We motivate it by the fact that at such a high level of quality, the user is likely to stop annotating the instance as the errors would be minimal, and even the radius of the click may exceed the size of the erroneous area.
In total, we collected 475 thousand valid clicks from users. The number of valid clicks annotated for each dataset and interaction round is presented in Table~\ref{tab:collected_clicks}.

\section{Click Simulation}

In this section, we explore models for predicting user clicks.
Firstly, the baselines are described~\eqref{sec:pred:baselines}.
Secondly, we introduce \textit{a clickability model} used for click prediction \eqref{sec:pred:model} in our interactive segmentation benchmark.
Thirdly, in \eqref{sec:pred:click_map} we describe the construction of training dataset for \textit{clickability model}.
Finally, we compare our \textit{clickability model} with baselines~\eqref{sec:pred:ablation}.

\subsection{Baseline Models}
\label{sec:pred:baselines}

\begin{figure}[!hb]
\begin{minipage}[t]{0.59\textwidth}
    \centering
    \begin{subfigure}[t]{0.19\textwidth}
        \centering
        \includegraphics[width=\textwidth]{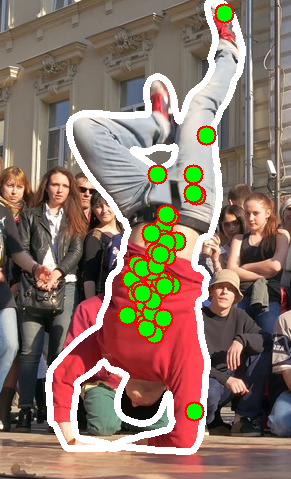}
        \caption{Clicks}
        \label{fig:click_models:fixations}
    \end{subfigure}
    \begin{subfigure}[t]{0.19\textwidth}
        \centering
        \includegraphics[width=\textwidth]{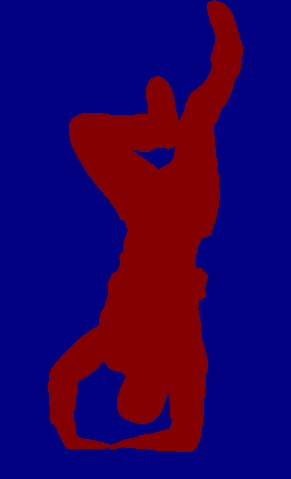}
        \caption{UD}
        \label{fig:click_models:uniform}
    \end{subfigure}
    \begin{subfigure}[t]{0.19\textwidth}
        \centering
        \includegraphics[width=\textwidth]{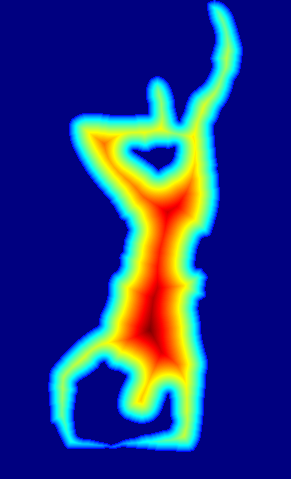}
        \caption{DT}
        \label{fig:click_models:dt}
    \end{subfigure}
    \begin{subfigure}[t]{0.19\textwidth}
        \centering
        \includegraphics[width=\textwidth]{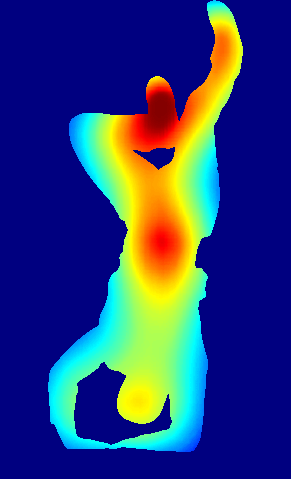}
        \caption{SM}
        \label{fig:click_models:saliency}
    \end{subfigure}
    \begin{subfigure}[t]{0.19\textwidth}
        \centering
        \includegraphics[width=\textwidth]{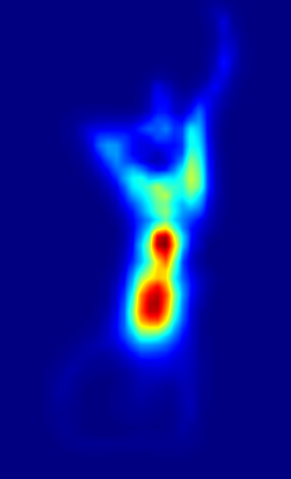}
        \caption{Ours}
        \label{fig:click_models:ours}
    \end{subfigure}
    \caption{Examples of considered clickability models:
    \subref{fig:click_models:fixations} visualizes target object (white contour) and ground-truth clicks (green points);
    \subref{fig:click_models:uniform} -- \subref{fig:click_models:saliency} depict uniform distribution (UD), distance transform (DT), and saliency map (SM) respectively;
    \subref{fig:click_models:ours} -- our predicted clickability map.
    }
    \label{fig:click_models}
\end{minipage} \hfill
\begin{minipage}[t]{0.39\textwidth}
    \vspace{-7.5em}
    \captionof{table}{Evaluation of various clickability models on real-user clicks of TETRIS validation part. Our approach outperforms existing clicking strategies in terms of the proximity of samples to real-user clicks.
    }
    \label{tab:models:eval}
    \fontsize{8pt}{11pt}\selectfont
    \tabcolsep=2pt
    \centering
    \begin{tabular}{cccccc}
    \toprule
    Model & KS\,$\uparrow$ & PL$_1$\,$\downarrow$ & WD\,$\downarrow$ & NSS\,$\uparrow$ & PDE\,$\uparrow$ \\
    \midrule
    UD & 0.10 & 0.57 & 0.17 & 3.99 & 1.36E-05 \\
    DT & 0.14 & 0.52 & 0.16 & 6.45 & 2.76E-05 \\
    SM & 0.13 & 0.51 & 0.15 & 4.79 & 1.83E-05 \\
    Ours & \textbf{0.55} & \textbf{0.40} & \textbf{0.08} & \textbf{9.11} & \textbf{4.69E-05} \\
    \bottomrule
    \end{tabular}
\end{minipage}
\end{figure}

As baselines for comparison, we considered uniform, distance, and saliency distribution models.
The uniform hypothesis postulates that the clickability of all pixels within the target area is equally distributed (see Figure~\ref{fig:click_models:uniform}).
When the area of interest is relatively small, the uniform assumption is reasonable.
However, according to this assumption, the click probability of object boundaries and their centers is equal, which is not necessarily true.
The distance transform~\cite{moskalenko2024tetris, ritm} addresses this issue by assigning greater weight to pixels in the center of the object than to those on the boundary (see Figure~\ref{fig:click_models:dt}).
Nevertheless, this transform considers only the shape of the object, neglecting human perception.
To account for human perception, saliency distribution can be used.
This is a reasonable baseline, as users look at the target area of interaction when clicking.
For the saliency baseline, we utilized a state-of-the-art model, TranSalNet~\cite{lou2022transalnet}.
The example of constructed saliency distribution is presented in Figure~\ref{fig:click_models:saliency}, details of how such map is constructed can be found in the Appendix~\ref{app:model_generalizability}.
However, saliency models are trained for free-viewing task, and do not take into consideration the setup of our task.

\subsection{Clickability Prediction Model}
\label{sec:pred:model}

\begin{wrapfigure}{r}{0.65\textwidth}
\vspace{-1em}
    \centering
    \includegraphics[width=0.65\textwidth]{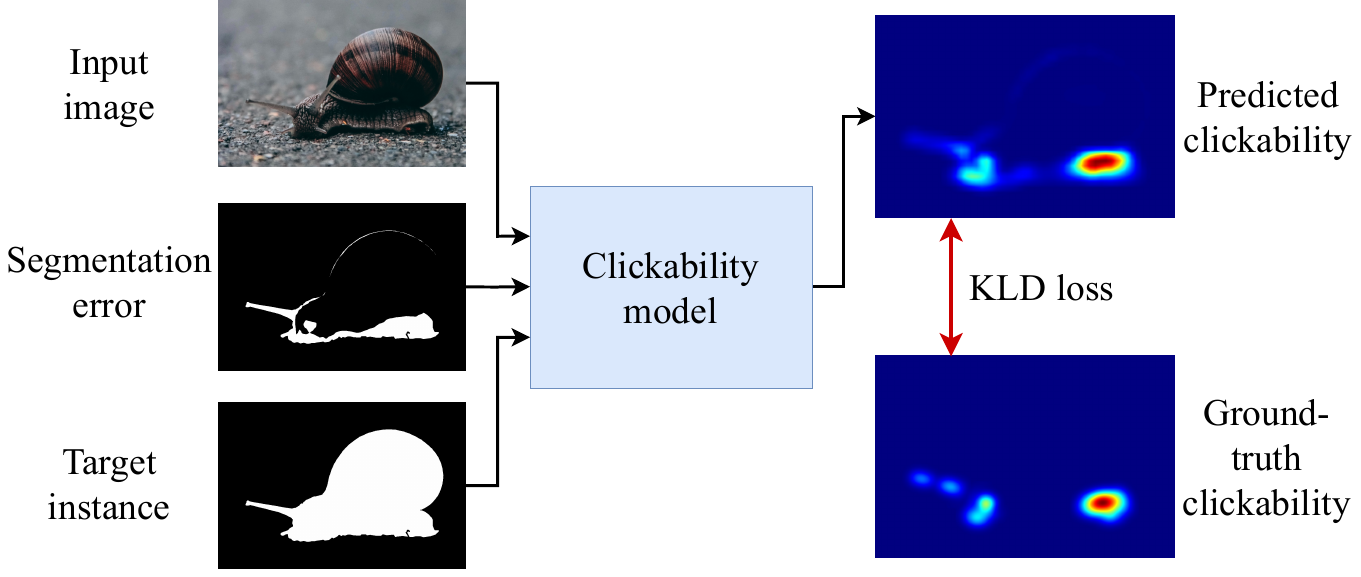}
    \caption{Proposed clickability prediction pipeline.
    }
    \label{fig:prediction_pipeline}
\vspace{-0.5em}
\end{wrapfigure}
Similar to the saliency prediction task, we formulate the task of simulating user clicks as a probabilistic problem.
Given an image, a ground truth object mask, and a segmentation error mask (FP or FN), the model should predict at each pixel the probability of being clicked.
We refer to this as a \textit{clickability map} (see details in Section~\ref{sec:pred:click_map}). 
The proposed pipeline is shown in Figure~\ref{fig:prediction_pipeline}.
As a base architecture for our model, we adapted state-of-the-art SegNeXt segmentation network~\cite{guo2022segnext}.
We input the original image into the network and concatenate the ground truth mask with the error mask, feeding the resulting tensor as an additional input to the network, a technique inspired by the Conv1S~\cite{ritm}.
We use the Kullback-Leibler divergence (KLD) loss function between the predicted and ground truth distributions.

\subsection{Clickability Maps Dataset}
\label{sec:pred:click_map}

We introduce the concept of a \textit{clickability map} as a single-channel image, such that the value of each pixel corresponds to the probability that the user will click on it during the interaction round.
We propose to use such maps to train \textit{clickability models}.

Given an image, error mask, and user clicks, the \textit{clickability map} is constructed as follows:
\begin{enumerate*}[label=(\arabic*)]
    \item initialize the map as an image of zero values;
    \item at each pixel position that was clicked, add one;
    \item \label{constuct:click_map:smooth_clicks} smooth the map by a Gaussian with some sigma, where sigma is a hyperparameter;
    \item \label{constuct:click_map:soft_mask} multiply pixel values of the map by corresponding pixel values from a \textit{soft error mask}, obtained by smoothing the original error mask by a Gaussian;
    \item normalize pixels by the map sum.
\end{enumerate*}
The proposed method is analogous to the construction of saliency maps from human eye fixations, except for step~\ref{constuct:click_map:soft_mask}.

Unlike saliency, we need to somehow constrain the most likely click positions within the boundaries of the mask.
Moreover, recall that during the collection of clicks, we considered clicks as valid if they were inside the mask or close to its border.
For these reasons, in step~\ref{constuct:click_map:soft_mask} the \textit{clickability map} is conditioned by multiplying it on the \textit{soft error mask} -- smoothing the error mask we consider the allowed radius of the border vicinity.
We smooth the error mask by Gaussian blurring with a sigma equal to the radius of the click used in the user interface (i.e., 1\% of image diagonal).
The sigma in step~\ref{constuct:click_map:smooth_clicks} simulates the probability density of clicks inside the mask.

We constructed train and validation datasets as follows.
We split images of TETRIS dataset into non-overlapping train and validation parts.
Since we do not know the real click density, for model training and validation, several sets of \textit{clickability maps} were constructed with varying magnitudes of sigma from step~\ref{constuct:click_map:smooth_clicks}.
To choose the best sigma, we conducted an ablation study, which can be found in Appendix~\ref{app:click_maps_ablation}.
Note that we constructed \textit{clickability maps} using clicks from both smartphones and PCs.
This was done to ensure that the model would predict clicks regardless of the device type.

\subsection{Models Evaluation}
\label{sec:pred:ablation}

To choose the best \textit{clickability model}, we evaluated considered models on the real-user clicks of TETRIS validation part.
Here, in addition to sample-based metrics considered above, we calculated additional metrics, that were computed based on the ground-truth clicks positions and predicted clicks distribution: PDE -- likelihood of ground-truth clicks, and NSS from saliency benchmarks~\cite{kummerer2018saliency}.
Evaluation results are presented in Table~\ref{tab:models:eval}.
Our clickability model shows the best performance.
We address the question of model generalizability in the Appendix~\ref{app:model_generalizability}.

\section{Benchmarking Interactive Segmentation}

\begin{wrapfigure}{r}{0.20\textwidth}
\vspace{-1.3em}
    \centering
    \includegraphics[width=0.2\textwidth]{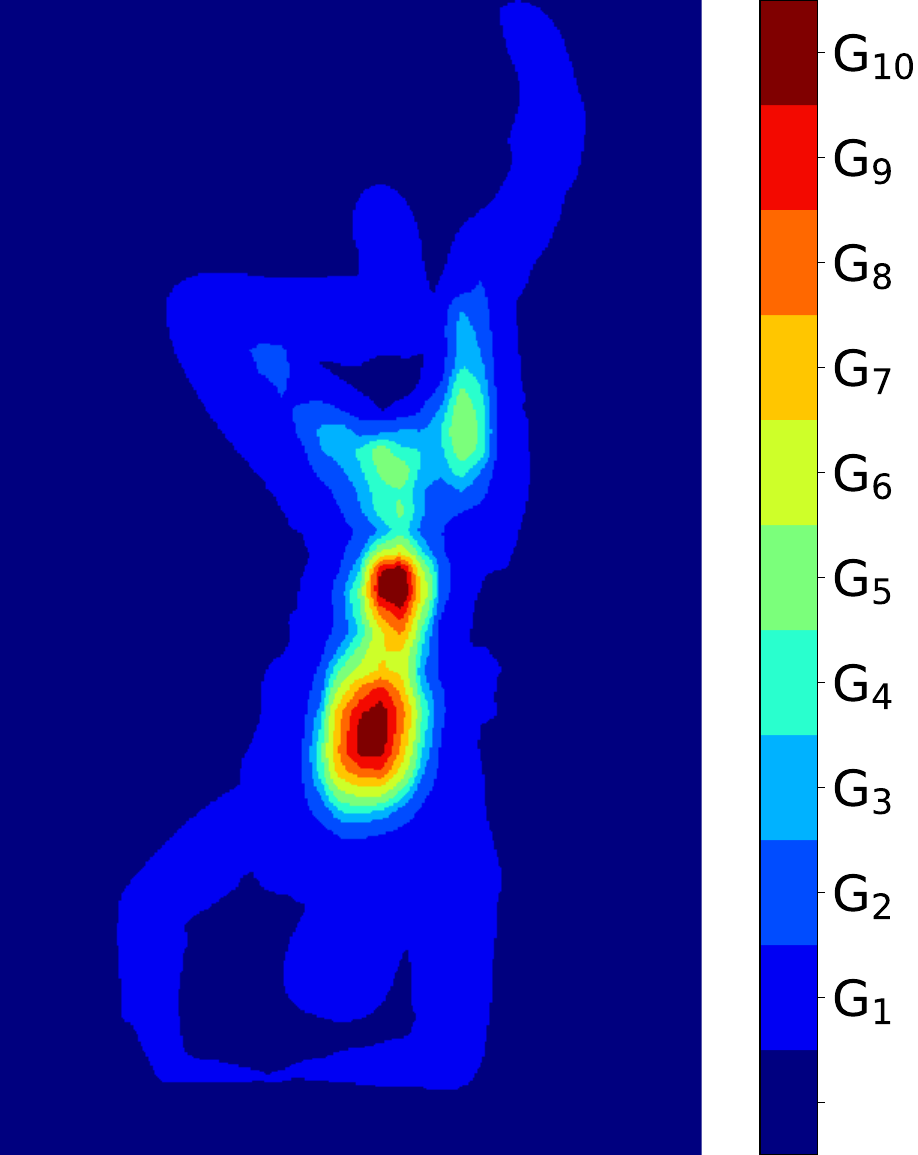}
    \caption{Spatial distribution of clicking groups for the instance in Figure~\ref{fig:click_models:fixations}.
    }
    \label{fig:groups_clicks}
\vspace{-1.5em}
\end{wrapfigure}
In this section, we introduce \textbf{RClicks} benchmark that evaluates the interactive segmentation methods according to the proposed \textit{clickability model}.
Our evaluation protocol aims to estimate not only the average annotating time but also the spread w.r.t. \textit{clicking groups}.
Our model returns a probability density for an instance. For every possible click, we have $(x, y)$ coordinates and probability. We sort clicks according to their probabilities and split them into 10 intervals (called \textit{clicking groups}) \{G$_i$\}$_{i=1}^{10}$ s.t. every interval has 10\% of total probability mass. We interpret these \textit{groups} as different user clicking patterns, and evaluate methods for each group separately.
Visualization of \textit{clicking groups} for an instance may be seen in Figure~\ref{fig:groups_clicks}.
Note, that even the probability mass of each group G$_i$ is equal, the average probability of clicks in each group increases with increase of $i$.
  
 We modify a common evaluation protocol~\cite{ritm} by replacing the \textit{baseline sampling strategy} with sampling from different \textit{clicking groups}, obtained through the \textit{clickability model}.
Specifically, (1) interactive segmentation metrics (e.g. NoC) are calculated for every instance in a dataset and group G$_i$ by sampling click from G$_i$ (weighted by the \textit{clickability model}) for every interaction round (in our experiments -- 20 rounds); 
(2) then for each instance statistics (e.g. mean and standard deviation) of sampled metrics over \textit{clicking groups} are estimated; 
(3) finally, these statistics are averaged over all instances in the dataset.

In interactive segmentation, a commonly used metric is $\textrm{NoC}_{\textrm{20}}\textrm{@}\textrm{90}$, which estimates the annotation time (in clicks, not more than 20) to achieve 90\% IoU using a particular method. We use this metric with our sampling strategy to estimate the following averaged statistics:
\begin{enumerate*}[label=(\roman*)]
\item Mean and standard deviation, which are denoted as \textbf{Sample} NoC.
\item Relative increase of Sample NoC compared to the \textit{baseline strategy} NoC. 
This statistic indicates how much extra annotation time an average user spends compared to the \textit{baseline strategy}.
We denote it as \textbf{${\Delta}$SB}.
\item Relative increase of annotation time using clicks from group G$_{1}$ over using clicks from group G$_{10}$. This metric is denoted as \textbf{${\Delta}$GR}. 
This metric represents a difference in annotation speed between two \textit{clicking groups}, that have a maximum difference of average clicking probabilities.
\end{enumerate*}

\begin{wrapfigure}{r}{0.65\textwidth}
% \vspace{-1em}
    \centering
    \includegraphics[width=0.65\textwidth]{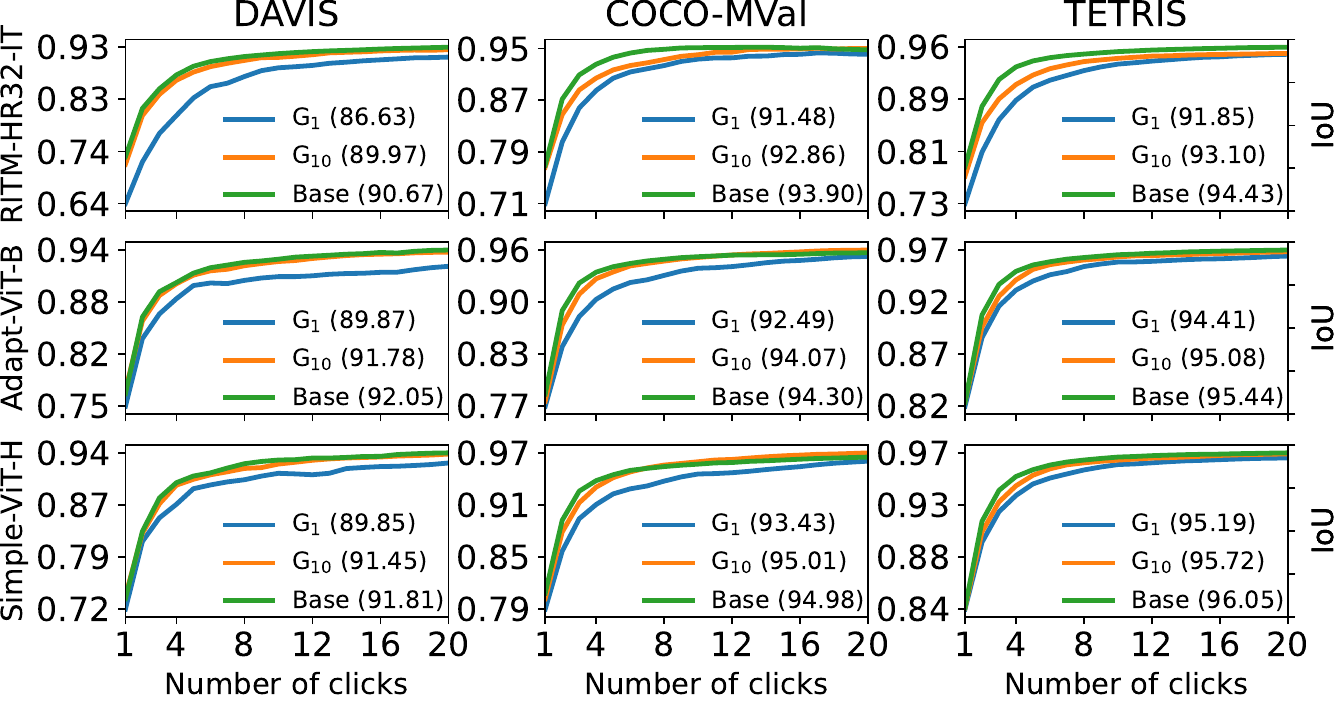}
    \caption{Mean IoU for varying number of clicks for \textit{baseline strategy}, G$_1$ and G$_{10}$ \textit{clicking groups}. IoU-AuC under these curves provided in brackets. 
    }
    \label{fig:spread}
    \vspace{2em}
    \includegraphics[width=0.65\textwidth]{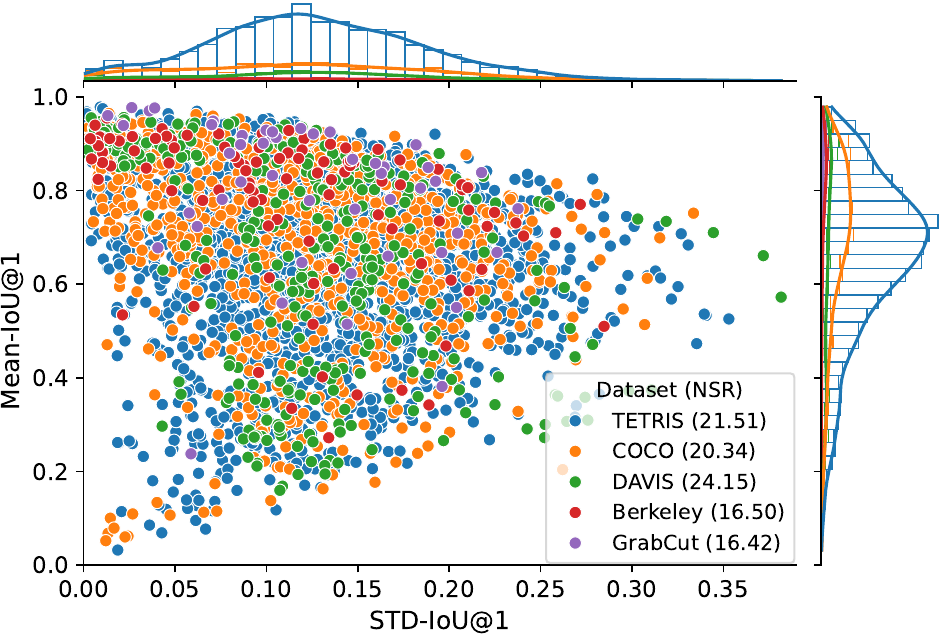}
    \caption{
    A scatter plot of the mean vs. standard deviation (STD) of IoU for the first real-users clicks.
    Each point represents the statistics for each instance, averaged across all considered segmentation methods and real clicks.
    An average NSR for each dataset is provided in brackets in the legend.
    }
    \label{fig:mean_std}
\vspace{-2em}
\end{wrapfigure}
Figure~\ref{fig:spread} shows plots of the averaged IoU versus the number of clicks for various segmentation methods when sampling clicks according to the \textit{baseline strategy}, G$_{1}$ and G$_{10}$ groups.
Overall, clicks from G$_{10}$ outperform clicks from G$_{1}$ in terms of IoU-AuC, while the \textit{baseline strategy} mostly outperforms clicks from both groups.
With a sufficient number of interactions, clicks from both G$_{10}$ and G$_{1}$ achieve high IoU.
However, clicks from G$_{1}$ require more interactions.

Estimated statistics for $\textrm{NoC}_{\textrm{20}}\textrm{@}\textrm{90}$ are presented in Table~\ref{tbl:base-d-og}.
Additional evaluation results for $\textrm{NoF}_{\textrm{20}}\textrm{@}\textrm{90}$, $\textrm{IoU-AuC}_{\textrm{20}}$ are provided in Appendix~\ref{app:additional_results}.

In addition to the evaluation results on simulated clicks, we provide evaluation results on the first round real clicks.
The evaluation on the subsequent real clicks is infeasible, since interactive segmentation in a subsequent round depends on a model output from a previous round.
However, there is no such problem in the first round, and actual performance metrics can be computed on the real clicks of the first round.
Therefore, we employed the real clicks from the first round as follows: (1) computed real-world accuracy (see Appendix~\ref{app:additional_results}); (2) compared accuracy of interactive segmentation methods using real and simulated clicks (see Appendix~\ref{app:additional_results}); and (3) estimated real-world robustness of the methods for each instance in the dataset.
We estimated the latter through IoU noise-to-signal ratio (NSR).
The greater the value of NSR, the more difficult such an instance is for segmentation.
Figure~\ref{fig:mean_std} plots a scatter of mean vs. standard deviation of IoU over real-users first clicks and segmentation methods from Table~\ref{tbl:base-d-og}.

\begin{table*}[!ht]
\centering
% Making it any smaller than 9 point with 10 point linespacing is not allowed.
\caption{Evaluation results of state-of-the-art interactive segmentation methods.
Statistics of $\textrm{NoC}_{\textrm{20}}\textrm{@}\textrm{90}$ on clicking groups \{G$_i$\}$^{10}_{i=1}$, averaged over datasets: Sample -- mean and standard deviation (std); ${\Delta}$SB -- relative increase of Sample NoC compared to \textit{baseline strategy} NoC; ${\Delta}$GR -- relative NoC increase between G$_{10}$ and G$_{1}$. 
In Data column: C+L denotes COCO+LVIS~\cite{ritm}; SBD, SA-1B and SA-V -- datasets from~\cite{sbd},~\cite{kirillov2023segment} and~\cite{ravi2024sam} respectively.
The best results are in bold, the second best are underlined, and the third best are in italics.
}
\label{tbl:base-d-og}
\fontsize{8pt}{11pt}\selectfont
\tabcolsep=2.5pt
\begin{tabular}{ccccccccccccccc}
\toprule
\multirow{3}{*}{Method}                                                   & \multirow{3}{*}{Backbone} & \multirow{3}{*}{Data} &  & \multicolumn{3}{c}{DAVIS \citep{davis}}                                                                                                                              &  & \multicolumn{3}{c}{COCO-MVal \citep{coco}}                                                                                                                           &  & \multicolumn{3}{c}{TETRIS \citep{moskalenko2024tetris}}                                                                                                              \\
                                                                          &                           &                       &  & \multicolumn{3}{c}{$\textrm{NoC}_{\textrm{20}}\textrm{@}\textrm{90}$}                                                                                                                 &  & \multicolumn{3}{c}{$\textrm{NoC}_{\textrm{20}}\textrm{@}\textrm{90}$}                                                                                                                 &  & \multicolumn{3}{c}{$\textrm{NoC}_{\textrm{20}}\textrm{@}\textrm{90}$}                                                                                                                 \\
                                                                          &                           &                       &  & \begin{tabular}[c]{@{}c@{}}Sample\\ (±std)\end{tabular} & \begin{tabular}[c]{@{}c@{}}${\Delta}$SB\\ (+\%)\end{tabular} & \begin{tabular}[c]{@{}c@{}}${\Delta}$GR\\ (+\%)\end{tabular} &  & \begin{tabular}[c]{@{}c@{}}Sample\\ (±std)\end{tabular} & \begin{tabular}[c]{@{}c@{}}${\Delta}$SB\\ (+\%)\end{tabular} & \begin{tabular}[c]{@{}c@{}}${\Delta}$GR\\ (+\%)\end{tabular} &  & \begin{tabular}[c]{@{}c@{}}Sample\\ (±std)\end{tabular} & \begin{tabular}[c]{@{}c@{}}${\Delta}$SB\\ (+\%)\end{tabular} & \begin{tabular}[c]{@{}c@{}}${\Delta}$GR\\ (+\%)\end{tabular} \\ \cline{1-3} \cline{5-7} \cline{9-11} \cline{13-15} 
GPCIS \citep{zhou2023interactivegpcis}                   & RN50                      & C+L                   &  & 6.44±0.85                                               & 16.88                                                        & 53.65                                                        &  & 4.74±1.31                                               & 26.43                                                        & 79.00                                                        &  & 3.87±0.79                                               & 19.55                                                        & 56.43                                                        \\ \cline{1-3} \cline{5-7} \cline{9-11} \cline{13-15} 
\multirow{2}{*}{CDNet \citep{chen2021conditional}}       & RN34                      & C+L                   &  & 5.95±0.73                                               & 14.95                                                        & 45.88                                                        &  & 4.13±0.85                                               & 15.15                                                        & 49.79                                                        &  & 3.10±0.53                                               & 14.16                                                        & 44.06                                                        \\
                                                                          & RN34                      & SBD                   &  & 7.87±1.25                                               & 23.39                                                        & 64.95                                                        &  & 6.36±1.29                                               & 20.86                                                        & 51.58                                                        &  & 4.51±0.76                                               & 17.08                                                        & 55.22                                                        \\ \cline{1-3} \cline{5-7} \cline{9-11} \cline{13-15} 
\multirow{5}{*}{RITM \citep{ritm}}                       & HR18                      & C+L                   &  & 6.23±0.67                                               & \underline{6.92}                                            & 16.13                                                        &  & 3.71±0.78                                               & 10.27                                                        & 20.22                                                        &  & 3.69±0.52                                               & 7.02                                                         & 13.95                                                        \\
                                                                          & HR18s-IT                  & C+L                   &  & 6.71±0.99                                               & 20.88                                                        & 54.15                                                        &  & 3.65±0.92                                               & 16.81                                                        & 33.89                                                        &  & 3.80±0.67                                               & 15.79                                                        & 32.66                                                        \\
                                                                          & HR18-IT                   & C+L                   &  & 6.15±0.83                                               & 11.37                                                        & 31.14                                                        &  & 3.22±0.83                                               & 15.84                                                        & 37.01                                                        &  & 3.48±0.60                                               & 11.59                                                        & 23.99                                                        \\
                                                                          & HR32-IT                   & C+L                   &  & 5.90±0.89                                               & 18.34                                                        & 51.07                                                        &  & 3.24±0.83                                               & 15.50                                                        & 37.31                                                        &  & 3.44±0.65                                               & 17.47                                                        & 30.69                                                        \\
                                                                          & HR18-IT                   & SBD                   &  & 7.42±1.03                                               & 17.85                                                        & 38.39                                                        &  & 4.81±1.24                                               & 17.17                                                        & 43.23                                                        &  & 4.80±0.74                                               & 11.96                                                        & 25.31                                                        \\ \cline{1-3} \cline{5-7} \cline{9-11} \cline{13-15} 
\multirow{2}{*}{AdaptClick \citep{lin2024adaptiveclick}} & ViT-B                     & C+L                   &  & 4.97±0.40                                               & 8.60                                                         & 15.14                                                        &  & 2.93±0.58                                               & 9.44                                                         & 19.75                                                        &  & 2.62±0.37                                               & 6.99                                                         & 12.94                                                        \\
                                                                          & ViT-B                     & SBD                   &  & 5.37±0.49                                               & 8.69                                                         & 17.94                                                        &  & 4.33±1.06                                               & 14.59                                                        & 35.07                                                        &  & 3.49±0.50                                               & 8.40                                                         & 16.44                                                        \\ \cline{1-3} \cline{5-7} \cline{9-11} \cline{13-15} 
\multirow{7}{*}{SimpleClick \citep{liu2022simpleclick}}  & ViT-B                     & C+L                   &  & 5.32±0.54                                               & 9.05                                                         & 26.33                                                        &  & 3.07±0.70                                               & 11.72                                                        & 23.60                                                        &  & 2.73±0.41                                               & 8.86                                                         & 16.64                                                        \\
                                                                          & ViT-L                     & C+L                   &  & 5.03±0.42                                               & 8.71                                                         & 16.67                                                        &  & \underline{2.67±0.56}                                  & 8.05                                                         & 20.88                                                        &  & \textit{2.46±0.35}                                      & 7.11                                                         & 10.01                                                        \\
                                                                          & ViT-H                     & C+L                   &  & 5.00±0.42                                               & \textit{7.06}                                                & 12.29                                                        &  & \textbf{2.57±0.54}                                      & \underline{6.14}                                            & 17.65                                                        &  & \underline{2.36±0.33}                                  & 6.94                                                         & 10.83                                                        \\
                                                                          & ViT-XT                    & SBD                   &  & 8.35±1.36                                               & 18.67                                                        & 51.05                                                        &  & 5.86±1.65                                               & 26.28                                                        & 61.63                                                        &  & 5.49±1.22                                               & 28.57                                                        & 35.40                                                        \\
                                                                          & ViT-B                     & SBD                   &  & 5.77±0.58                                               & 8.44                                                         & 25.72                                                        &  & 4.52±1.07                                               & 17.58                                                        & 36.18                                                        &  & 3.75±0.54                                               & 11.42                                                        & 19.50                                                        \\
                                                                          & ViT-L                     & SBD                   &  & 5.56±0.53                                               & 7.26                                                         & 15.97                                                        &  & 3.83±0.88                                               & 10.02                                                        & 33.06                                                        &  & 3.40±0.43                                               & 7.32                                                         & 16.15                                                        \\
                                                                          & ViT-H                     & SBD                   &  & 5.49±0.55                                               & 7.67                                                         & 23.69                                                        &  & 3.74±0.86                                               & 10.46                                                        & 31.97                                                        &  & 3.32±0.43                                               & 7.37                                                         & 16.88                                                        \\ \cline{1-3} \cline{5-7} \cline{9-11} \cline{13-15} 
CFR-ICL \citep{sun2023cfricl}                            & ViT-H                     & C+L                   &  & \textit{4.53±0.46}                                      & 9.32                                                         & 18.47                                                        &  & \textit{2.70±0.63}                                      & 9.58                                                         & 24.13                                                        &  & \textbf{2.12±0.34}                                      & 8.76                                                         & 14.33                                                        \\ \cline{1-3} \cline{5-7} \cline{9-11} \cline{13-15} 
MobileSAM \citep{zhang2023faster}                        & ViT-Tiny                  & SA-1B                 &  & 5.96±0.56                                               & 8.63                                                         & 15.39                                                        &  & 5.25±0.78                                               & 9.79                                                         & 19.78                                                        &  & 3.42±0.48                                               & 7.47                                                         & 12.69                                                        \\ \cline{1-3} \cline{5-7} \cline{9-11} \cline{13-15} 
\multirow{3}{*}{SAM \citep{kirillov2023segment}}         & ViT-B                     & SA-1B                 &  & 5.30±0.53                                               & 8.26                                                         & \textit{11.27}                                               &  & 4.91±0.79                                               & 9.88                                                         & 15.73                                                        &  & 3.04±0.51                                               & 11.17                                                        & 10.06                                                        \\
                                                                          & ViT-L                     & SA-1B                 &  & 5.21±0.41                                               & 8.82                                                         & 11.59                                                        &  & 4.81±0.63                                               & 8.89                                                         & 14.97                                                        &  & 2.60±0.40                                               & 8.11                                                         & 7.08                                                         \\
                                                                          & ViT-H                     & SA-1B                 &  & 5.42±0.49                                               & 8.00                                                         & 15.02                                                        &  & 5.14±0.68                                               & 7.63                                                         & 15.61                                                        &  & 2.66±0.38                                               & \textbf{5.95}                                                & 8.50                                                         \\ \cline{1-3} \cline{5-7} \cline{9-11} \cline{13-15} 
\multirow{3}{*}{SAM-HQ \citep{ke2023segment}}            & ViT-B                     & SA-1B                 &  & 5.32±0.50                                               & 7.45                                                         & 13.03                                                        &  & 5.39±0.89                                               & 12.48                                                        & 16.68                                                        &  & 3.35±0.67                                               & 16.41                                                        & 10.74                                                        \\
                                                                          & ViT-L                     & SA-1B                 &  & 5.19±0.48                                               & 8.58                                                         & 15.69                                                        &  & 5.05±0.74                                               & 9.64                                                         & 13.50                                                        &  & 2.81±0.51                                               & 11.02                                                        & 7.69                                                         \\
                                                                          & ViT-H                     & SA-1B                 &  & 5.16±0.44                                               & 8.15                                                         & 18.36                                                        &  & 4.97±0.68                                               & 7.71                                                         & \textit{12.36}                                               &  & 2.75±0.41                                               & \textit{6.78}                                                & 7.95                                                         \\ \cline{1-7} \cline{9-11} \cline{13-15} 
\multirow{4}{*}{SAM 2 \citep{ravi2024sam}}               & Hiera-T                   & SA-V                  &  & 4.65±0.28                                               & \textbf{4.86}                                                & \textbf{7.46}                                                &  & 3.86±0.64                                               & 7.79                                                         & 13.14                                                        &  & 3.11±0.50                                               & 9.45                                                         & \underline{3.57}                                  \\
                                                                          & Hiera-B+                  & SA-V                  &  & 4.67±0.33                                               & 8.49                                                         & 15.86                                                        &  & 3.75±0.61                                               & \textit{7.44}                                                & 12.67                                                        &  & 3.02±0.47                                               & 9.51                                                         & \textit{4.79}                                                \\
                                                                          & Hiera-L                   & SA-V                  &  & \textit{4.61±0.29}                                      & 9.51                                                         & 13.28                                                        &  & 3.84±0.62                                               & 9.12                                                         & 12.35                                                        &  & 2.83±0.41                                               & 7.46                                                         & \textit{4.10}                                                \\
                                                                          & Hiera-H                   & SA-V                  &  & \textbf{4.39±0.23}                                      & 7.55                                                         & 10.03                                                        &  & 3.42±0.51                                               & \textbf{6.12}                                                & \textbf{9.34}                                                &  & 2.74±0.38                                               & \underline{6.51}                                            & 4.87                                                         \\ \cline{1-3} \cline{5-7} \cline{9-11} \cline{13-15} 
\multirow{4}{*}{SAM 2.1 \citep{ravi2024sam}}             & Hiera-T                   & SA-V                  &  & 4.67±0.32                                               & 7.08                                                         & \underline{8.99}                                            &  & 3.91±0.68                                               & 8.45                                                         & \textit{11.88}                                               &  & 3.11±0.50                                               & 9.75                                                         & \textbf{3.35}                                                \\
                                                                          & Hiera-B+                  & SA-V                  &  & 4.63±0.32                                               & 9.72                                                         & 14.30                                                        &  & 3.76±0.62                                               & 8.16                                                         & 12.35                                                        &  & 3.04±0.49                                               & 9.59                                                         & 4.70                                                         \\
                                                                          & Hiera-L                   & SA-V                  &  & 4.67±0.32                                               & 11.75                                                        & 15.39                                                        &  & 3.88±0.62                                               & 7.47                                                         & 11.95                                                        &  & 2.87±0.43                                               & 8.35                                                         & 4.51                                                         \\
                                                                          & Hiera-H                   & SA-V                  &  & \underline{4.44±0.25}                                  & 10.35                                                        & \textit{9.48}                                                &  & 3.51±0.52                                               & \textit{6.78}                                                & \underline{9.91}                                            &  & 2.81±0.39                                               & 7.41                                                         & 4.50                                            
                                                                          
                                                                          \\
                                                                          \bottomrule            
\end{tabular}
\end{table*}

\FloatBarrier

\section{Discussion}
A review of Table~\ref{tbl:base-d-og}, Figures~\ref{fig:spread} and~\ref{fig:mean_std} leads to the following conclusions.
First, according to ${\Delta}$SB,  \textbf{\textit{baseline strategy} underestimates the real-world annotation time from 5\% up to 29\%}.
This implies that the baseline benchmark may significantly underestimate the real-world annotation costs. 
Consequently, our benchmark may be employed for a more accurate estimation of annotation costs.

Then, according to ${\Delta}$GR, \textbf{annotation time of users from different \textit{clicking groups} varies from 3\% up to 79\%}.
The observed variations in ${\Delta}$GR indicate that segmentation methods are unstable w.r.t. click positions in the image.

According to Sample and ${\Delta}$GR values, the best annotation time is achieved by SAM 2 Hiera-H (on DAVIS), CFR-ICL (on TETRIS) and SimpleClick ViT-H (on COCO-MVal).
The two latter methods are less robust compared to SAM-like methods, which perform more consistently across \textit{clicking groups}.
However, SAM 2 Hiera-H backbone is less robust than Hiera-T backbone on DAVIS dataset. 
This indicates that \textbf{there is currently no segmentation method that is optimal in terms of both performance and robustness on all datasets}. 
Consequently, \textbf{developers should select a method in accordance with their requirements}.

Finally, points in the bottom-right part of Figure~\ref{fig:mean_std} correspond to instances with high NSR. These values may be utilized \textbf{to identify and analyze hard instances in the datasets}. 
Additionally, according to averaged NSR, we \textbf{identified the hardest dataset for annotation}, it is DAVIS with 24.15 NSR.

\section{Conclusion}
\label{concl_ref}
In this paper, we presented RClicks -- a benchmark for interactive segmentation methods, that evaluates both real-world quality and robustness with respect to different clicking patterns.
Using the developed unbiased presentation strategy, we collected the multi-round real-user click dataset.
We developed the \textit{clickability model} that can be utilized to estimate click probabilities and sample realistic user clicks.
By employing this model in our benchmark, we demonstrated that \textit{baseline strategy} may overestimate methods' performance in the real world. 
Furthermore, our analysis showed that there is currently no interactive segmentation method that is optimal in terms of both performance and robustness on all datasets.
Additionally, we evaluated segmentation methods using real-user clicks of the first round and proposed a methodology to estimate the instance difficulty for state-of-the-art methods.
We hope RClicks will facilitate the advancement of interactive segmentation methods that provide optimal user experiences in real-world scenarios.

\FloatBarrier

% \bibliographystyle{unsrtnat}
% \bibliography{paper}

% \newpage

\addcontentsline{toc}{section}{References}
\bibliographystyle{unsrtnat}
\bibliography{main}

\begin{thebibliography}{54}
\providecommand{\natexlab}[1]{#1}
\providecommand{\url}[1]{\texttt{#1}}
\expandafter\ifx\csname urlstyle\endcsname\relax
  \providecommand{\doi}[1]{doi: #1}\else
  \providecommand{\doi}{doi: \begingroup \urlstyle{rm}\Url}\fi

\bibitem[Kirillov et~al.(2023)Kirillov, Mintun, Ravi, Mao, Rolland, Gustafson, Xiao, Whitehead, Berg, Lo, et~al.]{kirillov2023segment}
Alexander Kirillov, Eric Mintun, Nikhila Ravi, Hanzi Mao, Chloe Rolland, Laura Gustafson, Tete Xiao, Spencer Whitehead, Alexander~C Berg, Wan-Yen Lo, et~al.
\newblock Segment anything.
\newblock \emph{arXiv preprint arXiv:2304.02643}, 2023.

\bibitem[Ravi et~al.(2024)Ravi, Gabeur, Hu, Hu, Ryali, Ma, Khedr, R{\"a}dle, Rolland, Gustafson, et~al.]{ravi2024sam}
Nikhila Ravi, Valentin Gabeur, Yuan-Ting Hu, Ronghang Hu, Chaitanya Ryali, Tengyu Ma, Haitham Khedr, Roman R{\"a}dle, Chloe Rolland, Laura Gustafson, et~al.
\newblock Sam 2: Segment anything in images and videos.
\newblock \emph{arXiv preprint arXiv:2408.00714}, 2024.

\bibitem[Ke et~al.(2023)Ke, Ye, Danelljan, Liu, Tai, Tang, and Yu]{ke2023segment}
Lei Ke, Mingqiao Ye, Martin Danelljan, Yifan Liu, Yu-Wing Tai, Chi-Keung Tang, and Fisher Yu.
\newblock Segment anything in high quality.
\newblock \emph{arXiv preprint arXiv:2306.01567}, 2023.

\bibitem[Xie et~al.(2024)Xie, Guan, Jiang, Yi, Ding, Lu, and Zhang]{xie2024pasam}
Zhaozhi Xie, Bochen Guan, Weihao Jiang, Muyang Yi, Yue Ding, Hongtao Lu, and Lei Zhang.
\newblock Pa-sam: Prompt adapter sam for high-quality image segmentation.
\newblock \emph{2024 IEEE International Conference on Multimedia and Expo (ICME)}, 2024.

\bibitem[Hu et~al.(2023)Hu, Xia, Ju, and Li]{hu2023sam}
Chuanfei Hu, Tianyi Xia, Shenghong Ju, and Xinde Li.
\newblock When sam meets medical images: An investigation of segment anything model (sam) on multi-phase liver tumor segmentation, 2023.

\bibitem[Zhou et~al.(2023{\natexlab{a}})Zhou, Zhang, Zhou, Wu, and Gong]{zhou2023sam}
Tao Zhou, Yizhe Zhang, Yi~Zhou, Ye~Wu, and Chen Gong.
\newblock Can sam segment polyps?, 2023{\natexlab{a}}.

\bibitem[Biswas(2023)]{biswas2023polypsam}
Risab Biswas.
\newblock Polyp-sam++: Can a text guided sam perform better for polyp segmentation?, 2023.

\bibitem[Zhang and Jiao(2023)]{zhang2023segment}
Yichi Zhang and Rushi Jiao.
\newblock How segment anything model (sam) boost medical image segmentation?
\newblock \emph{arXiv preprint arXiv:2305.03678}, 2023.

\bibitem[Zhang et~al.(2024)Zhang, Shen, and Jiao]{SAM4MIS}
Yichi Zhang, Zhenrong Shen, and Rushi Jiao.
\newblock Segment anything model for medical image segmentation: Current applications and future directions.
\newblock \emph{Computers in Biology and Medicine}, 171:\penalty0 108238, 2024.

\bibitem[Mazurowski et~al.(2023)Mazurowski, Dong, Gu, Yang, Konz, and Zhang]{mazurowski2023segment}
Maciej~A Mazurowski, Haoyu Dong, Hanxue Gu, Jichen Yang, Nicholas Konz, and Yixin Zhang.
\newblock Segment anything model for medical image analysis: an experimental study.
\newblock \emph{Medical Image Analysis}, 89:\penalty0 102918, 2023.

\bibitem[Yang et~al.(2023)Yang, Wu, He, Zhao, and Liu]{yang2023sam3d}
Yunhan Yang, Xiaoyang Wu, Tong He, Hengshuang Zhao, and Xihui Liu.
\newblock Sam3d: Segment anything in 3d scenes, 2023.

\bibitem[Xu et~al.(2023)Xu, Yin, Qiu, Liu, Tong, and Han]{xu2023sampro3d}
Mutian Xu, Xingyilang Yin, Lingteng Qiu, Yang Liu, Xin Tong, and Xiaoguang Han.
\newblock Sampro3d: Locating sam prompts in 3d for zero-shot scene segmentation.
\newblock \emph{arXiv preprint arXiv:2311.17707}, 2023.

\bibitem[Cheng et~al.(2023)Cheng, Li, Xu, Li, Yang, Wang, and Yang]{cheng2023segment}
Yangming Cheng, Liulei Li, Yuanyou Xu, Xiaodi Li, Zongxin Yang, Wenguan Wang, and Yi~Yang.
\newblock Segment and track anything.
\newblock \emph{arXiv preprint arXiv:2305.06558}, 2023.

\bibitem[Moskalenko et~al.(2024)Moskalenko, Shakhuro, Vorontsova, Konushin, Antonov, Krapukhin, Shepelev, and Soshin]{moskalenko2024tetris}
Andrey Moskalenko, Vlad Shakhuro, Anna Vorontsova, Anton Konushin, Anton Antonov, Alexander Krapukhin, Denis Shepelev, and Konstantin Soshin.
\newblock Tetris: Towards exploring the robustness of interactive segmentation.
\newblock \emph{arXiv preprint arXiv:2402.06132}, 2024.

\bibitem[Rother et~al.(2004)Rother, Kolmogorov, and Blake]{grabcut}
Carsten Rother, Vladimir Kolmogorov, and Andrew Blake.
\newblock Grabcut -- interactive foreground extraction using iterated graph cuts.
\newblock \emph{ACM transactions on graphics (TOG)}, 23\penalty0 (3):\penalty0 309--314, 2004.

\bibitem[McGuinness and O’connor(2010)]{berkeley-intro}
Kevin McGuinness and Noel~E O’connor.
\newblock A comparative evaluation of interactive segmentation algorithms.
\newblock \emph{Pattern Recognition}, 43\penalty0 (2):\penalty0 434--444, 2010.

\bibitem[Lin et~al.(2014{\natexlab{a}})Lin, Maire, Belongie, Hays, Perona, Ramanan, Doll{\'a}r, and Zitnick]{coco}
Tsung-Yi Lin, Michael Maire, Serge Belongie, James Hays, Pietro Perona, Deva Ramanan, Piotr Doll{\'a}r, and C~Lawrence Zitnick.
\newblock Microsoft coco: Common objects in context.
\newblock In \emph{ECCV}, 2014{\natexlab{a}}.

\bibitem[Perazzi et~al.(2016)Perazzi, Pont-Tuset, McWilliams, Van~Gool, Gross, and Sorkine-Hornung]{davis}
Federico Perazzi, Jordi Pont-Tuset, Brian McWilliams, Luc Van~Gool, Markus Gross, and Alexander Sorkine-Hornung.
\newblock A benchmark dataset and evaluation methodology for video object segmentation.
\newblock In \emph{CVPR}, 2016.

\bibitem[Sofiiuk et~al.(2022)Sofiiuk, Petrov, and Konushin]{ritm}
Konstantin Sofiiuk, Ilya~A. Petrov, and Anton Konushin.
\newblock Reviving iterative training with mask guidance for interactive segmentation.
\newblock In \emph{2022 IEEE International Conference on Image Processing (ICIP)}, pages 3141--3145, 2022.
\newblock \doi{10.1109/ICIP46576.2022.9897365}.

\bibitem[Agustsson et~al.(2019)Agustsson, Uijlings, and Ferrari]{ferrari2019scribbles}
Eirikur Agustsson, Jasper R.~R. Uijlings, and Vittorio Ferrari.
\newblock Interactive full image segmentation by considering all regions jointly.
\newblock In \emph{CVPR}, 2019.

\bibitem[Gueziri et~al.(2017)Gueziri, McGuffin, and Laporte]{gueziri2017latency}
Houssem-Eddine Gueziri, Michael McGuffin, and Catherine Laporte.
\newblock Latency management in scribble-based interactive segmentation of medical images.
\newblock \emph{IEEE Transactions on Biomedical Engineering}, 2017.

\bibitem[Popenova et~al.(2023)Popenova, Galeev, Vorontsova, and Konushin]{popenova2023contour}
Polina Popenova, Danil Galeev, Anna Vorontsova, and Anton Konushin.
\newblock Contour-based interactive segmentation.
\newblock In \emph{Proceedings of the Thirty-Second International Joint Conference on Artificial Intelligence}, pages 1322--1330, 2023.

\bibitem[Cheng et~al.(2021)Cheng, Xu, Jiang, and Wang]{cheng2021flexible}
Hang Cheng, Shugong Xu, Xiufeng Jiang, and Rongrong Wang.
\newblock Deep image matting with flexible guidance input.
\newblock In \emph{BMVC}, 2021.

\bibitem[Xu et~al.(2016)Xu, Price, Cohen, Yang, and Huang]{deep-object-selection}
Ning Xu, Brian Price, Scott Cohen, Jimei Yang, and Thomas~S Huang.
\newblock Deep interactive object selection.
\newblock In \emph{CVPR}, 2016.

\bibitem[Liu et~al.(2022)Liu, Xu, Bertasius, and Niethammer]{liu2022simpleclick}
Qin Liu, Zhenlin Xu, Gedas Bertasius, and Marc Niethammer.
\newblock Simpleclick: Interactive image segmentation with simple vision transformers.
\newblock \emph{arXiv preprint arXiv:2210.11006}, 2022.

\bibitem[Sun et~al.(2023)Sun, Xian, Xu, Yao, and Capriotti]{sun2023cfricl}
Shoukun Sun, Min Xian, Fei Xu, Tiankai Yao, and Luca Capriotti.
\newblock Cfr-icl: Cascade-forward refinement with iterative click loss for interactive image segmentation.
\newblock \emph{arXiv preprint arXiv:2303.05620}, 2023.

\bibitem[Zhou et~al.(2023{\natexlab{b}})Zhou, Wang, Zhao, Li, Huang, Meng, and Zheng]{zhou2023interactivegpcis}
Minghao Zhou, Hong Wang, Qian Zhao, Yuexiang Li, Yawen Huang, Deyu Meng, and Yefeng Zheng.
\newblock Interactive segmentation as gaussion process classification.
\newblock In \emph{Proceedings of the IEEE/CVF Conference on Computer Vision and Pattern Recognition}, pages 19488--19497, 2023{\natexlab{b}}.

\bibitem[Zhang et~al.(2023)Zhang, Han, Qiao, Kim, Bae, Lee, and Hong]{zhang2023faster}
Chaoning Zhang, Dongshen Han, Yu~Qiao, Jung~Uk Kim, Sung-Ho Bae, Seungkyu Lee, and Choong~Seon Hong.
\newblock Faster segment anything: Towards lightweight sam for mobile applications.
\newblock \emph{arXiv preprint arXiv:2306.14289}, 2023.

\bibitem[Martin et~al.(2001)Martin, Fowlkes, Tal, and Malik]{berkeley}
David Martin, Charless Fowlkes, Doron Tal, and Jitendra Malik.
\newblock A database of human segmented natural images and its application to evaluating segmentation algorithms and measuring ecological statistics.
\newblock In \emph{ICCV}, 2001.

\bibitem[Everingham et~al.(2012)Everingham, Van~Gool, Williams, Winn, and Zisserman]{pascal-voc-2012}
M.~Everingham, L.~Van~Gool, C.~K.~I. Williams, J.~Winn, and A.~Zisserman.
\newblock The {PASCAL} {V}isual {O}bject {C}lasses {C}hallenge 2012 {(VOC2012)} {R}esults, 2012.

\bibitem[Li et~al.(2018)Li, Chen, and Koltun]{latent-diversity}
Zhuwen Li, Qifeng Chen, and Vladlen Koltun.
\newblock Interactive image segmentation with latent diversity.
\newblock In \emph{CVPR}, 2018.

\bibitem[Hariharan et~al.(2011)Hariharan, Arbelaez, Bourdev, Maji, and Malik]{sbd}
Bharath Hariharan, Pablo Arbelaez, Lubomir Bourdev, Subhransu Maji, and Jitendra Malik.
\newblock Semantic contours from inverse detectors.
\newblock In \emph{ICCV}, 2011.

\bibitem[Lou et~al.(2022{\natexlab{a}})Lou, Lin, Marshall, Saupe, and Liu]{TranSalNet}
Jianxun Lou, Hanhe Lin, David Marshall, Dietmar Saupe, and Hantao Liu.
\newblock Transalnet: Towards perceptually relevant visual saliency prediction.
\newblock \emph{Neurocomputing}, 2022{\natexlab{a}}.
\newblock ISSN 0925-2312.
\newblock \doi{https://doi.org/10.1016/j.neucom.2022.04.080}.

\bibitem[Kroner et~al.(2020)Kroner, Senden, Driessens, and Goebel]{kroner2020contextual}
Alexander Kroner, Mario Senden, Kurt Driessens, and Rainer Goebel.
\newblock Contextual encoder-decoder network for visual saliency prediction.
\newblock \emph{Neural Networks}, 129:\penalty0 261--270, 2020.
\newblock ISSN 0893-6080.
\newblock \doi{https://doi.org/10.1016/j.neunet.2020.05.004}.
\newblock URL \url{http://www.sciencedirect.com/science/article/pii/S0893608020301660}.

\bibitem[Judd et~al.(2009)Judd, Ehinger, Durand, and Torralba]{judd2009learning}
Tilke Judd, Krista Ehinger, Frédo Durand, and Antonio Torralba.
\newblock Learning to predict where humans look.
\newblock In \emph{2009 IEEE 12th International Conference on Computer Vision}, pages 2106--2113, 2009.
\newblock \doi{10.1109/ICCV.2009.5459462}.

\bibitem[Judd et~al.(2012)Judd, Durand, and Torralba]{judd2012benchmark}
Tilke Judd, Fr{\'e}do Durand, and Antonio Torralba.
\newblock A benchmark of computational models of saliency to predict human fixations.
\newblock In \emph{MIT Technical Report}, 2012.

\bibitem[Borji and Itti(2015)]{CAT2000}
Ali Borji and Laurent Itti.
\newblock Cat2000: A large scale fixation dataset for boosting saliency research.
\newblock \emph{CVPR 2015 workshop on "Future of Datasets"}, 2015.
\newblock arXiv preprint arXiv:1505.03581.

\bibitem[Jiang et~al.(2015)Jiang, Huang, Duan, and Zhao]{jiang2015salicon}
Ming Jiang, Shengsheng Huang, Juanyong Duan, and Qi~Zhao.
\newblock Salicon: Saliency in context.
\newblock In \emph{Proceedings of the IEEE conference on computer vision and pattern recognition}, pages 1072--1080, 2015.

\bibitem[Tavakoli et~al.(2017)Tavakoli, Ahmed, Borji, and Laaksonen]{tavakoli2017saliency}
Hamed~R Tavakoli, Fawad Ahmed, Ali Borji, and Jorma Laaksonen.
\newblock Saliency revisited: Analysis of mouse movements versus fixations.
\newblock In \emph{Proceedings of the ieee conference on computer vision and pattern recognition}, pages 1774--1782, 2017.

\bibitem[Lin et~al.(2014{\natexlab{b}})Lin, Maire, Belongie, Hays, Perona, Ramanan, Doll{\'a}r, and Zitnick]{lin2014microsoft}
Tsung-Yi Lin, Michael Maire, Serge Belongie, James Hays, Pietro Perona, Deva Ramanan, Piotr Doll{\'a}r, and C~Lawrence Zitnick.
\newblock Microsoft coco: Common objects in context.
\newblock In \emph{Computer Vision--ECCV 2014: 13th European Conference, Zurich, Switzerland, September 6-12, 2014, Proceedings, Part V 13}, pages 740--755. Springer, 2014{\natexlab{b}}.

\bibitem[Peyré and Cuturi(2020)]{peyre2020computational}
Gabriel Peyré and Marco Cuturi.
\newblock Computational optimal transport, 2020.

\bibitem[Christian(2020)]{christian2020re}
Sam Christian.
\newblock {Re-examining the evidence of the Hercules–Corona Borealis Great Wall}.
\newblock \emph{Monthly Notices of the Royal Astronomical Society}, 495\penalty0 (4):\penalty0 4291--4296, 05 2020.
\newblock ISSN 0035-8711.
\newblock \doi{10.1093/mnras/staa1448}.

\bibitem[Peacock(1983)]{peacock1983two}
John~A Peacock.
\newblock Two-dimensional goodness-of-fit testing in astronomy.
\newblock \emph{Monthly Notices of the Royal Astronomical Society}, 202\penalty0 (3):\penalty0 615--627, 1983.

\bibitem[Fasano and Franceschini(1987)]{fasano1987multidimensional}
Giovanni Fasano and Alberto Franceschini.
\newblock A multidimensional version of the kolmogorov--smirnov test.
\newblock \emph{Monthly Notices of the Royal Astronomical Society}, 225\penalty0 (1):\penalty0 155--170, 1987.

\bibitem[Lou et~al.(2022{\natexlab{b}})Lou, Lin, Marshall, Saupe, and Liu]{lou2022transalnet}
Jianxun Lou, Hanhe Lin, David Marshall, Dietmar Saupe, and Hantao Liu.
\newblock Transalnet: Towards perceptually relevant visual saliency prediction.
\newblock \emph{Neurocomputing}, 494:\penalty0 455--467, 2022{\natexlab{b}}.

\bibitem[Guo et~al.(2022)Guo, Lu, Hou, Liu, Cheng, and Hu]{guo2022segnext}
Meng-Hao Guo, Cheng-Ze Lu, Qibin Hou, Zhengning Liu, Ming-Ming Cheng, and Shi-Min Hu.
\newblock Segnext: Rethinking convolutional attention design for semantic segmentation.
\newblock \emph{arXiv preprint arXiv:2209.08575}, 2022.

\bibitem[Kummerer et~al.(2018)Kummerer, Wallis, and Bethge]{kummerer2018saliency}
Matthias Kummerer, Thomas S.~A. Wallis, and Matthias Bethge.
\newblock Saliency benchmarking made easy: Separating models, maps and metrics.
\newblock In \emph{Proceedings of the European Conference on Computer Vision (ECCV)}, September 2018.

\bibitem[Chen et~al.(2021)Chen, Zhao, Yu, Zhang, and Duan]{chen2021conditional}
Xi~Chen, Zhiyan Zhao, Feiwu Yu, Yilei Zhang, and Manni Duan.
\newblock Conditional diffusion for interactive segmentation.
\newblock In \emph{Proceedings of the IEEE/CVF International Conference on Computer Vision}, pages 7345--7354, 2021.

\bibitem[Lin et~al.(2024)Lin, Chen, Yang, Roitberg, Li, Li, and Li]{lin2024adaptiveclick}
Jiacheng Lin, Jiajun Chen, Kailun Yang, Alina Roitberg, Siyu Li, Zhiyong Li, and Shutao Li.
\newblock Adaptiveclick: Click-aware transformer with adaptive focal loss for interactive image segmentation.
\newblock \emph{IEEE Transactions on Neural Networks and Learning Systems}, 2024.

\bibitem[Delatolas et~al.(2024)Delatolas, Kalogeiton, and Papadopoulos]{delatolas2024learning}
Thanos Delatolas, Vicky Kalogeiton, and Dim~P Papadopoulos.
\newblock Learning the what and how of annotation in video object segmentation.
\newblock In \emph{Proceedings of the IEEE/CVF Winter Conference on Applications of Computer Vision}, pages 6951--6961, 2024.

\bibitem[Papadopoulos et~al.(2014)Papadopoulos, Clarke, Keller, and Ferrari]{Papadopoulos2014TrainingOC}
Dim~P. Papadopoulos, Alasdair D.~F. Clarke, Frank Keller, and Vittorio Ferrari.
\newblock Training object class detectors from eye tracking data.
\newblock In \emph{European Conference on Computer Vision}, 2014.
\newblock URL \url{https://api.semanticscholar.org/CorpusID:14119147}.

\bibitem[Papadopoulos et~al.(2017)Papadopoulos, Uijlings, Keller, and Ferrari]{papadopoulos2017training}
Dim~P Papadopoulos, Jasper~RR Uijlings, Frank Keller, and Vittorio Ferrari.
\newblock Training object class detectors with click supervision.
\newblock In \emph{Proceedings of the IEEE Conference on Computer Vision and Pattern Recognition}, pages 6374--6383, 2017.

\bibitem[Bearman et~al.(2016)Bearman, Russakovsky, Ferrari, and Fei-Fei]{bearman2016s}
Amy Bearman, Olga Russakovsky, Vittorio Ferrari, and Li~Fei-Fei.
\newblock What’s the point: Semantic segmentation with point supervision.
\newblock In \emph{European conference on computer vision}, pages 549--565. Springer, 2016.

\bibitem[Gebru et~al.(2021)Gebru, Morgenstern, Vecchione, Vaughan, Wallach, Daumé~III, and Crawford]{datasheet}
Timnit Gebru, Jamie Morgenstern, Briana Vecchione, Jennifer~Wortman Vaughan, Hanna Wallach, Hal Daumé~III, and Kate Crawford.
\newblock Datasheets for datasets.
\newblock volume~64, pages 86--92, 2021.

\end{thebibliography}


\begin{thebibliography}{21}
\providecommand{\natexlab}[1]{#1}
\providecommand{\url}[1]{\texttt{#1}}
\expandafter\ifx\csname urlstyle\endcsname\relax
  \providecommand{\doi}[1]{doi: #1}\else
  \providecommand{\doi}{doi: \begingroup \urlstyle{rm}\Url}\fi

\bibitem[Guo et~al.(2022)Guo, Lu, Hou, Liu, Cheng, and Hu]{guo2022segnext}
Meng-Hao Guo, Cheng-Ze Lu, Qibin Hou, Zhengning Liu, Ming-Ming Cheng, and Shi-Min Hu.
\newblock Segnext: Rethinking convolutional attention design for semantic segmentation.
\newblock \emph{arXiv preprint arXiv:2209.08575}, 2022.

\bibitem[Sofiiuk et~al.(2022)Sofiiuk, Petrov, and Konushin]{ritm}
Konstantin Sofiiuk, Ilya~A. Petrov, and Anton Konushin.
\newblock Reviving iterative training with mask guidance for interactive segmentation.
\newblock In \emph{2022 IEEE International Conference on Image Processing (ICIP)}, pages 3141--3145, 2022.
\newblock \doi{10.1109/ICIP46576.2022.9897365}.

\bibitem[Delatolas et~al.(2024)Delatolas, Kalogeiton, and Papadopoulos]{delatolas2024learning}
Thanos Delatolas, Vicky Kalogeiton, and Dim~P Papadopoulos.
\newblock Learning the what and how of annotation in video object segmentation.
\newblock In \emph{Proceedings of the IEEE/CVF Winter Conference on Applications of Computer Vision}, pages 6951--6961, 2024.

\bibitem[Ravi et~al.(2024)Ravi, Gabeur, Hu, Hu, Ryali, Ma, Khedr, R{\"a}dle, Rolland, Gustafson, et~al.]{ravi2024sam}
Nikhila Ravi, Valentin Gabeur, Yuan-Ting Hu, Ronghang Hu, Chaitanya Ryali, Tengyu Ma, Haitham Khedr, Roman R{\"a}dle, Chloe Rolland, Laura Gustafson, et~al.
\newblock Sam 2: Segment anything in images and videos.
\newblock \emph{arXiv preprint arXiv:2408.00714}, 2024.

\bibitem[Papadopoulos et~al.(2014)Papadopoulos, Clarke, Keller, and Ferrari]{Papadopoulos2014TrainingOC}
Dim~P. Papadopoulos, Alasdair D.~F. Clarke, Frank Keller, and Vittorio Ferrari.
\newblock Training object class detectors from eye tracking data.
\newblock In \emph{European Conference on Computer Vision}, 2014.
\newblock URL \url{https://api.semanticscholar.org/CorpusID:14119147}.

\bibitem[Papadopoulos et~al.(2017)Papadopoulos, Uijlings, Keller, and Ferrari]{papadopoulos2017training}
Dim~P Papadopoulos, Jasper~RR Uijlings, Frank Keller, and Vittorio Ferrari.
\newblock Training object class detectors with click supervision.
\newblock In \emph{Proceedings of the IEEE Conference on Computer Vision and Pattern Recognition}, pages 6374--6383, 2017.

\bibitem[Bearman et~al.(2016)Bearman, Russakovsky, Ferrari, and Fei-Fei]{bearman2016s}
Amy Bearman, Olga Russakovsky, Vittorio Ferrari, and Li~Fei-Fei.
\newblock What’s the point: Semantic segmentation with point supervision.
\newblock In \emph{European conference on computer vision}, pages 549--565. Springer, 2016.

\bibitem[Rother et~al.(2004)Rother, Kolmogorov, and Blake]{grabcut}
Carsten Rother, Vladimir Kolmogorov, and Andrew Blake.
\newblock Grabcut -- interactive foreground extraction using iterated graph cuts.
\newblock \emph{ACM transactions on graphics (TOG)}, 23\penalty0 (3):\penalty0 309--314, 2004.

\bibitem[Martin et~al.(2001)Martin, Fowlkes, Tal, and Malik]{berkeley}
David Martin, Charless Fowlkes, Doron Tal, and Jitendra Malik.
\newblock A database of human segmented natural images and its application to evaluating segmentation algorithms and measuring ecological statistics.
\newblock In \emph{ICCV}, 2001.

\bibitem[Perazzi et~al.(2016)Perazzi, Pont-Tuset, McWilliams, Van~Gool, Gross, and Sorkine-Hornung]{davis}
Federico Perazzi, Jordi Pont-Tuset, Brian McWilliams, Luc Van~Gool, Markus Gross, and Alexander Sorkine-Hornung.
\newblock A benchmark dataset and evaluation methodology for video object segmentation.
\newblock In \emph{CVPR}, 2016.

\bibitem[Lin et~al.(2014)Lin, Maire, Belongie, Hays, Perona, Ramanan, Doll{\'a}r, and Zitnick]{coco}
Tsung-Yi Lin, Michael Maire, Serge Belongie, James Hays, Pietro Perona, Deva Ramanan, Piotr Doll{\'a}r, and C~Lawrence Zitnick.
\newblock Microsoft coco: Common objects in context.
\newblock In \emph{ECCV}, 2014.

\bibitem[Moskalenko et~al.(2024)Moskalenko, Shakhuro, Vorontsova, Konushin, Antonov, Krapukhin, Shepelev, and Soshin]{moskalenko2024tetris}
Andrey Moskalenko, Vlad Shakhuro, Anna Vorontsova, Anton Konushin, Anton Antonov, Alexander Krapukhin, Denis Shepelev, and Konstantin Soshin.
\newblock Tetris: Towards exploring the robustness of interactive segmentation.
\newblock \emph{arXiv preprint arXiv:2402.06132}, 2024.

\bibitem[Zhou et~al.(2023)Zhou, Wang, Zhao, Li, Huang, Meng, and Zheng]{zhou2023interactivegpcis}
Minghao Zhou, Hong Wang, Qian Zhao, Yuexiang Li, Yawen Huang, Deyu Meng, and Yefeng Zheng.
\newblock Interactive segmentation as gaussion process classification.
\newblock In \emph{Proceedings of the IEEE/CVF Conference on Computer Vision and Pattern Recognition}, pages 19488--19497, 2023.

\bibitem[Chen et~al.(2021)Chen, Zhao, Yu, Zhang, and Duan]{chen2021conditional}
Xi~Chen, Zhiyan Zhao, Feiwu Yu, Yilei Zhang, and Manni Duan.
\newblock Conditional diffusion for interactive segmentation.
\newblock In \emph{Proceedings of the IEEE/CVF International Conference on Computer Vision}, pages 7345--7354, 2021.

\bibitem[Lin et~al.(2024)Lin, Chen, Yang, Roitberg, Li, Li, and Li]{lin2024adaptiveclick}
Jiacheng Lin, Jiajun Chen, Kailun Yang, Alina Roitberg, Siyu Li, Zhiyong Li, and Shutao Li.
\newblock Adaptiveclick: Click-aware transformer with adaptive focal loss for interactive image segmentation.
\newblock \emph{IEEE Transactions on Neural Networks and Learning Systems}, 2024.

\bibitem[Liu et~al.(2022)Liu, Xu, Bertasius, and Niethammer]{liu2022simpleclick}
Qin Liu, Zhenlin Xu, Gedas Bertasius, and Marc Niethammer.
\newblock Simpleclick: Interactive image segmentation with simple vision transformers.
\newblock \emph{arXiv preprint arXiv:2210.11006}, 2022.

\bibitem[Sun et~al.(2023)Sun, Xian, Xu, Yao, and Capriotti]{sun2023cfricl}
Shoukun Sun, Min Xian, Fei Xu, Tiankai Yao, and Luca Capriotti.
\newblock Cfr-icl: Cascade-forward refinement with iterative click loss for interactive image segmentation.
\newblock \emph{arXiv preprint arXiv:2303.05620}, 2023.

\bibitem[Zhang et~al.(2023)Zhang, Han, Qiao, Kim, Bae, Lee, and Hong]{zhang2023faster}
Chaoning Zhang, Dongshen Han, Yu~Qiao, Jung~Uk Kim, Sung-Ho Bae, Seungkyu Lee, and Choong~Seon Hong.
\newblock Faster segment anything: Towards lightweight sam for mobile applications.
\newblock \emph{arXiv preprint arXiv:2306.14289}, 2023.

\bibitem[Kirillov et~al.(2023)Kirillov, Mintun, Ravi, Mao, Rolland, Gustafson, Xiao, Whitehead, Berg, Lo, et~al.]{kirillov2023segment}
Alexander Kirillov, Eric Mintun, Nikhila Ravi, Hanzi Mao, Chloe Rolland, Laura Gustafson, Tete Xiao, Spencer Whitehead, Alexander~C Berg, Wan-Yen Lo, et~al.
\newblock Segment anything.
\newblock \emph{arXiv preprint arXiv:2304.02643}, 2023.

\bibitem[Ke et~al.(2023)Ke, Ye, Danelljan, Liu, Tai, Tang, and Yu]{ke2023segment}
Lei Ke, Mingqiao Ye, Martin Danelljan, Yifan Liu, Yu-Wing Tai, Chi-Keung Tang, and Fisher Yu.
\newblock Segment anything in high quality.
\newblock \emph{arXiv preprint arXiv:2306.01567}, 2023.

\bibitem[Gebru et~al.(2021)Gebru, Morgenstern, Vecchione, Vaughan, Wallach, Daumé~III, and Crawford]{datasheet}
Timnit Gebru, Jamie Morgenstern, Briana Vecchione, Jennifer~Wortman Vaughan, Hanna Wallach, Hal Daumé~III, and Kate Crawford.
\newblock Datasheets for datasets.
\newblock volume~64, pages 86--92, 2021.

\end{thebibliography}

\newpage

\appendix

% \title{\TITLEVAL \\[5pt] \large \mdseries Supplementary Materials}

% \settitle

\section{Benchmark Discussion}
\label{sec:bench-info}

{\bfseries Access.} The benchmark is publicly available at \weburl repository.

{\bfseries License.} Clicks are under \href{https://creativecommons.org/licenses/by-nc/4.0/deed.en}{CC BY-NC 4.0}, evaluation code and baseline models are under \href{https://opensource.org/licenses/MIT}{MIT}.

{\bfseries Ethical issues.} The benchmark is created for testing interactive segmentation methods.
To the best of our knowledge, interactive segmentation has two applications: image editing and assisted image labeling.
Since interactive segmentation methods still do require user input, we believe their emergence will not make image labeling and image editing jobs redundant.

{\bfseries Limitations of work.} We observe the following major limitations of our benchmark:
\begin{enumerate}
    \item \label{lim:1} To estimate the sample statistic, we split clicks into 10 pattern groups and sample 1 click from each group per round (10 samples per round).
    While increasing the number of sampled clicks improves performance estimations per image, we believe that, for a sufficiently large dataset, 10 sampled clicks per round is sufficient to accurately estimate average performance over the dataset.
  
    \item \label{lim:2} We trained our \textit{clickability model} on the TETRIS subset.
    The click patterns observed in our study may be domain-specific to the train images. 
    Different types of images (e.g., medical images, satellite images) might result in different click behaviors that our model does not account for.
    \item \label{lim:3} Our dataset of clicks was collected from an online annotation platform, which means our model may reflect biases specific to the individuals who participated in this platform. This could include biases related to their annotation habits, demographic backgrounds, or other unmeasured factors, potentially affecting the model's performance when applied to a wider or different user base.
    \item \label{lim:4} Our model and benchmark are based on click interactions, but interactive segmentation can involve other modalities such as scribbles, contours, or voice commands. We have not considered these types of inputs.
\end{enumerate}

To overcome Limitation~\ref{lim:1}, we would need to evaluate the model for every possible click position, which is computationally impossible.

We believe that Limitation~\ref{lim:2} is inevitable due to the fact that we cannot annotate all possible image scenarios.

Limitation~\ref{lim:3} cannot be avoided because it is not feasible to obtain a fully representative population of all people with different patterns.

To overcome Limitation~\ref{lim:4}, it would be necessary to initiate a separate research project dedicated to studying different input types for interactive segmentation. Unfortunately, this is beyond the scope of this paper.

\newpage

\section{Clickability Model}
\label{sec:training_details}

\subsection{Training and model details}
\label{app:training_details}

In this subsection, we provide \textit{clickability model} architecture description and its training details.

{\bfseries Architecture details.}
As a predictor, we adapted the state-of-the-art SegNeXt-B segmentation network~\cite{guo2022segnext} with the MSCAN-B backbone. We input the original image into the network and concatenate the ground truth mask with the error mask, feeding the resulting tensor as an additional input to the network. This technique is inspired by Conv1S~\citep{ritm}. However, we used three convolutional layers with non-linear activation functions instead of one, as we needed to encode more complex features of the ground truth mask and error mask, rather than just clicks. Additionally, after a forward pass, we used min-max normalization to transform values to the interval from 0 to 1.

To compare our model's complexity with state-of-the-art interactive segmentation methods, we measured inference speed and resource usage of the state-of-the-art segmentation methods and our clickability model in Table~\ref{tab:models:comparison}. All evaluations were done on a single A100.
The \textit{clickability model} is about twice as fast as RITM and SimpleClick, and about five times faster than SAM. It also uses a small or comparable amount of GPU memory. Note that our model can easily fit on a consumer GPU with 8GB of VRAM.

{\bfseries Training details.} We trained our model for 20 epochs on the train part of TETRIS dataset (TETRIS splits can be found in \texttt{txt}-file of the benchmark code).
During training, samples were augmented using Horizontal Flip.
We minimized Kullback-Leibler Divergence Loss by Adam optimizer with CosineLRScheduler and initial 0.01 learning rate.
The training process took 3 hours on a single Nvidia Tesla A100 GPU.

\begin{table}[htbp]
    \caption{Comparison of the characteristics of different interactive segmentation methods with the clickability model.}
    \label{tab:models:comparison}
    \fontsize{10pt}{14pt}\selectfont
    \tabcolsep=6pt
    \centering
    \begin{tabular}{lccccc}
    \toprule
    Model & Params, M & GFLOPs & Mem, Gb & Inference, ms & Input size \\
    \midrule
    RITM HRNet-32 & 31.95 & 83 & 0.217 & 42 ± 2.0 & 400 × 400 \\
    SimpleClick-ViT-H & 659.39 & 1461 & 2.785 & 41 ± 2.3 & 448 × 448 \\
    SAM-ViT-H & 641.09 & 5473 & 5.745 & 150 ± 2.7 & 1024 × 1024 \\
    Clickability model & 27.59 & 64 & 0.378 & 23 ± 0.9 & 416 × 416 \\
    \bottomrule
    \end{tabular}
\end{table}

\FloatBarrier

\subsection{Model generalizability}
\label{app:model_generalizability}

\begin{table}[ht!]
    \captionof{table}{Evaluation of various \textit{clickability models} on real-user clicks. Our model was trained on the train part of TETRIS dataset. Our approach outperforms baseline clicking strategies in terms of the proximity of samples to real-user clicks on all considered interactive segmentation datasets. 
    }
    \label{tab:models:all_datasets}
    \fontsize{10pt}{14pt}\selectfont
    \tabcolsep=3pt
    \centering
    \begin{tabular}{ccccccc}
    \toprule
    Dataset & Model & KS\,$\uparrow$ & PL$_1$\,$\downarrow$ & WD\,$\downarrow$ & NSS\,$\uparrow$ & PDE\,$\uparrow$ \\
    \midrule
    \multirow{5}{*}{GrabCut} & UD & 0.15 & 0.55 & 0.16 & 2.44 & 4.06E-05 \\
     & DT & 0.22 & 0.49 & 0.14 & 3.94 & 7.75E-05 \\
     & SI & 0.10 & 0.52 & 0.17 & 2.27 & 4.64E-05 \\
     & SM & 0.14 & 0.52 & 0.16 & 2.75 & 5.30E-05 \\
     & Ours & \textbf{0.50} & \textbf{0.40} & \textbf{0.10} & \textbf{5.63} & \textbf{2.12E-04} \\
     \midrule
    \multirow{5}{*}{Berkeley} & UD & 0.16 & 0.54 & 0.16 & 3.39 & 1.20E-04 \\
     & DT & 0.23 & 0.48 & 0.14 & 5.28 & 2.13E-04 \\
     & SI & 0.11 & 0.51 & 0.17 & 3.08 & 1.38E-04 \\
     & SM & 0.17 & 0.51 & 0.15 & 3.69 & 1.50E-04 \\
     & Ours & \textbf{0.50} & \textbf{0.39} & \textbf{0.09} & \textbf{7.29} & \textbf{4.99E-04} \\
     \midrule
    \multirow{5}{*}{DAVIS} & UD & 0.15 & 0.56 & 0.17 & 4.31 & 6.49E-05 \\
     & DT & 0.25 & 0.49 & 0.14 & 6.68 & 1.21E-04 \\
     & SI & 0.11 & 0.53 & 0.17 & 3.96 & 7.22E-05 \\
     & SM & 0.16 & 0.53 & 0.16 & 4.50 & 7.84E-05 \\
     & Ours & \textbf{0.48} & \textbf{0.41} & \textbf{0.10} & \textbf{8.55} & \textbf{2.53E-04} \\
     \midrule
    \multirow{5}{*}{COCO-MVal} & UD & 0.20 & 0.79 & 0.21 & 6.08 & 4.04E-04 \\
     & DT & 0.26 & 0.73 & 0.19 & 8.34 & 5.74E-04 \\
     & SI & 0.16 & 0.73 & 0.22 & 6.26 & 5.12E-04 \\
     & SM & 0.21 & 0.74 & 0.20 & 6.80 & 5.14E-04 \\
     & Ours & \textbf{0.50} & \textbf{0.61} & \textbf{0.14} & \textbf{10.44} & \textbf{7.29E-04} \\
     \midrule
    \multirow{5}{*}{TETRIS (Val)} & UD & 0.10 & 0.57 & 0.17 & 3.99 & 1.36E-05 \\
     & DT & 0.14 & 0.52 & 0.16 & 6.45 & 2.76E-05 \\
     & SI & 0.09 & 0.51 & 0.16 & 4.13 & 1.66E-05 \\
     & SM & 0.13 & 0.51 & 0.15 & 4.79 & 1.83E-05 \\
     & Ours & \textbf{0.55} & \textbf{0.40} & \textbf{0.08} & \textbf{9.11} & \textbf{4.69E-05} \\
    \bottomrule
    \end{tabular}
\end{table}

To assess the generalizability of our \textit{clickability model}, we computed metrics on click samples for different datasets in Table~\ref{tab:models:all_datasets}.

Note, that in Tables~\ref{tab:models:eval},~\ref{tab:models:all_datasets} and~\ref{tab:models:ablation} KS, PL$_1$, WD were calculated using clicks bootstrapping from \textit{clickability model} (100 times per instance).

Additionally, to calculate saliency fairly, we assume the model should receive not the full image, but only a cropped region of the image with the object of interest.
This ensures that the model's attention is directed towards the relevant area, leading to a more accurate assessment of its saliency.
We have checked two variants of giving image with object of interest: \begin{enumerate}[label=(\arabic*)]
    \item Part of image with expand around the object of interest on 1.4 times (denoted as SI in Table~\ref{tab:models:all_datasets}).
    \item Part of image with expand around the object of interest on 1.4 times, but everything except the object of interest is gray (denoted as SM in Table~\ref{tab:models:all_datasets}).
\end{enumerate}

According to Table~\ref{tab:models:all_datasets}, our clickability model significantly outperforms all baselines on TETRIS and other datasets.
On other datasets the click sampling quality of our model mostly is slightly worse than on TETRIS, nevertheless we find these results comparable.
Therefore, we conclude that the clickability model is not limited to TETRIS dataset and generalizes well to other datasets of common images.

However, datasets from specific domains, e.g. medical, are out of the scope of our work.
For these datasets, additional data collection and evaluations should be done.

\FloatBarrier

\subsection{Clickability maps ablations}
\label{app:click_maps_ablation}

\begin{figure}[!ht]
    \centering
    \begin{subfigure}[t]{0.15\textwidth}
        \centering
        \includegraphics[width=\textwidth]{images/click_models/224_clicks.png}
        \caption{Clicks}
        \label{fig:click_maps:fixations}
    \end{subfigure}
    \begin{subfigure}[t]{0.15\textwidth}
        \centering
        \includegraphics[width=\textwidth]{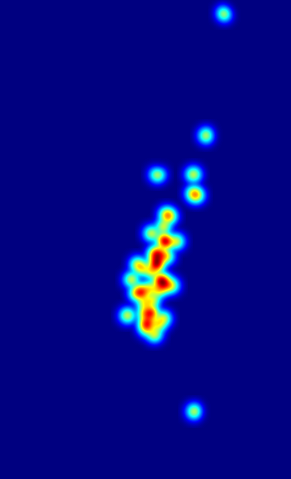}
        \caption{CM$_{10}$}
        \label{fig:click_maps:click_map_10}
    \end{subfigure}
    \begin{subfigure}[t]{0.15\textwidth}
        \centering
        \includegraphics[width=\textwidth]{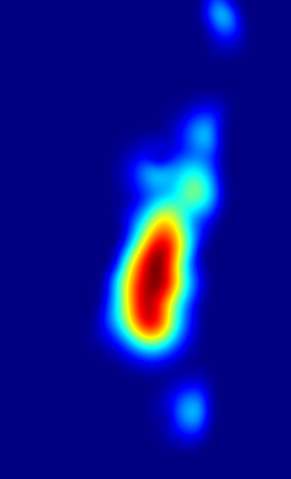}
        \caption{CM$_{30}$}
        \label{fig:click_maps:click_map_30}
    \end{subfigure}
    \begin{subfigure}[t]{0.15\textwidth}
        \centering
        \includegraphics[width=\textwidth]{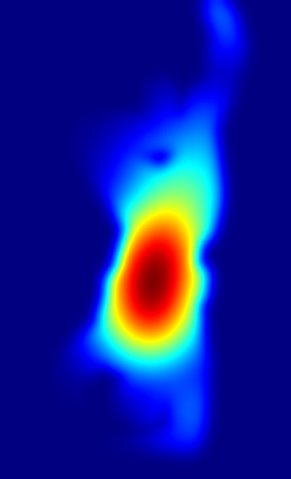}
        \caption{CM$_{60}$}
        \label{fig:click_maps:click_map_60}
    \end{subfigure}
    \begin{subfigure}[t]{0.15\textwidth}
        \centering
        \includegraphics[width=\textwidth]{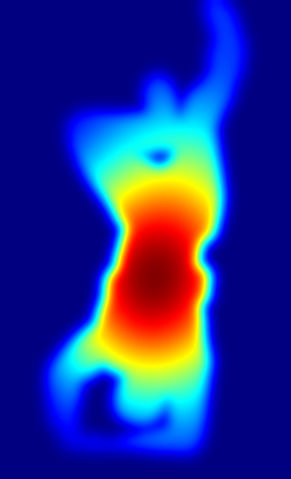}
        \caption{CM$_{120}$}
        \label{fig:click_maps:click_map_120}
    \end{subfigure} 
    \caption{Example of ground-turth clickability maps \subref{fig:click_maps:click_map_10} -- \subref{fig:click_maps:click_map_120} with various $\sigma$ (denoted CM$_{\sigma}$) constructed from ground-truth clicks and object mask \subref{fig:click_maps:fixations} used to train clickability model.
    }
    \label{fig:click_maps}
\end{figure}

Figure~\ref{fig:click_maps} illustrates examples of clickability maps with various sigmas (denoted CM$_\sigma$).
For each sigma, we trained a separate model and chose the best one according to our ablations on a validation set of clicks and images, Table~\ref{tab:models:ablation} contains evaluation results for 
our \textit{clickability models} with various sigmas.
Since CM$_{5}$ performed best on almost all metrics, we use it in our interactive segmentation benchmark.

\begin{table}[h]
    \captionof{table}{Evaluation of our \textit{clickability model} when training on ground-turth clickability maps with various $\sigma$ (denoted CM$_{\sigma}$) on validation part of TETRIS.
    }
    \label{tab:models:ablation}
    \fontsize{10pt}{14pt}\selectfont
    \tabcolsep=3pt
    \centering
    \begin{tabular}{cccccc}
    \toprule
    Train data  & KS\,$\uparrow$ & PL$_1$\,$\downarrow$ & WD\,$\downarrow$ & NSS\,$\uparrow$ & PDE\,$\uparrow$ \\
    \midrule
CM$_{0.01}$ & 0.48                          & 0.408                               & 0.09                            & 8.61                           & 4.44E-05                       \\
CM$_{0.25}$ & 0.46                          & 0.416                               & 0.10                            & 8.58                           & 4.40E-05                       \\
CM$_{0.5}$  & 0.49                          & 0.411                               & 0.09                            & 8.87                           & 4.70E-05                       \\
CM$_{1}$    & 0.51                          & 0.407                               & 0.09                            & 8.94                           & \textbf{4.86E-05}              \\
CM$_{2}$    & 0.53                          & 0.399                               & 0.09                            & 9.05                           & 4.83E-05                       \\
CM$_{5}$    & \textbf{0.55}                 & \textbf{0.397}                      & \textbf{0.08}                   & \textbf{9.11}                  & 4.69E-05                       \\
CM$_{10}$   & 0.51                          & 0.403                               & 0.09                            & 8.87                           & 4.17E-05                       \\
CM$_{20}$   & 0.51                          & 0.403                               & 0.09                            & 8.24                           & 3.11E-05                       \\
CM$_{30}$   & 0.48                          & 0.404                               & 0.09                            & 7.71                           & 2.48E-05                       \\
CM$_{60}$   & 0.37                          & 0.424                               & 0.10                            & 6.60                           & 1.67E-05                       \\
CM$_{90}$   & 0.31                          & 0.439                               & 0.10                            & 6.06                           & 1.40E-05                       \\
CM$_{120}$  & 0.26                          & 0.453                               & 0.11                            & 5.69                           & 1.26E-05  \\
    \bottomrule
    \end{tabular}
\end{table}

\FloatBarrier

\subsection{Examples of generated clicks}

\begin{figure}[!ht]
    \centering
    \begin{subfigure}[t]{0.45\textwidth}
        \centering
        \caption{GrabCut, KS-test -- True}
        \includegraphics[width=1\textwidth] {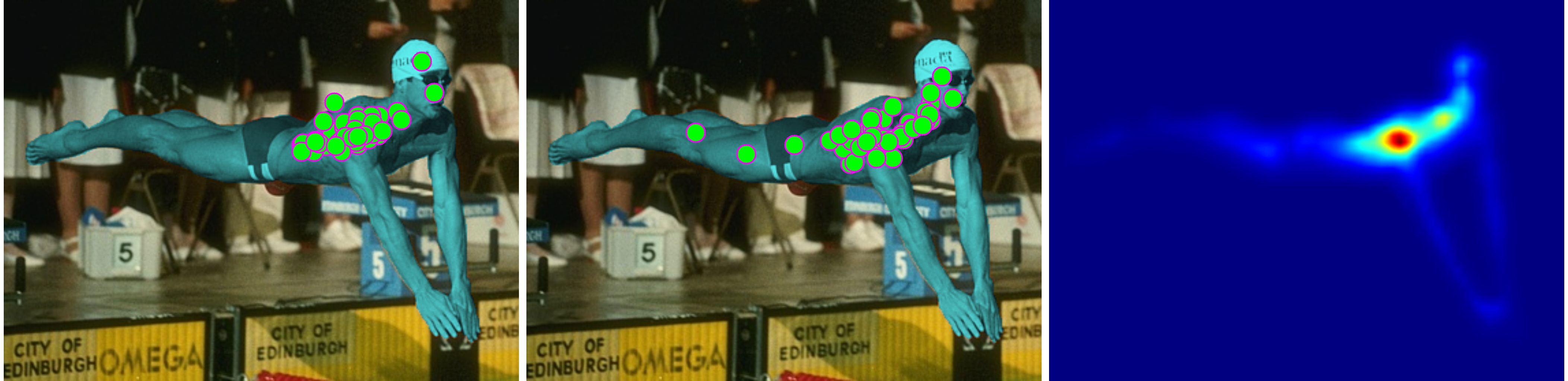}
    \end{subfigure}
    \begin{subfigure}[t]{0.45\textwidth}
        \centering
        \caption{GrabCut, KS-test -- False}        
        \includegraphics[width=1\textwidth]{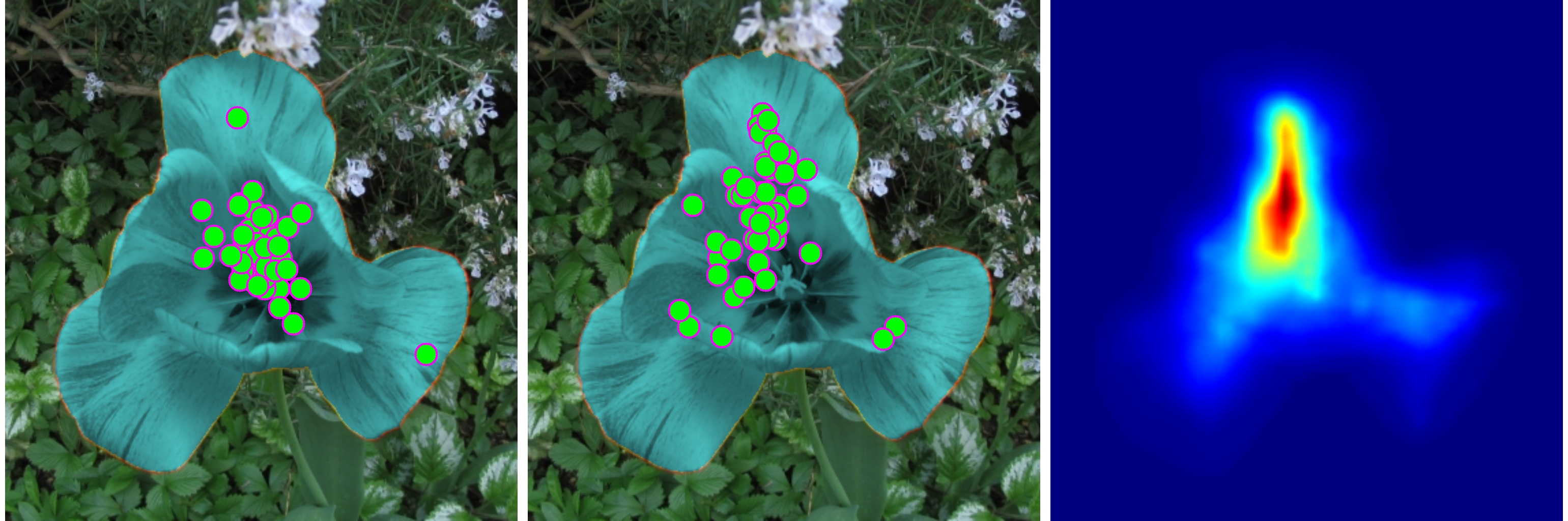}
    \end{subfigure}

    \begin{subfigure}[t]{0.45\textwidth}
        \centering
        \caption{Berkeley, KS-test -- True}
        \includegraphics[width=1\textwidth]{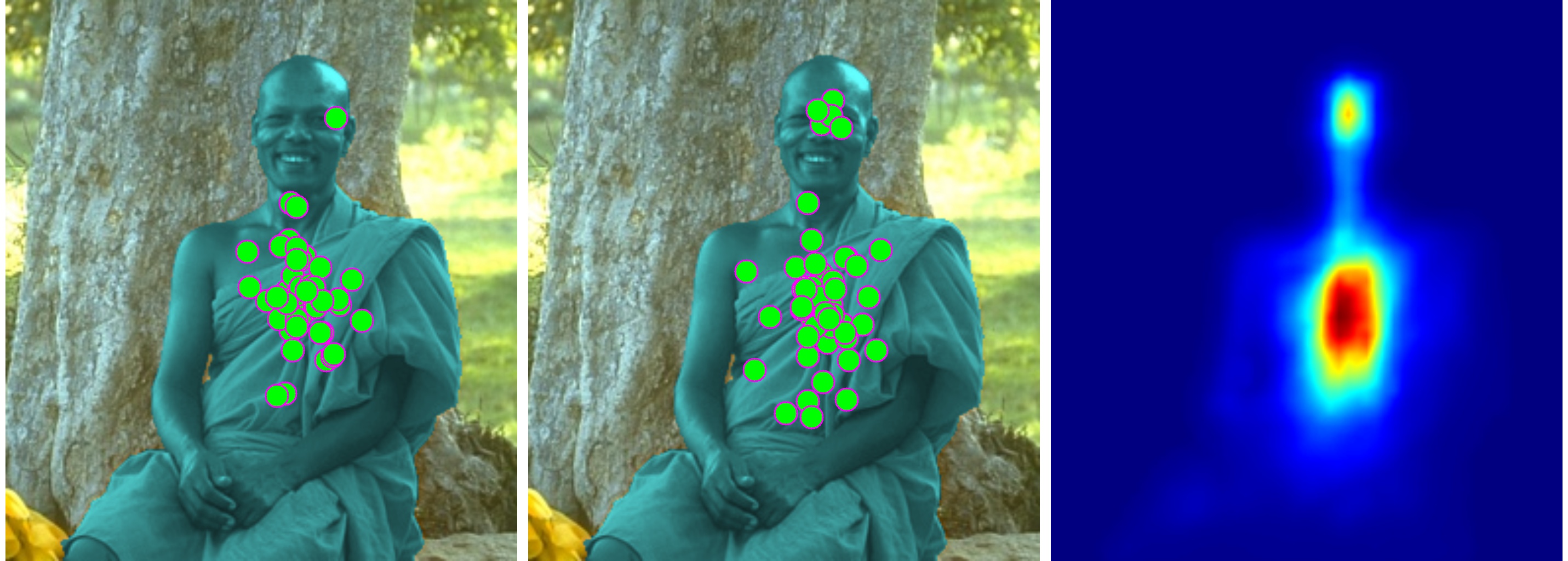}
    \end{subfigure}
    \begin{subfigure}[t]{0.45\textwidth}
        \centering
        \caption{Berkeley, KS-test -- False}        
        \includegraphics[width=1\textwidth]{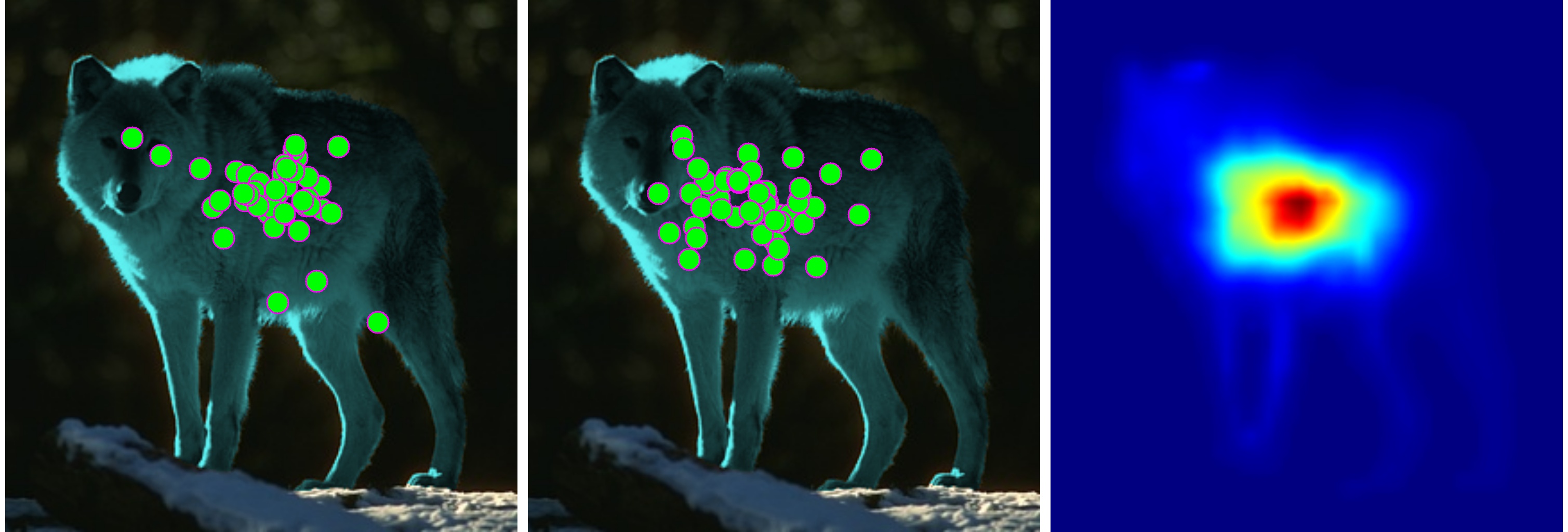}
    \end{subfigure}

    \begin{subfigure}[t]{0.45\textwidth}
        \centering
        \caption{DAVIS, KS-test -- True}
        \includegraphics[width=1\textwidth]{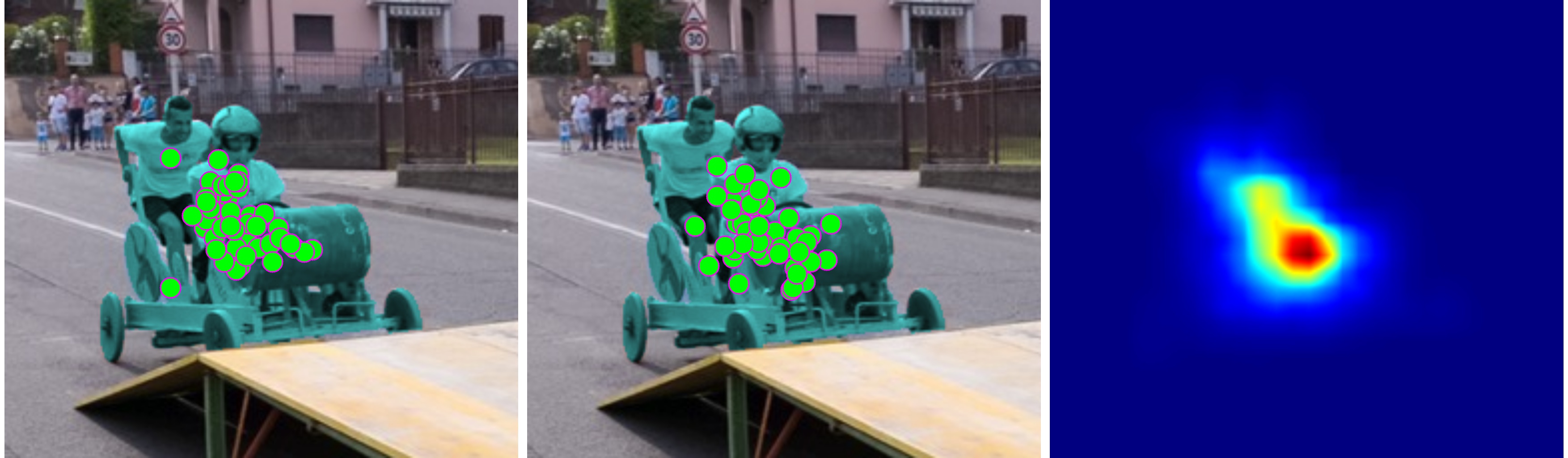}
    \end{subfigure}
    \begin{subfigure}[t]{0.45\textwidth}
        \centering
        \caption{DAVIS, KS-test -- False}        
        \includegraphics[width=1\textwidth]{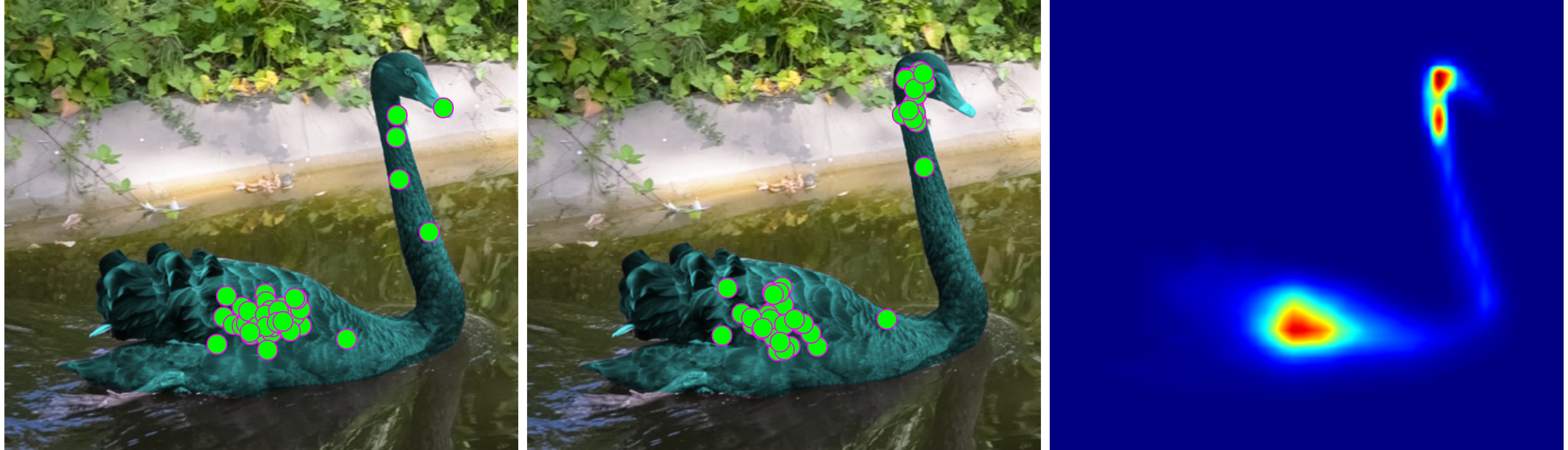}
    \end{subfigure}

    \begin{subfigure}[t]{0.45\textwidth}
        \centering
        \caption{COCO-MVal, KS-test -- True}
        \includegraphics[width=1\textwidth]{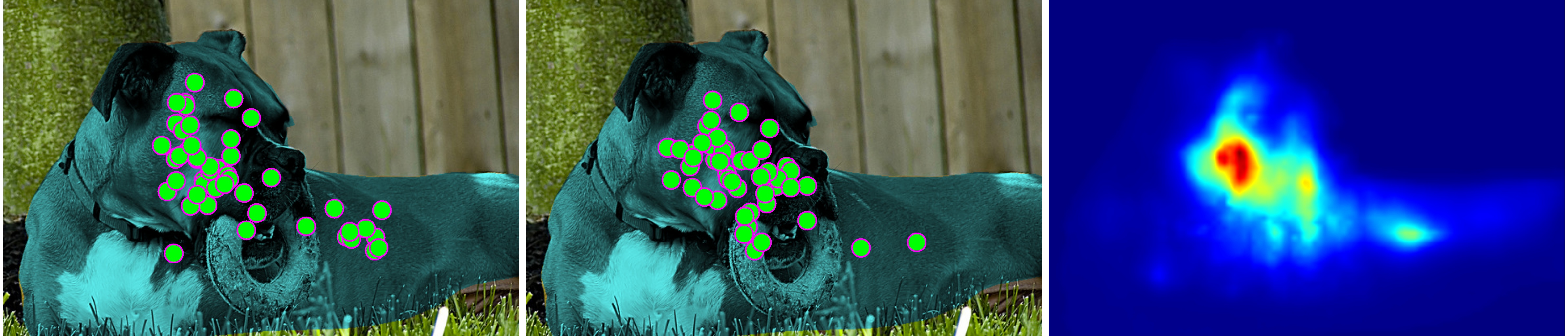}
    \end{subfigure}
    \begin{subfigure}[t]{0.45\textwidth}
        \centering
        \caption{COCO-MVal, KS-test -- False}        
        \includegraphics[width=1\textwidth]{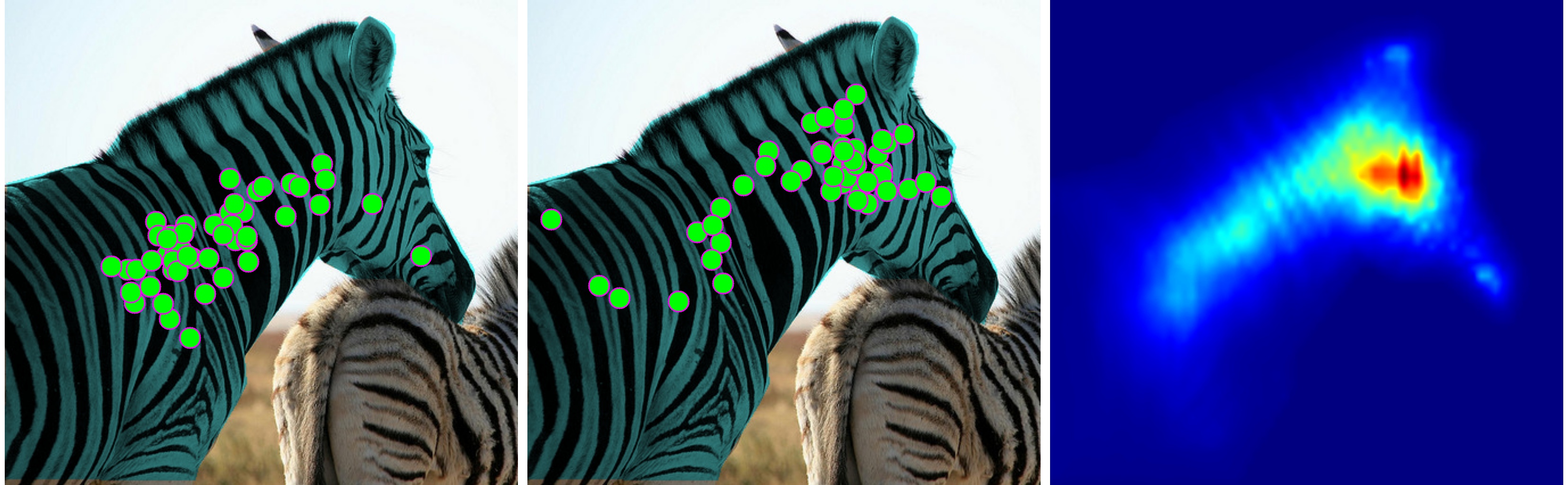}
    \end{subfigure}

    \begin{subfigure}[t]{0.45\textwidth}
        \centering
        \caption{TETRIS, KS-test -- True}
        \includegraphics[width=1\textwidth]{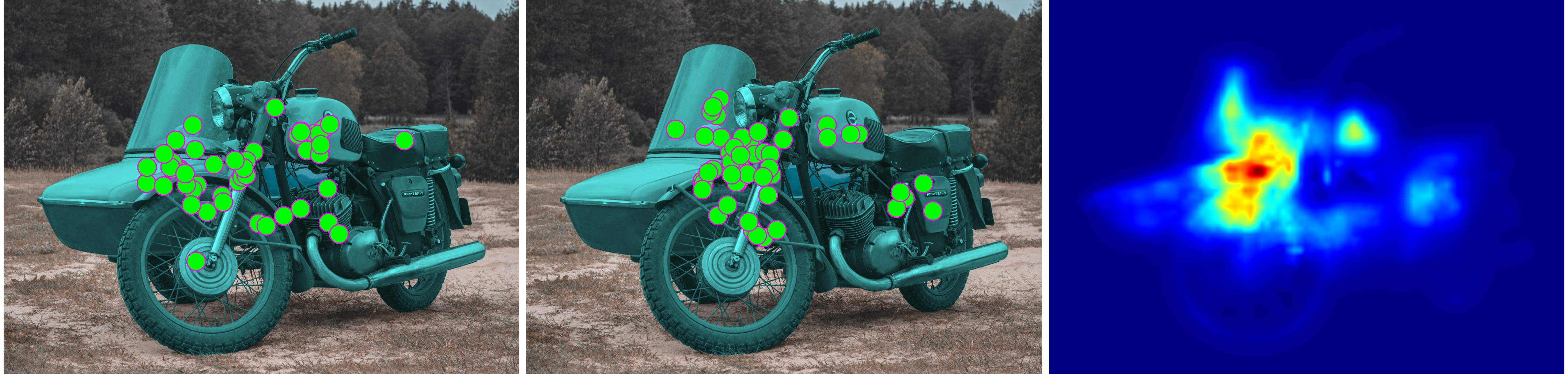}
    \end{subfigure}
    \begin{subfigure}[t]{0.45\textwidth}
        \centering
        \caption{TETRIS, KS-test -- False}        
        \includegraphics[width=1\textwidth]{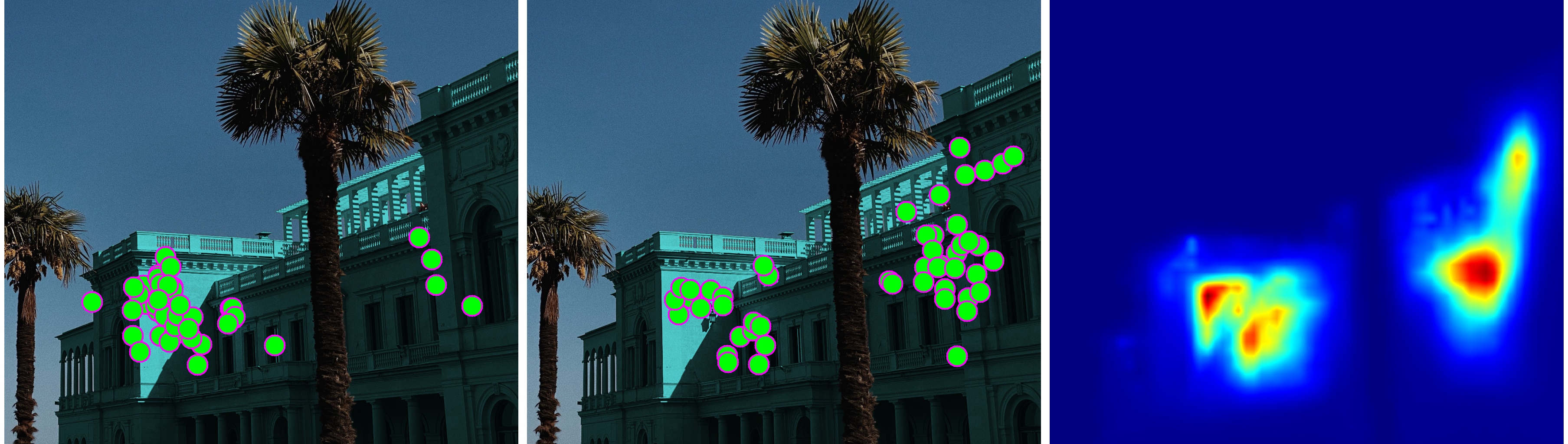}
    \end{subfigure}
    
    \caption{Examples of ground truth clicks (left), clicks (middle) and \textit{clickability map} (right), generated by our \textit{clickability model} for the first round.
    Green dots illustrate first round clicks, teal masks represent target regions that should be segmented.
    In the subcaptions the results of Kolmogorov-Smirnov test (True if p-value > 0.05, i.e. there are no significant differences between the distributions of clicks) are provided.
    }
    \label{fig:examples:first_round}
\end{figure}

\begin{figure}[!ht]
    \centering
    \begin{subfigure}[t]{0.45\textwidth}
        \centering
        \caption{GrabCut, KS-test -- True}
        \includegraphics[width=1\textwidth] {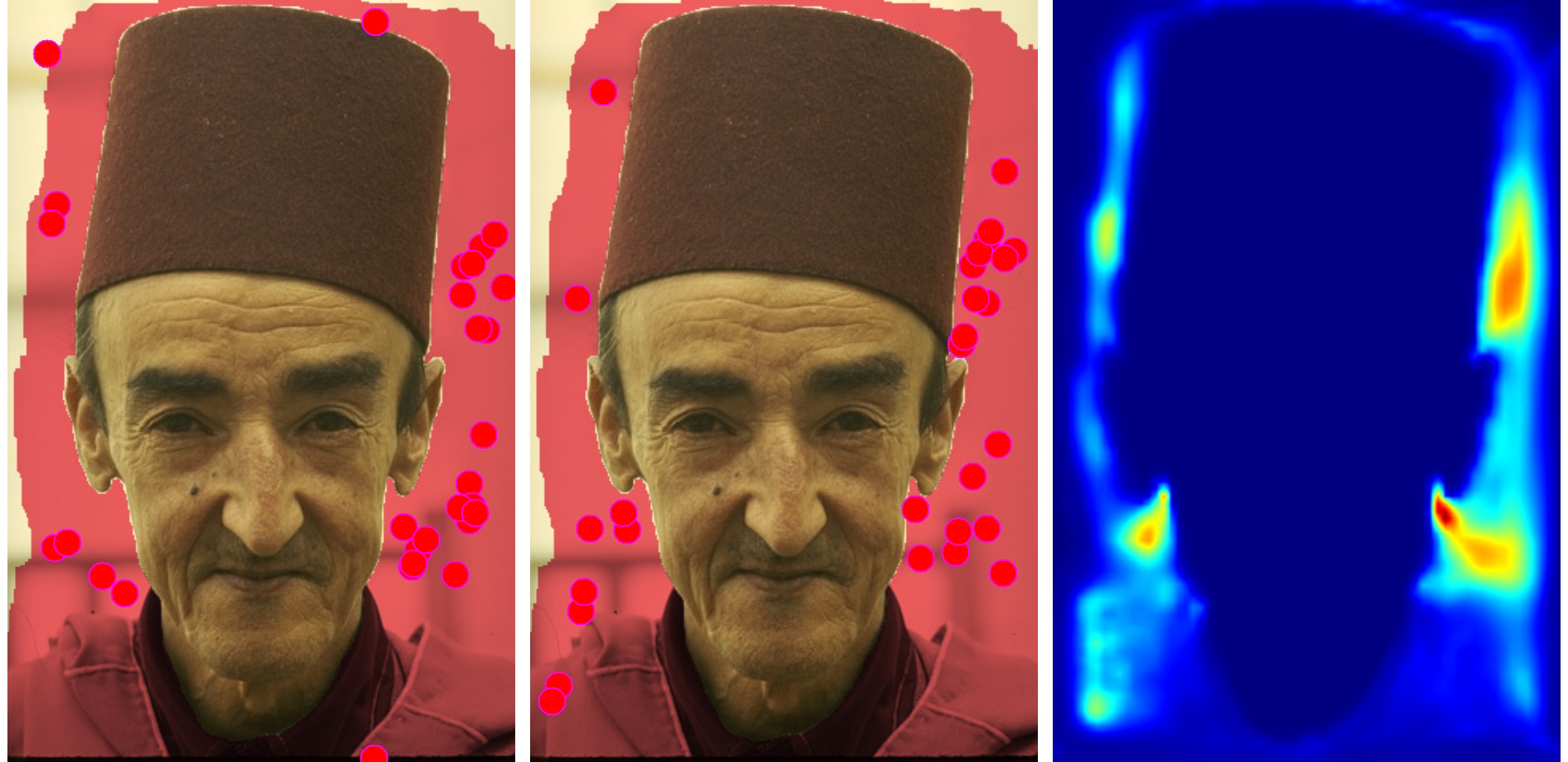}
    \end{subfigure}
    \begin{subfigure}[t]{0.45\textwidth}
        \centering
        \caption{GrabCut, KS-test -- False}        
        \includegraphics[width=1\textwidth]{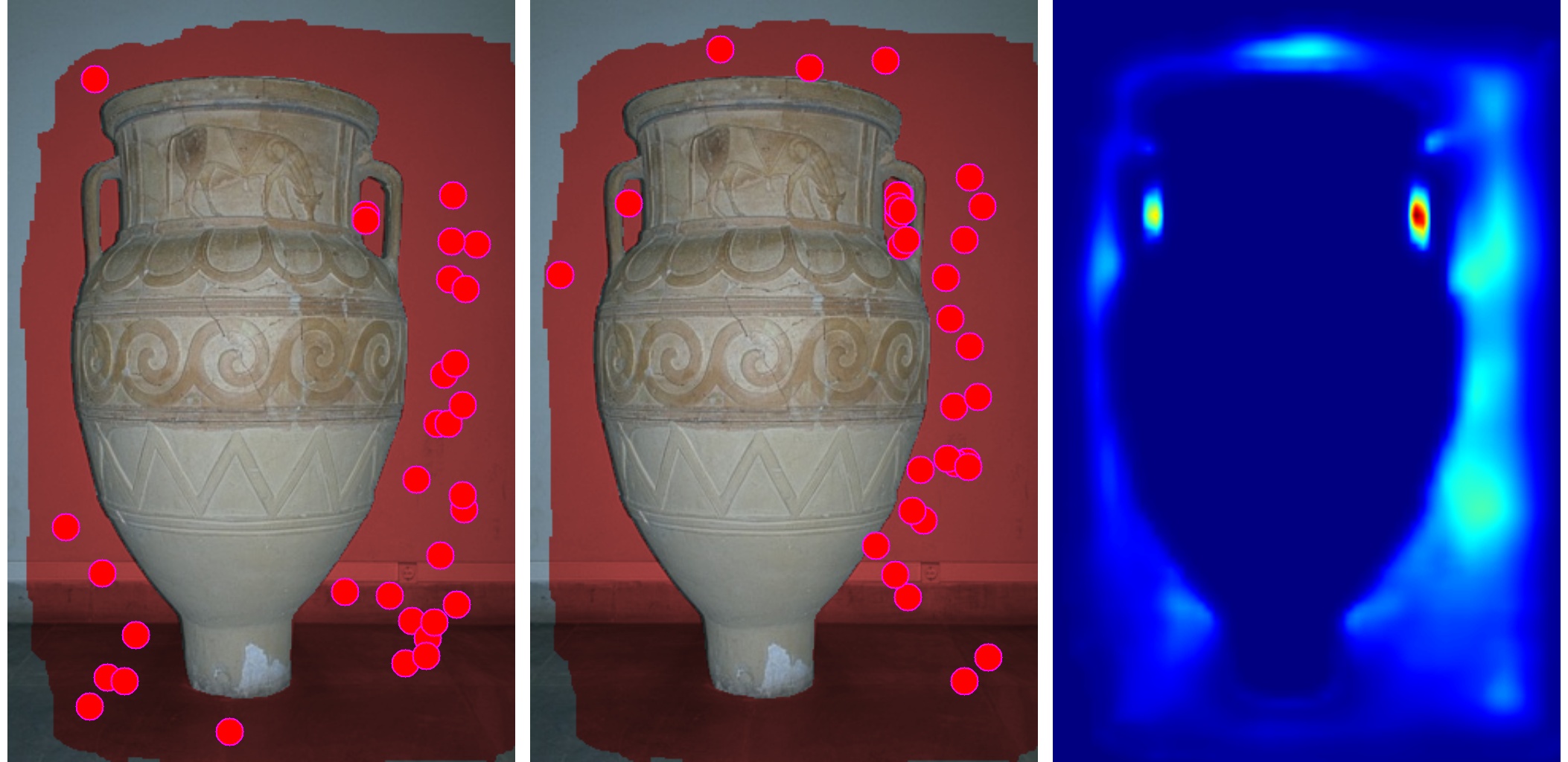}
    \end{subfigure}

    \begin{subfigure}[t]{0.45\textwidth}
        \centering
        \caption{Berkeley, KS-test -- True}
        \includegraphics[width=1\textwidth]{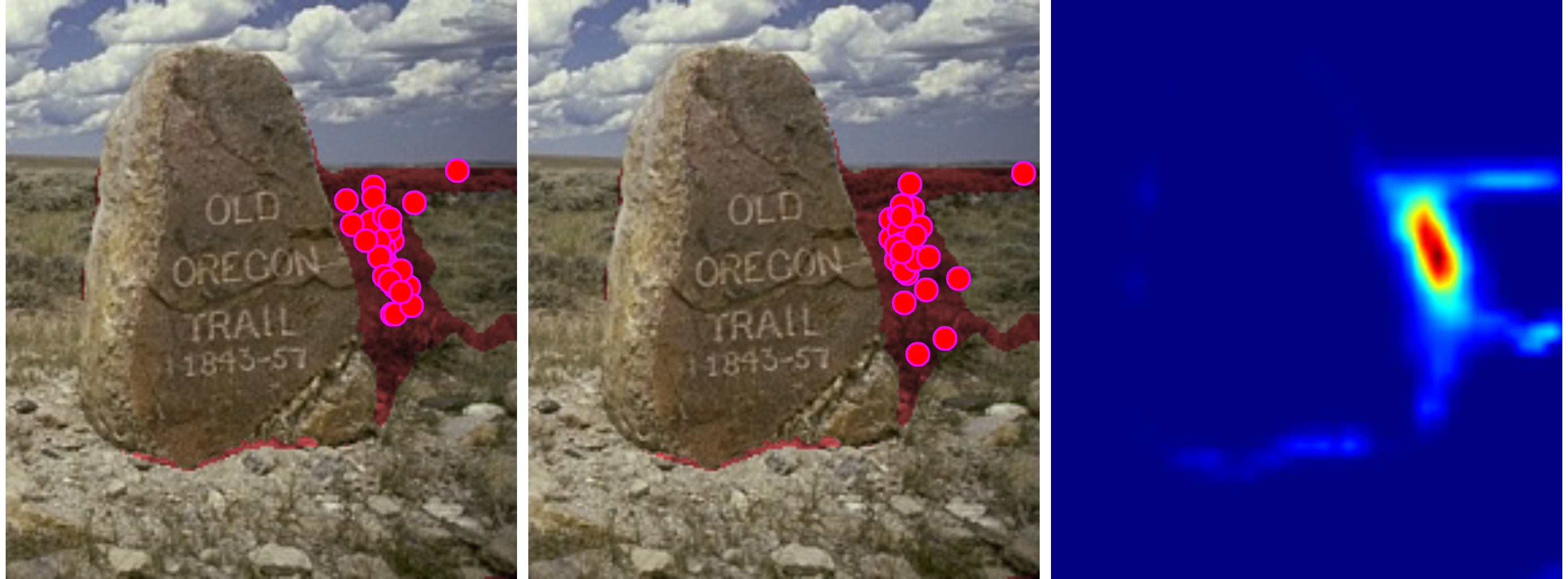}
    \end{subfigure}
    \begin{subfigure}[t]{0.45\textwidth}
        \centering
        \caption{Berkeley, KS-test -- False}        
        \includegraphics[width=1\textwidth]{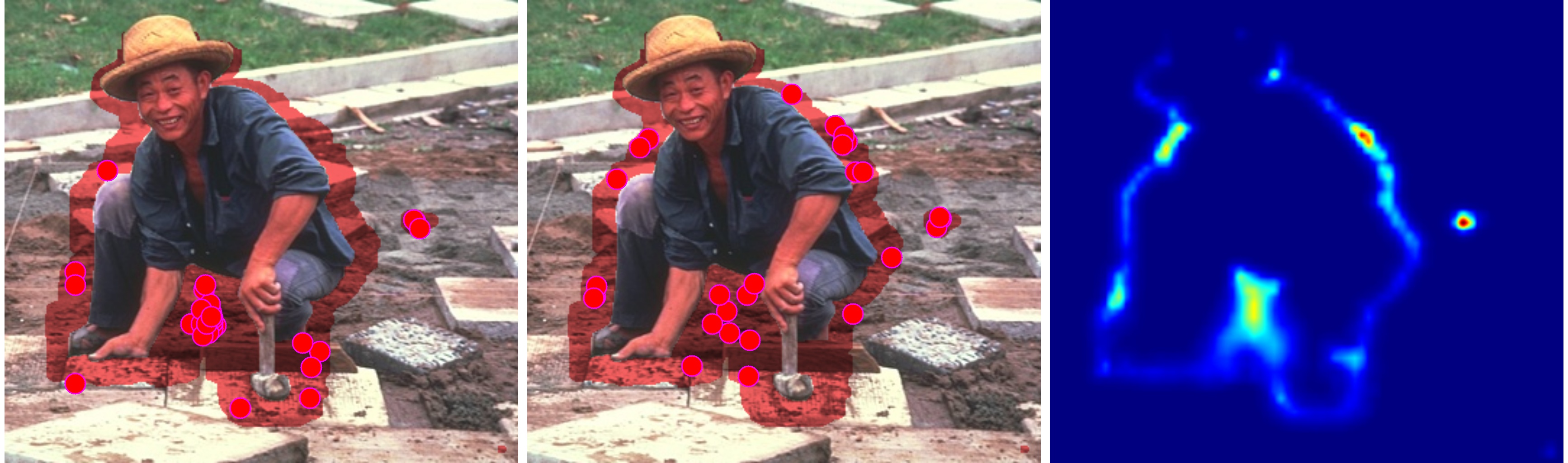}
    \end{subfigure}

    \begin{subfigure}[t]{0.45\textwidth}
        \centering
        \caption{DAVIS, KS-test -- True}
        \includegraphics[width=1\textwidth]{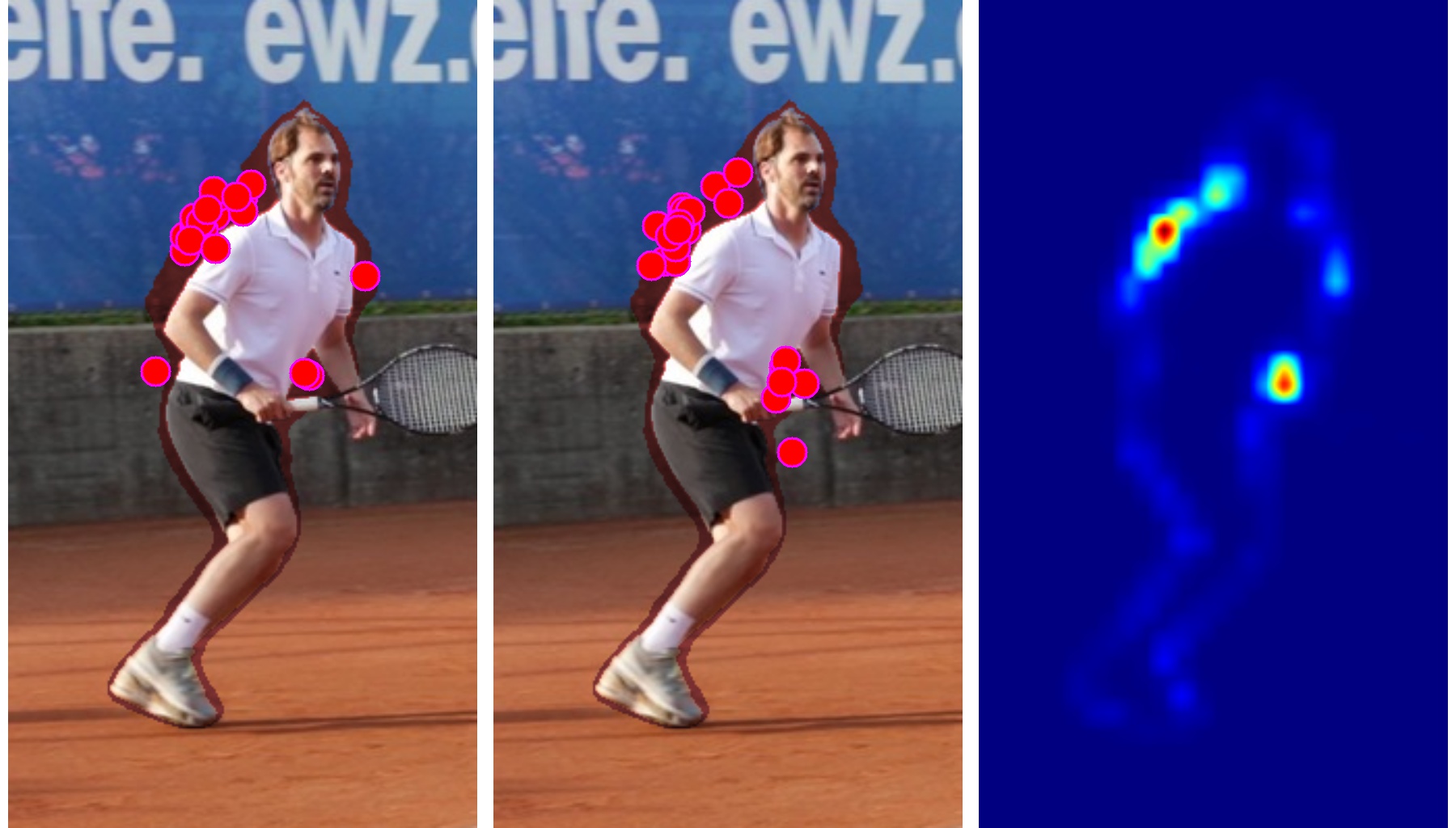}
    \end{subfigure}
    \begin{subfigure}[t]{0.45\textwidth}
        \centering
        \caption{DAVIS, KS-test -- False}        
        \includegraphics[width=1\textwidth]{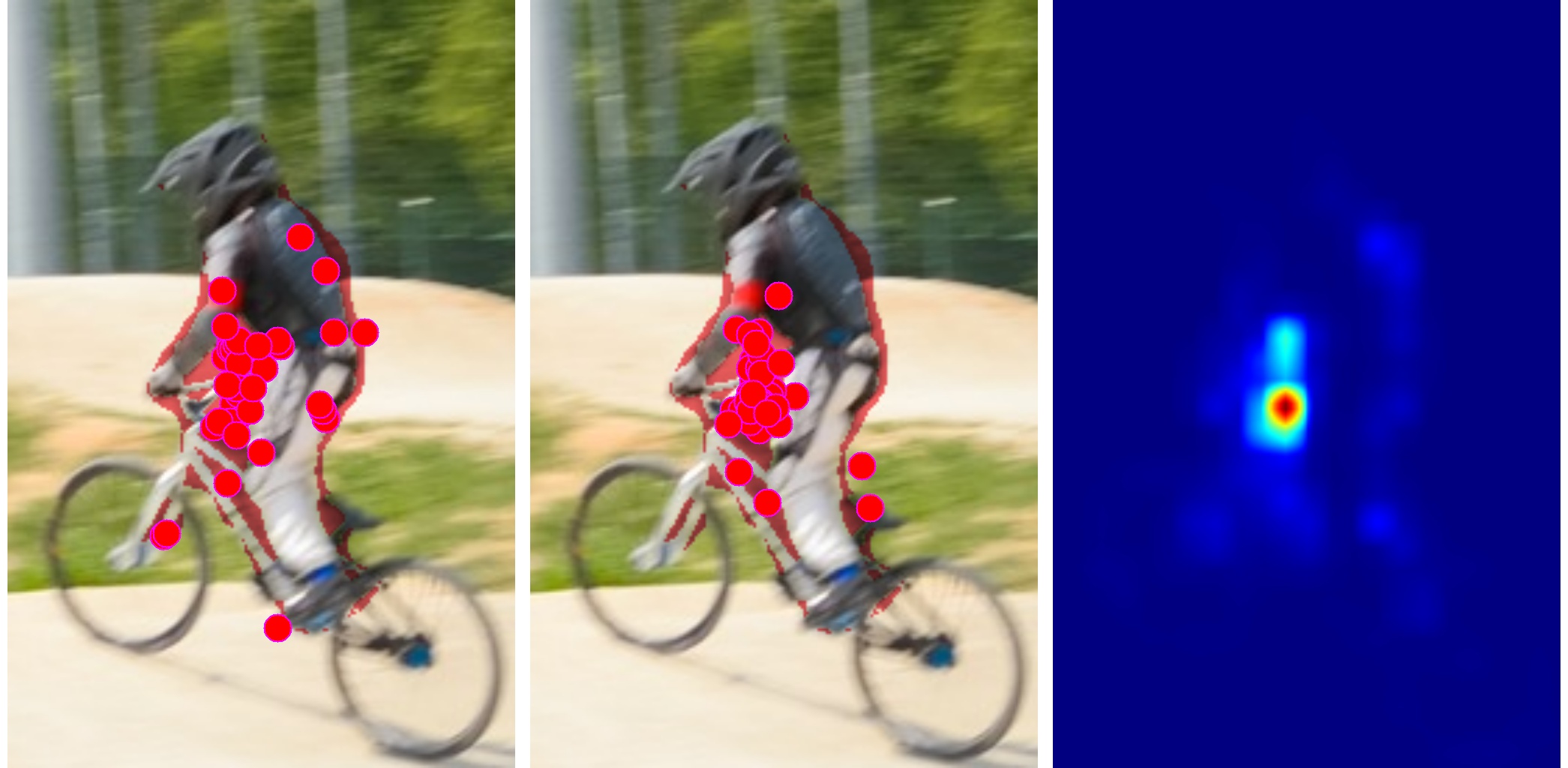}
    \end{subfigure}

    \begin{subfigure}[t]{0.45\textwidth}
        \centering
        \caption{COCO-MVal, KS-test -- True}
        \includegraphics[width=1\textwidth]{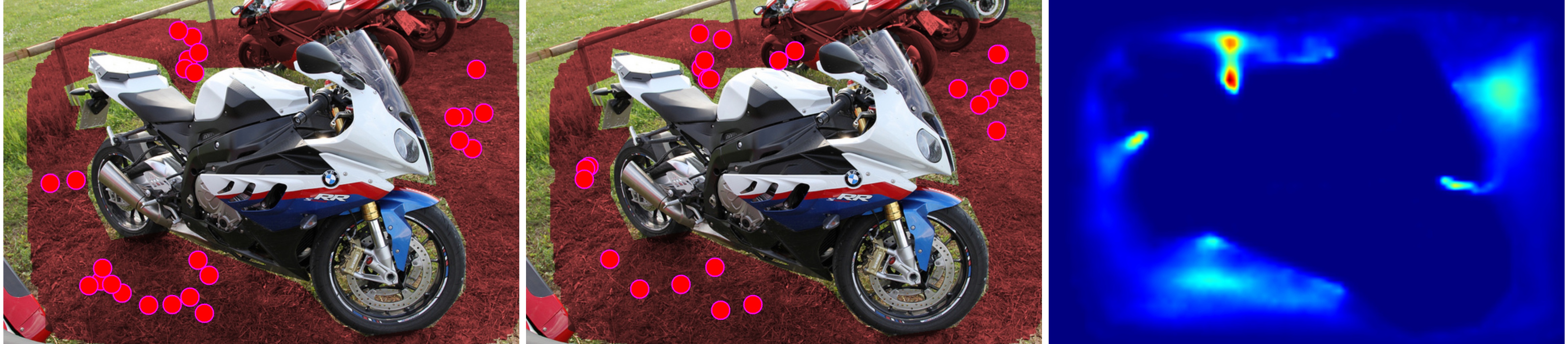}
    \end{subfigure}
    \begin{subfigure}[t]{0.45\textwidth}
        \centering
        \caption{COCO-MVal, KS-test -- False}        
        \includegraphics[width=1\textwidth]{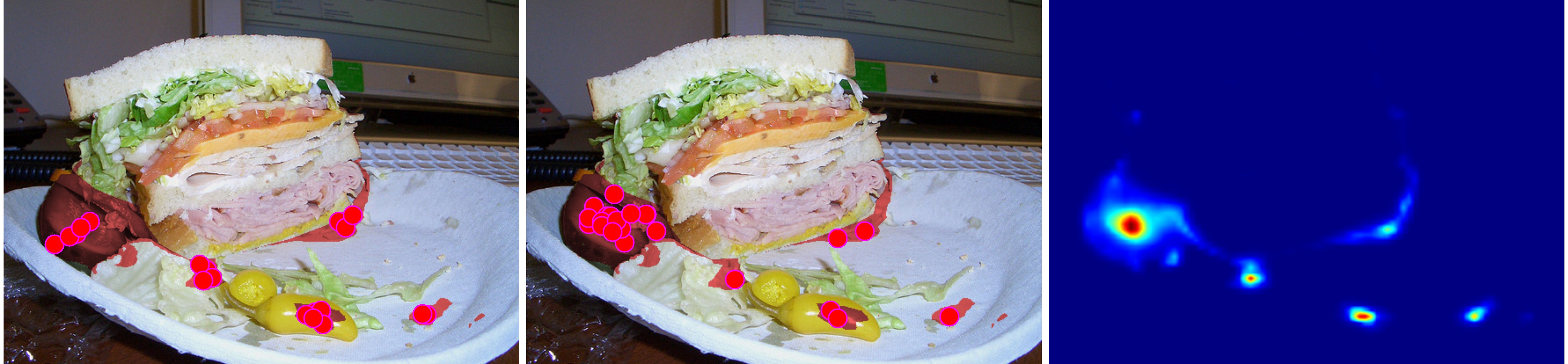}
    \end{subfigure}

    \begin{subfigure}[t]{0.45\textwidth}
        \centering
        \caption{TETRIS, KS-test -- True}
        \includegraphics[width=1\textwidth]{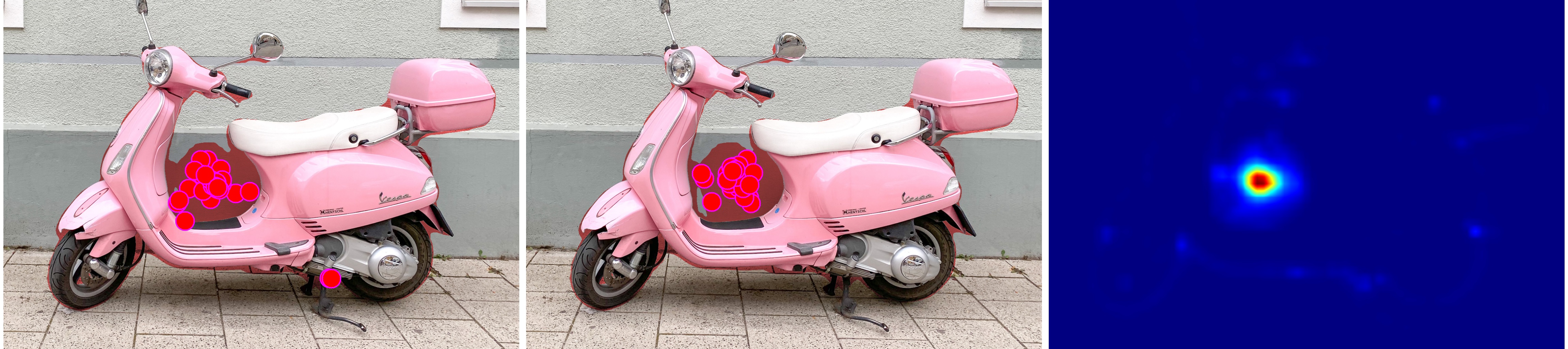}
    \end{subfigure}
    \begin{subfigure}[t]{0.45\textwidth}
        \centering
        \caption{TETRIS, KS-test -- False}        
        \includegraphics[width=1\textwidth]{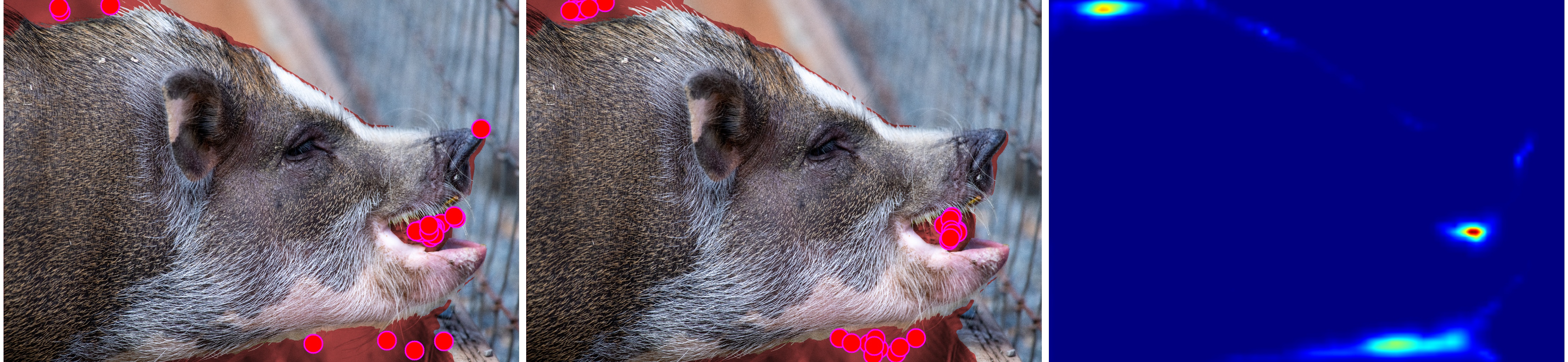}
    \end{subfigure}
    
    \caption{Examples of false-positive (FP) ground truth clicks (left), generated clicks (middle) and predicted \textit{clickability map} (right) by our \textit{clickability model} for the subsequent round. 
    Red dots illustrate FP clicks, red masks represent regions of segmentation errors.
    In the subcaptions the results of Kolmogorov-Smirnov test (True if p-value > 0.05, i.e. there are no significant differences between the distributions of clicks) are provided.
    }
    \label{fig:examples:fp}
\end{figure}

\begin{figure}[!ht]
    \centering
    \begin{subfigure}[t]{0.45\textwidth}
        \centering
        \caption{GrabCut, KS-test -- True}
        \includegraphics[width=1\textwidth] {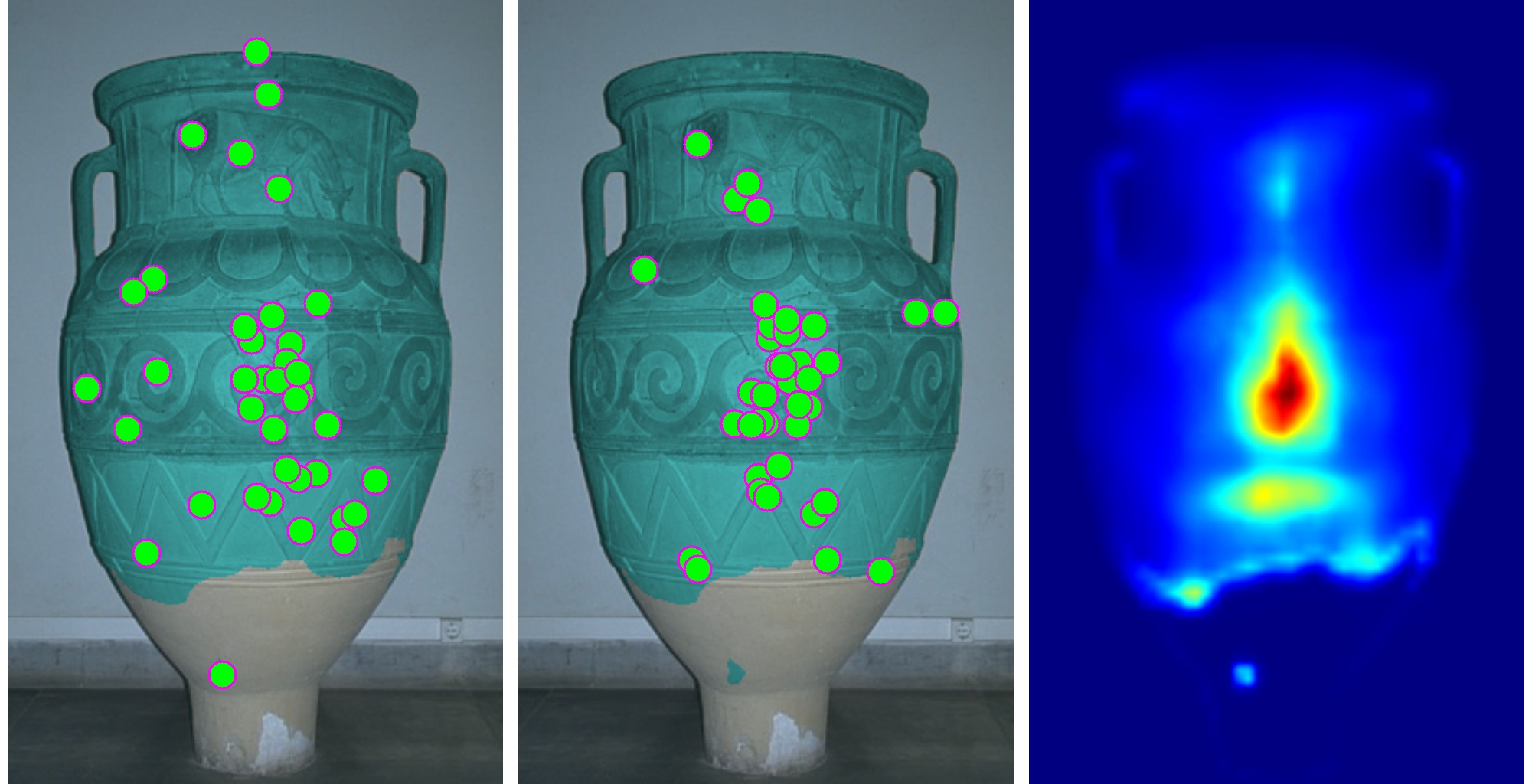}
    \end{subfigure}
    \begin{subfigure}[t]{0.45\textwidth}
        \centering
        \caption{GrabCut, KS-test -- False}        
        \includegraphics[width=1\textwidth]{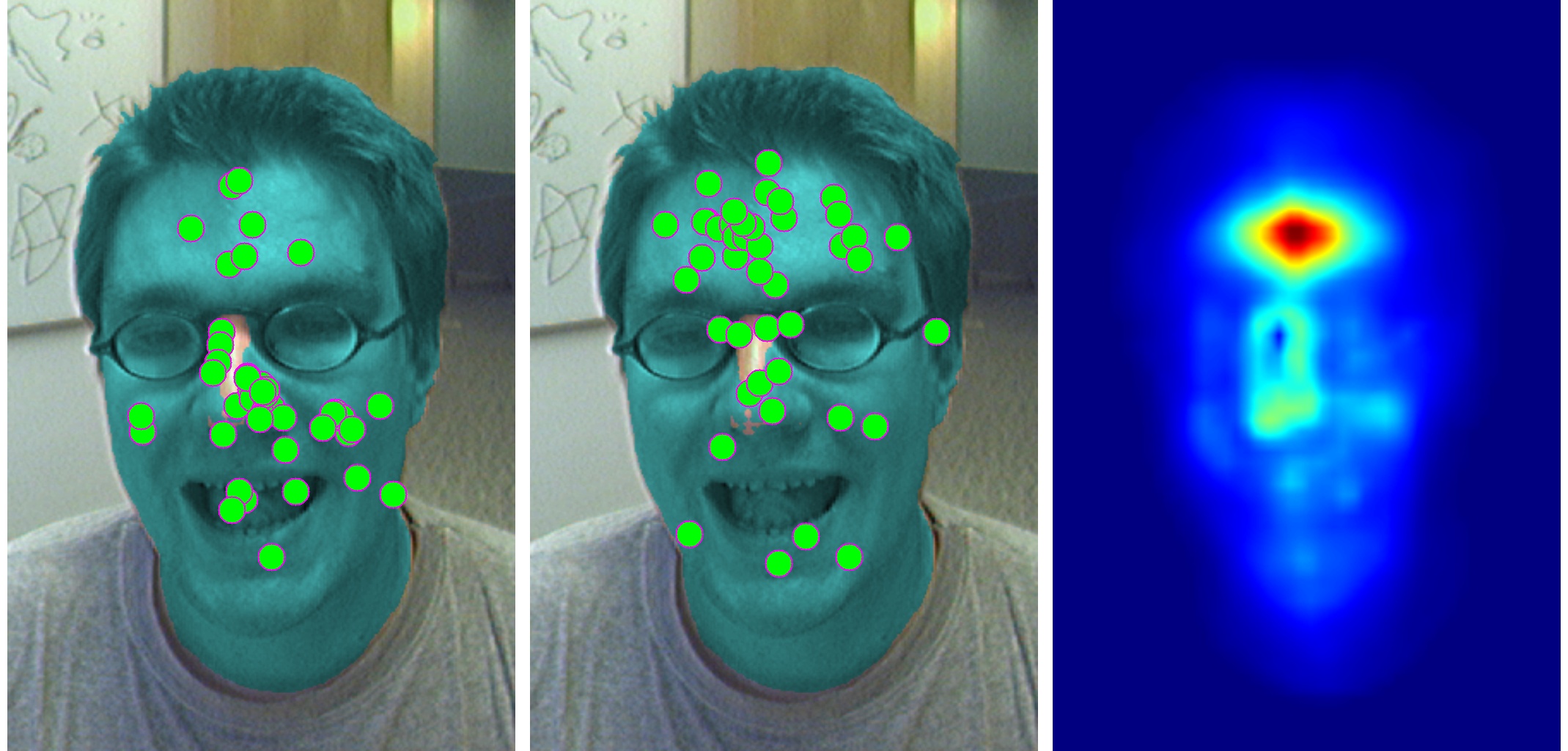}
    \end{subfigure}

    \begin{subfigure}[t]{0.45\textwidth}
        \centering
        \caption{Berkeley, KS-test -- True}
        \includegraphics[width=1\textwidth]{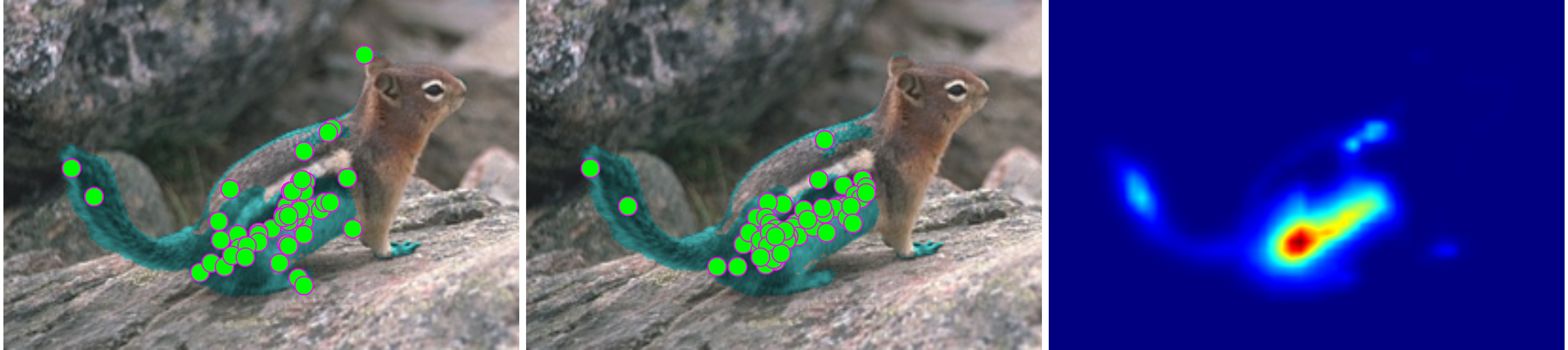}
    \end{subfigure}
    \begin{subfigure}[t]{0.45\textwidth}
        \centering
        \caption{Berkeley, KS-test -- False}        
        \includegraphics[width=1\textwidth]{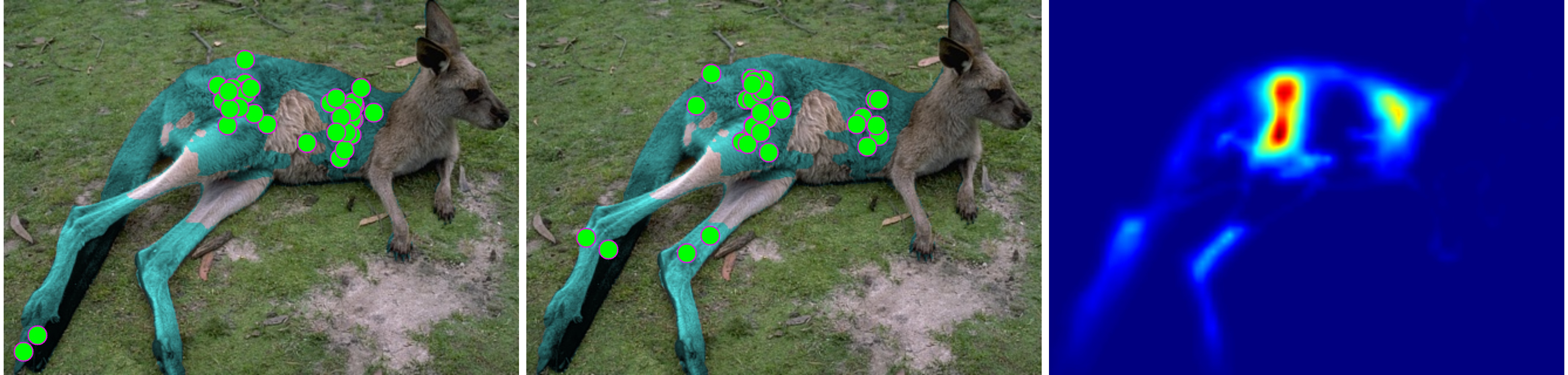}
    \end{subfigure}

    \begin{subfigure}[t]{0.45\textwidth}
        \centering
        \caption{DAVIS, KS-test -- True}
        \includegraphics[width=1\textwidth]{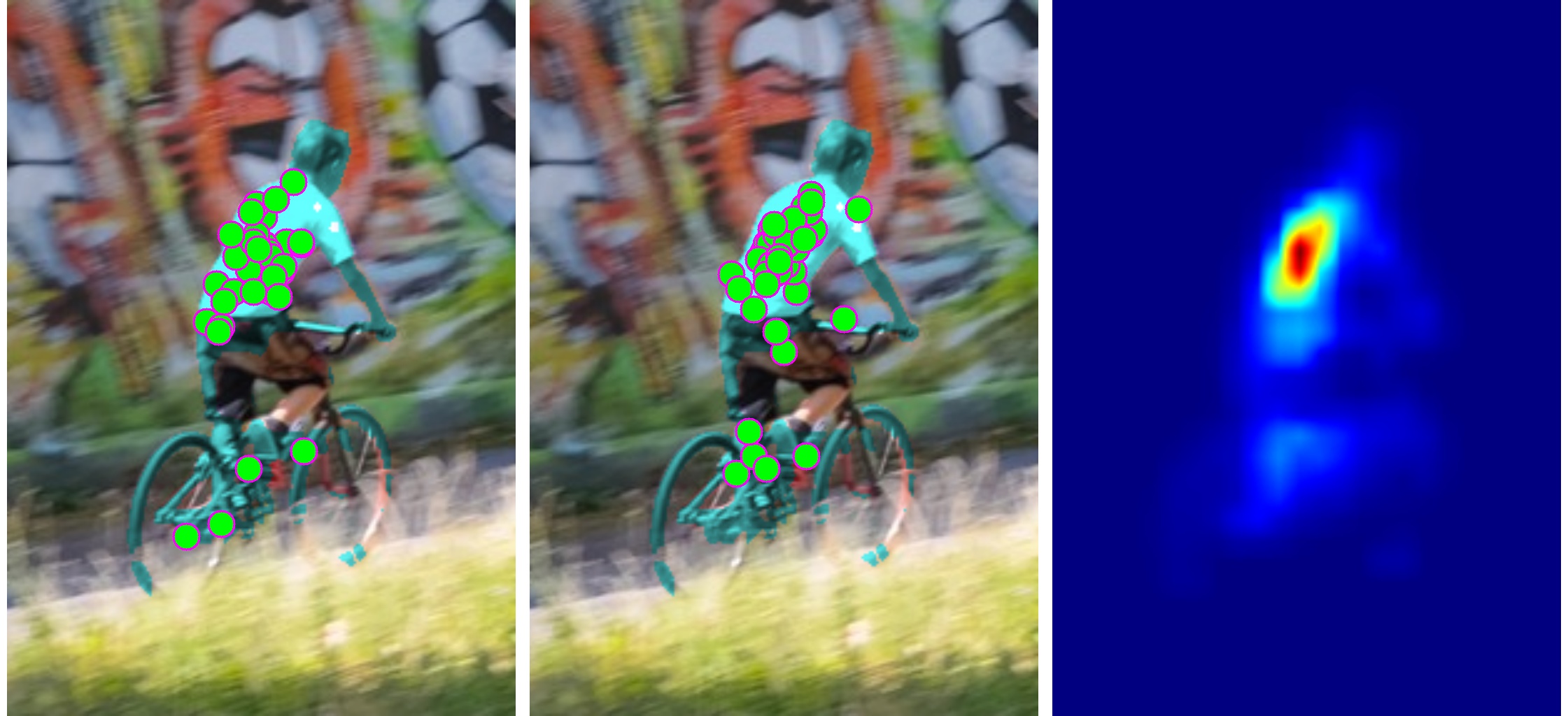}
    \end{subfigure}
    \begin{subfigure}[t]{0.45\textwidth}
        \centering
        \caption{DAVIS, KS-test -- False}        
        \includegraphics[width=1\textwidth]{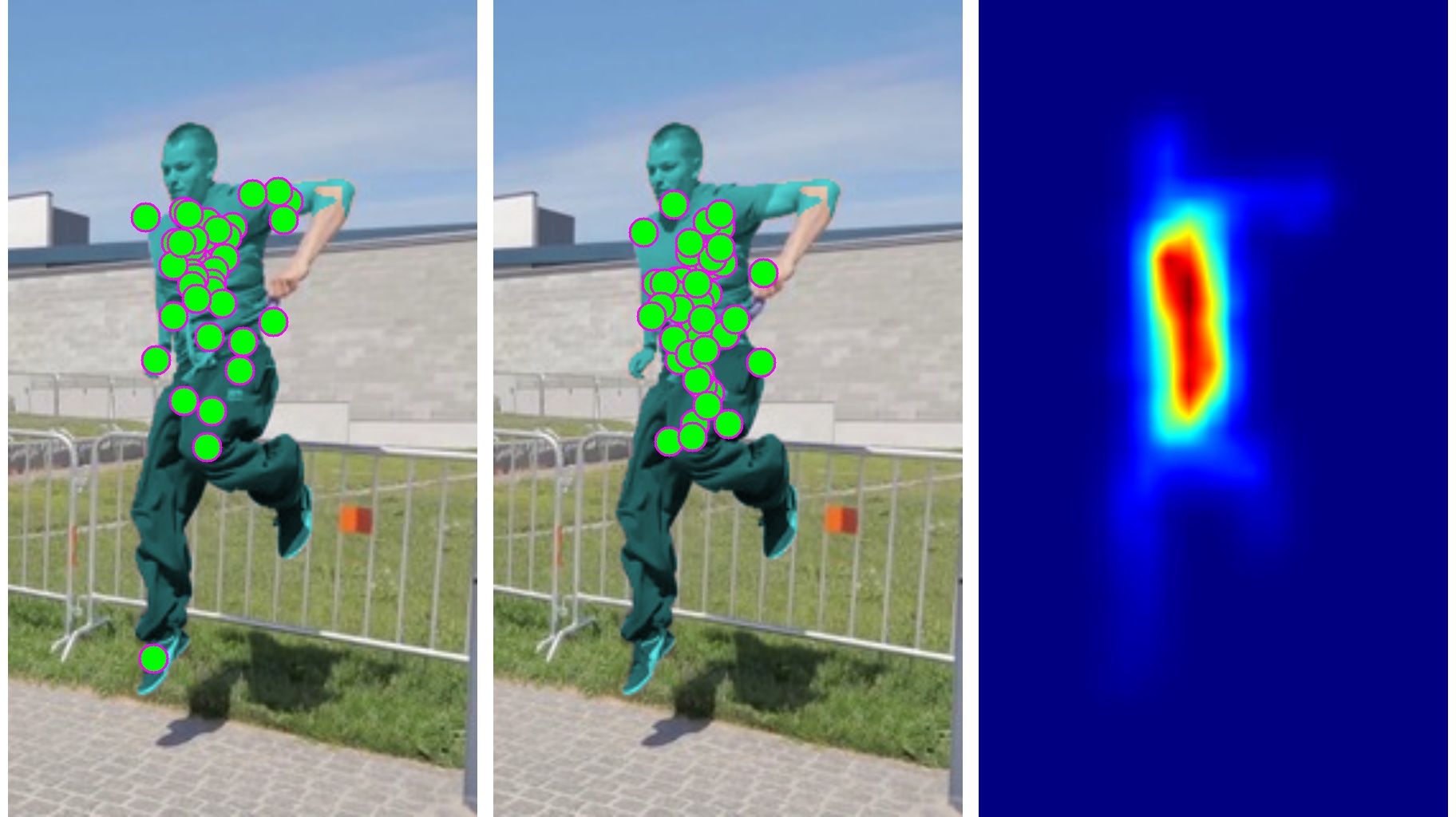}
    \end{subfigure}

    \begin{subfigure}[t]{0.45\textwidth}
        \centering
        \caption{COCO-MVal, KS-test -- True}
        \includegraphics[width=1\textwidth]{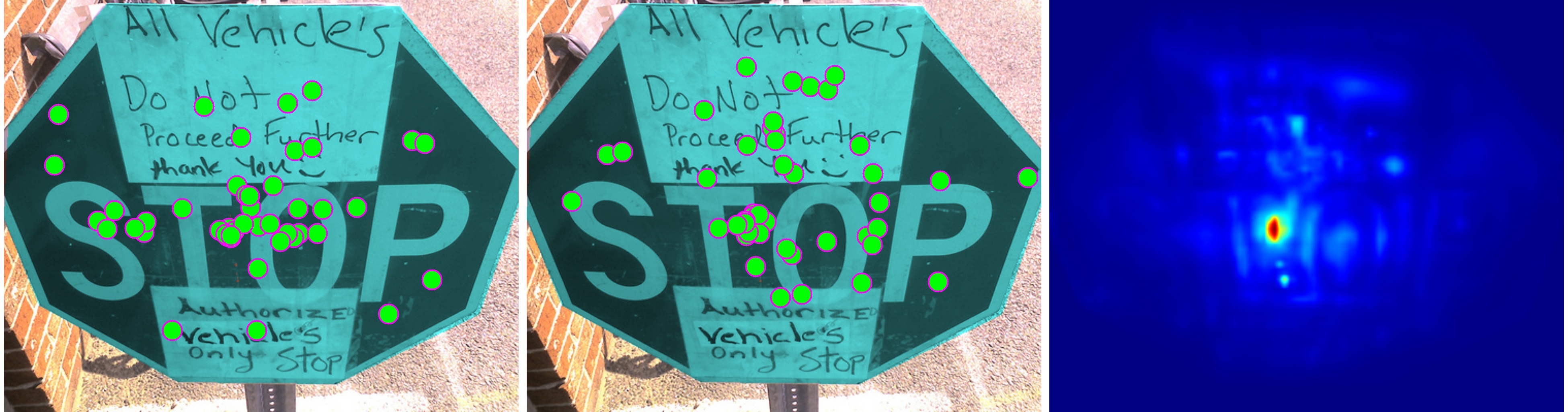}
    \end{subfigure}
    \begin{subfigure}[t]{0.45\textwidth}
        \centering
        \caption{COCO-MVal, KS-test -- False}        
        \includegraphics[width=1\textwidth]{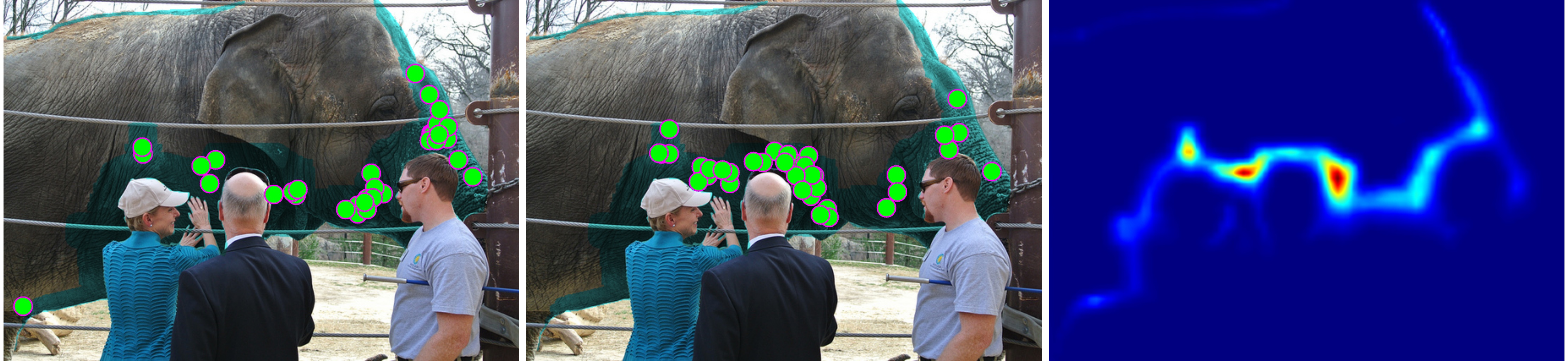}
    \end{subfigure}

    \begin{subfigure}[t]{0.45\textwidth}
        \centering
        \caption{TETRIS, KS-test -- True}
        \includegraphics[width=1\textwidth]{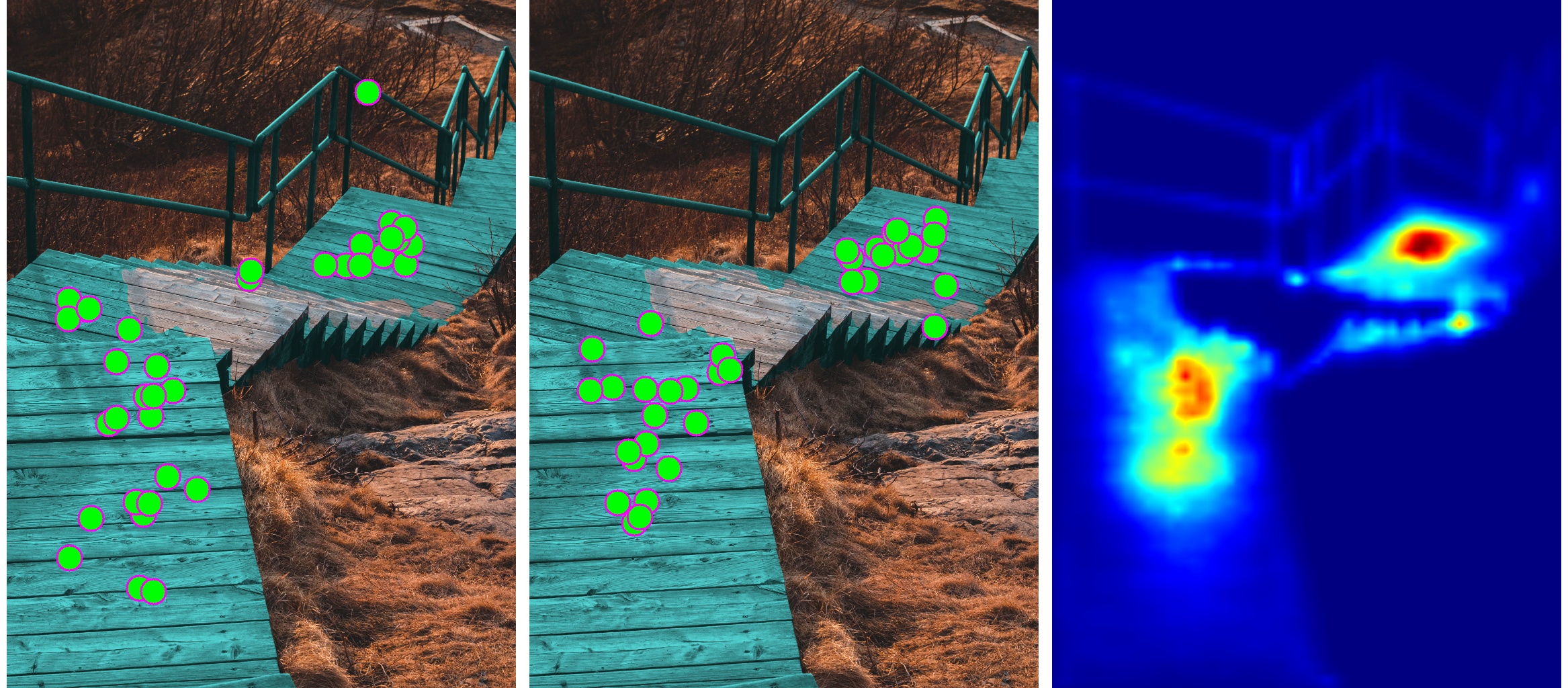}
    \end{subfigure}
    \begin{subfigure}[t]{0.45\textwidth}
        \centering
        \caption{TETRIS, KS-test -- False}        
        \includegraphics[width=1\textwidth]{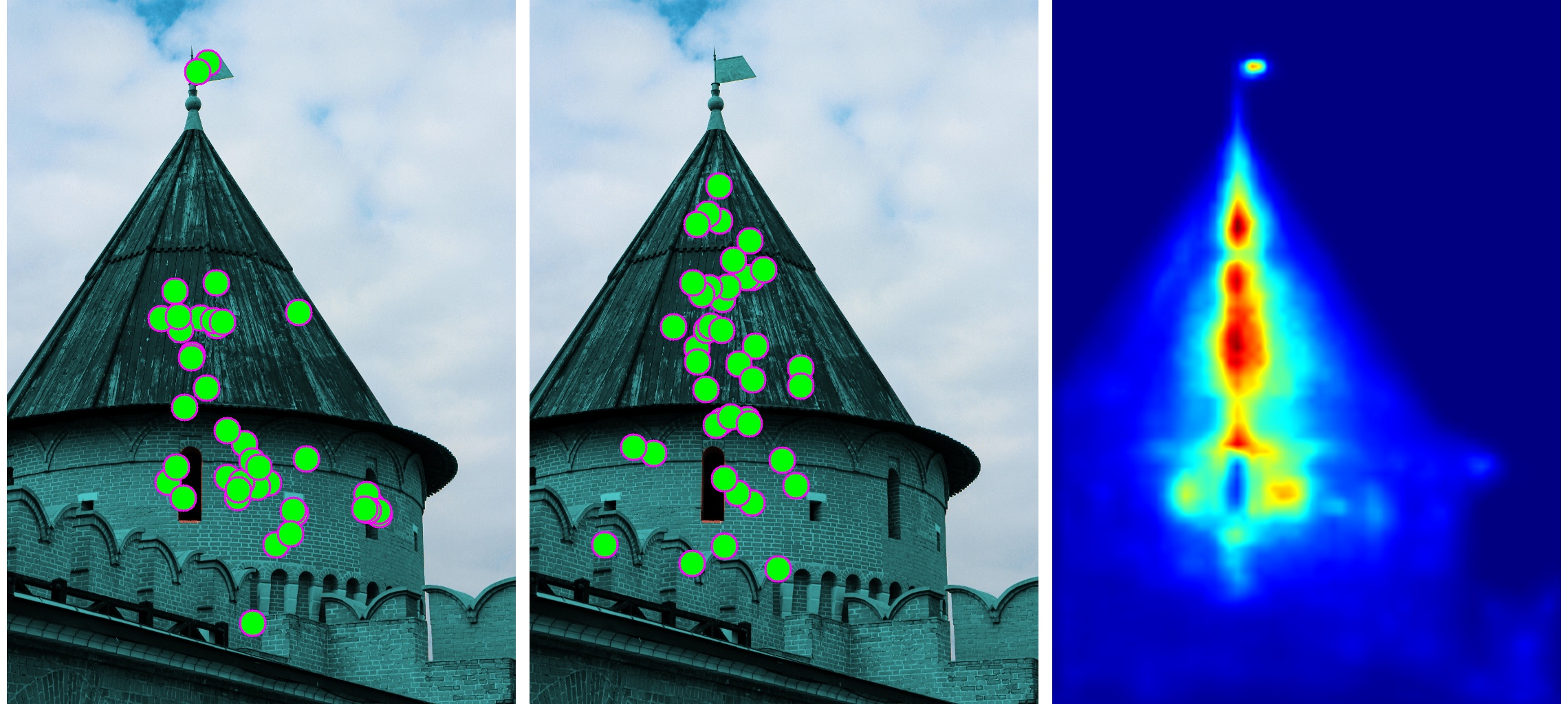}
    \end{subfigure}
    
    \caption{Examples of false-negative (FN) ground truth clicks (left), generated clicks (middle) and predicted \textit{clickability map} (right) by our \textit{clickability model} for the subsequent round.
    Green dots illustrate FN clicks, teal masks represent target regions that should be segmented.
    In the subcaptions the results of Kolmogorov-Smirnov test (True if p-value > 0.05, i.e. there are no significant differences between the distributions of clicks) are provided.
    }
    \label{fig:examples:fn}
\end{figure}

Figures~\ref{fig:examples:first_round},~\ref{fig:examples:fp},~\ref{fig:examples:fn} present examples of the collected ground truth clicks, generated clicks and predicted clickability maps by our clickability model.

\FloatBarrier

\section{Clicks Dataset}
\label{app:clicks_dataset}

\subsection{Collection procedure}
\label{sec:data_collection_procedure}

Before starting data collection, we considered various ways to visually present the mask and image to a person.
These are all the methods we found reasonable.

Here is a brief story behind choosing display mode:
\begin{enumerate}
    \item The most unbiased display mode to instruct a participant is considered to be Text Description.
    By avoiding any visual display of the mask, we eliminate potential biases. 
    However, we could not use it to annotate all images because such descriptions can be ambiguous.
    We also cannot use this method for subsequent rounds because we need to indicate the part of the mask that was not segmented in the previous round.
    \item The next display mode we tried was to show the ground truth black-and-white mask (Silhouette Mask).
    In that case, we simply show the mask as is without any modifications.
    We collected the first batch of data and noticed that participants tend to click to the geometric center of the object.
    Therefore, this strategy seemed to be biased.
    We started to look for the least unbiased one.
    \item To prevent people from selecting the geometric center of the object, we added Object CutOut mode.
    We hypothesize that it is harder to be biased on the geometric form of an object when it retains its appearance.
    \item After that, we tried Shifted CutOut mode.
    The idea of this mode was to eliminate dependency on initial cursor position.
    In that display mode, a person has to move the cursor to the object.
    \item We also tried Highlighted Instance mode, since it’s a natural way to highlight an object retaining its' background.
    We hypothesized that object background may influence click position.
\end{enumerate}

In determining the optimal display duration for our dataset collection, we conducted a thorough review of existing literature on annotation time requirements.
A number of studies have been conducted on this subject ~\citep{delatolas2024learning}\,(Sec. 4), ~\citep{ravi2024sam}\,(Sec. E.1.2), ~\citep{Papadopoulos2014TrainingOC}\,(Sec. 2), ~\citep{papadopoulos2017training}\,(Sec. 3.4), ~\citep{bearman2016s}\,(Sec. 4.1).
Researchers came to the conclusion that the required time plan formula is the following: 1 sec. for the annotator to visually locate an object and 1.5 sec. for adding each click, while total reported time to localize and click may vary between 1.87 sec. and 2.5 sec.
Since, we did not have the goal of speeding up the labeling of RClicks, but to collect high-quality data, we increased times as follows: showing an image for 1.5 sec., different mask displaying modes for 2 sec. (since time is needed to localize and remember the target object) and Text Description mode for 2.5 sec. (since more time is needed to read and understand the text).

To annotate datasets, we used the \href{https://toloka.ai}{toloka.ai} crowdsourcing platform.
Each crowd worker was paid \$0.02 for 10 clicks. On average, workers made 13 clicks per minute, earning approximately \$1.5 per hour. This level is above minimum hour wage for the countries the annotators were from.
Overall we spent about 1300\$ on annotation. 

According to the \href{https://join.toloka.ai/legal/user-agreement/}{user agreement} and \href{https://toloka.ai/privacy-notice}{privacy policy} of Toloka, personal data typically includes information that can identify an individual, such as name, contact information, and other personal identifiers. Annotators clicks on images do not fall under this category.
Moreover, we provide fully anonymized data, that can not be linked with people who clicked.
Toloka's policy allows for the sharing of anonymized data with third parties. If the collected click data is anonymized and cannot be traced back to individual annotators, it may be shared with third parties without violating the privacy terms.

Here is an example of an instruction, which annotators saw, for Object CutOut mode:

\tcbset{
    colframe=blue!75!black,
    colback=blue!5!white,
    coltitle=white,
    colbacktitle=blue!75!black,
    fonttitle=\bfseries\sffamily\large,
    fontupper=\sffamily,
    boxrule=0.5mm,
    sharp corners,
}

\begin{tcolorbox}[title=Instructions]

\begin{enumerate}[label=\bfseries\arabic*.]
    
\item After opening the task, there will be an image loading period of approximately 30 seconds.

\item After clicking the ``START'' button, an image will be shown for 1.5 seconds, followed by a demonstration of the object of interest on a gray background for 2 seconds.
Note that during this time, you cannot click on the object!

Example:

\begin{center}
\includegraphics[width=0.5\textwidth]{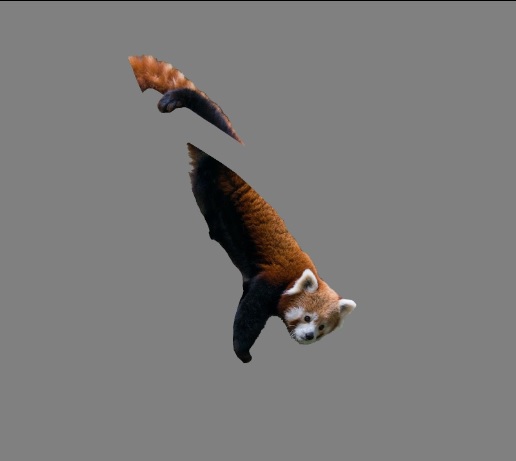}
\end{center}

\item Then you will see the original image again, which will displayed for 1.5 second. 
Your task is to select the object that was indicated in the previous step with one click.

\item To successfully complete the task, it is necessary to process 10 images in sequence.

\end{enumerate}
\end{tcolorbox}

\newpage

\subsection{Comparison of PC and mobile clicks}

\begin{figure}[!ht]
    \centering
    \begin{subfigure}[t]{0.45\textwidth}
        \centering
        \caption{First round, KS-test -- True}
        \includegraphics[width=1\textwidth] {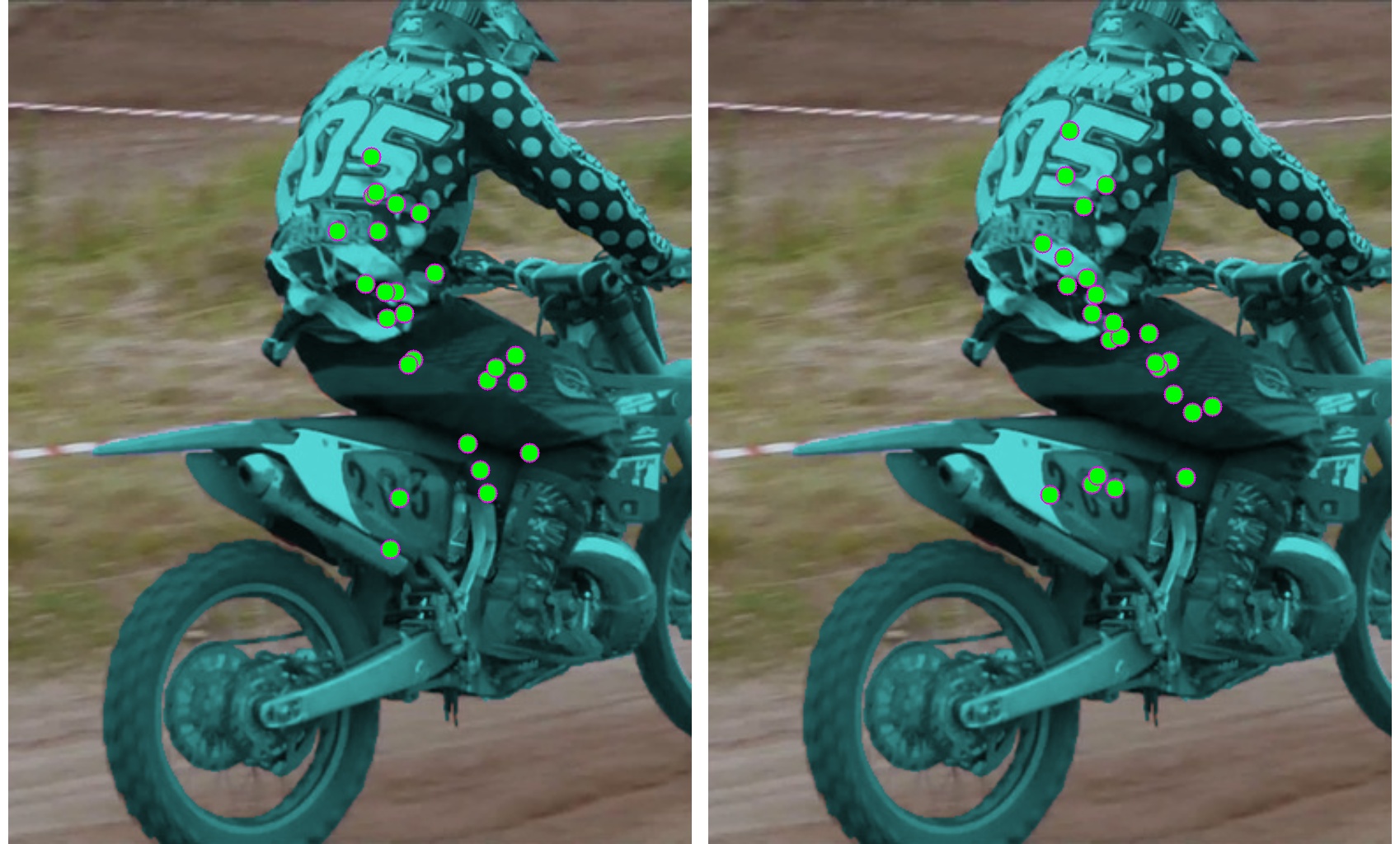}
    \end{subfigure}
    \begin{subfigure}[t]{0.45\textwidth}
        \centering
        \caption{First round, KS-test -- False}        
        \includegraphics[width=1\textwidth]{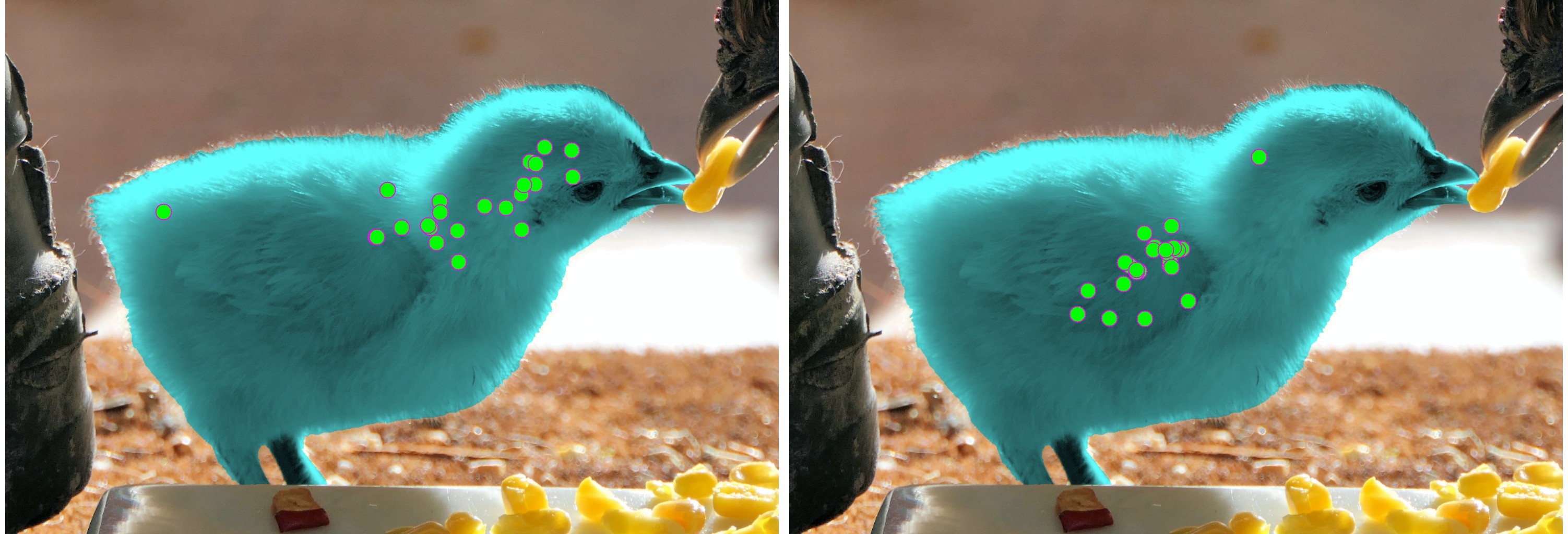}
    \end{subfigure}

    \begin{subfigure}[t]{0.45\textwidth}
        \centering
        \caption{Subsequent round (FP), KS-test -- True}
        \includegraphics[width=1\textwidth]{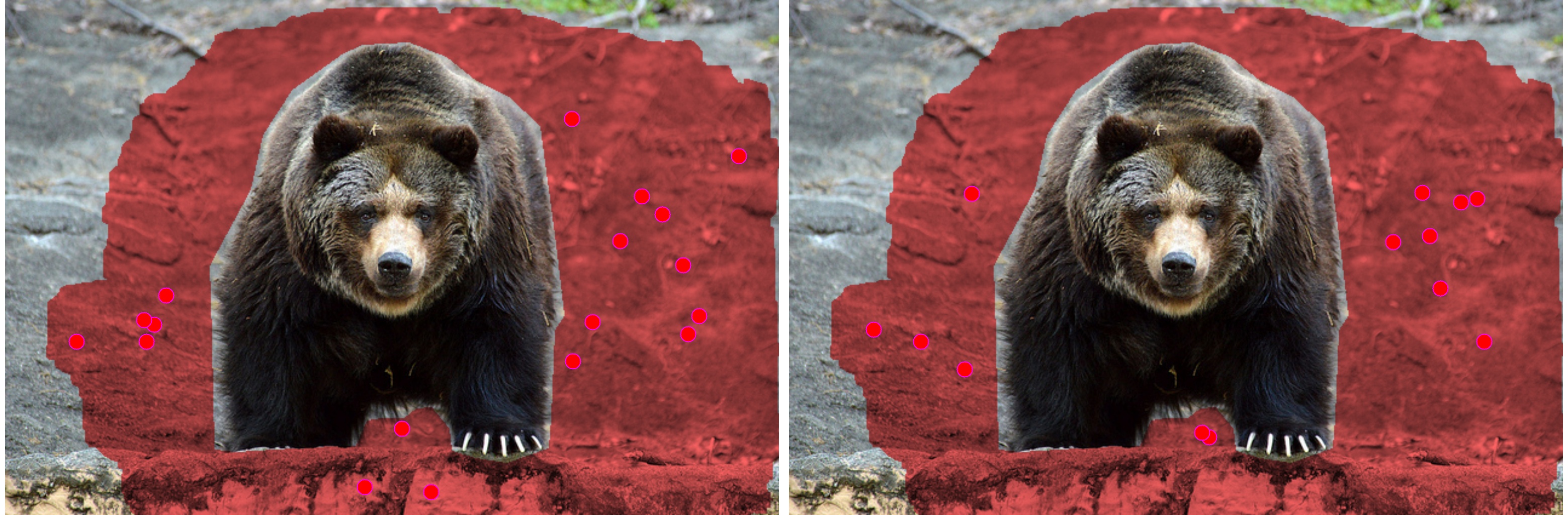}
    \end{subfigure}
    \begin{subfigure}[t]{0.45\textwidth}
        \centering
        \caption{Subsequent round (FP), KS-test -- False}        
        \includegraphics[width=1\textwidth]{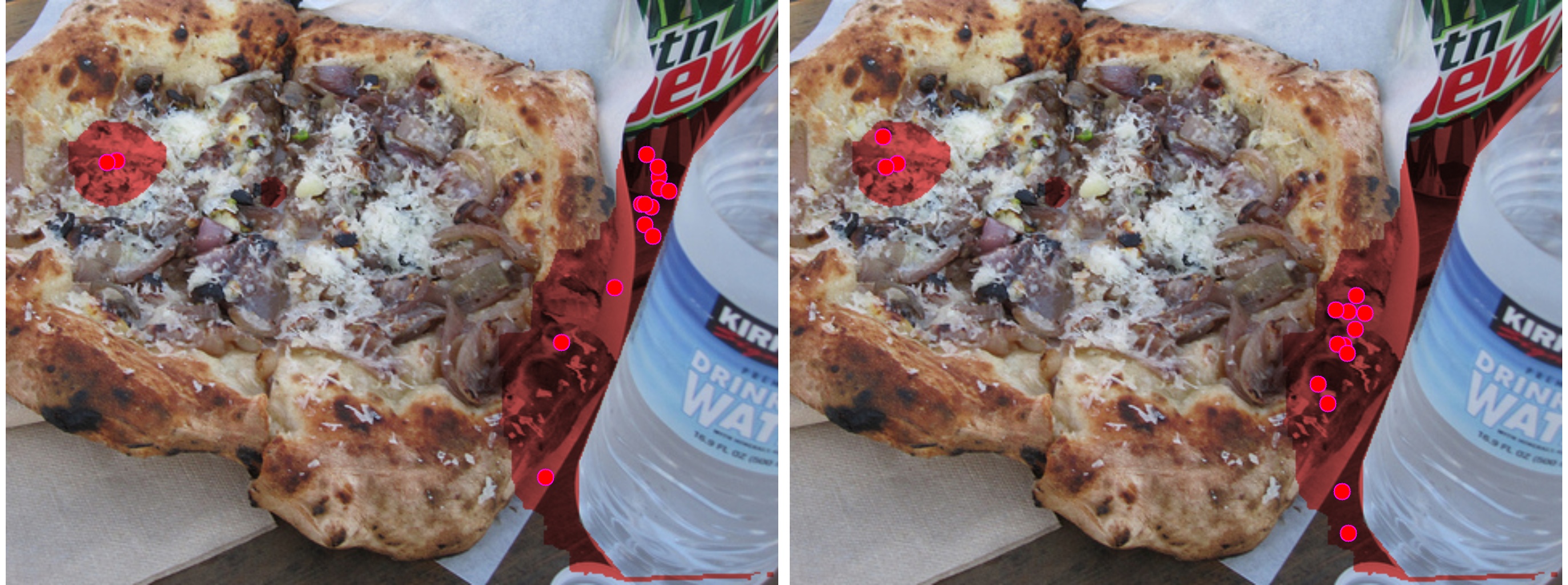}
    \end{subfigure}

    \begin{subfigure}[t]{0.45\textwidth}
        \centering
        \caption{Subsequent round (FN), KS-test -- True}
        \includegraphics[width=1\textwidth]{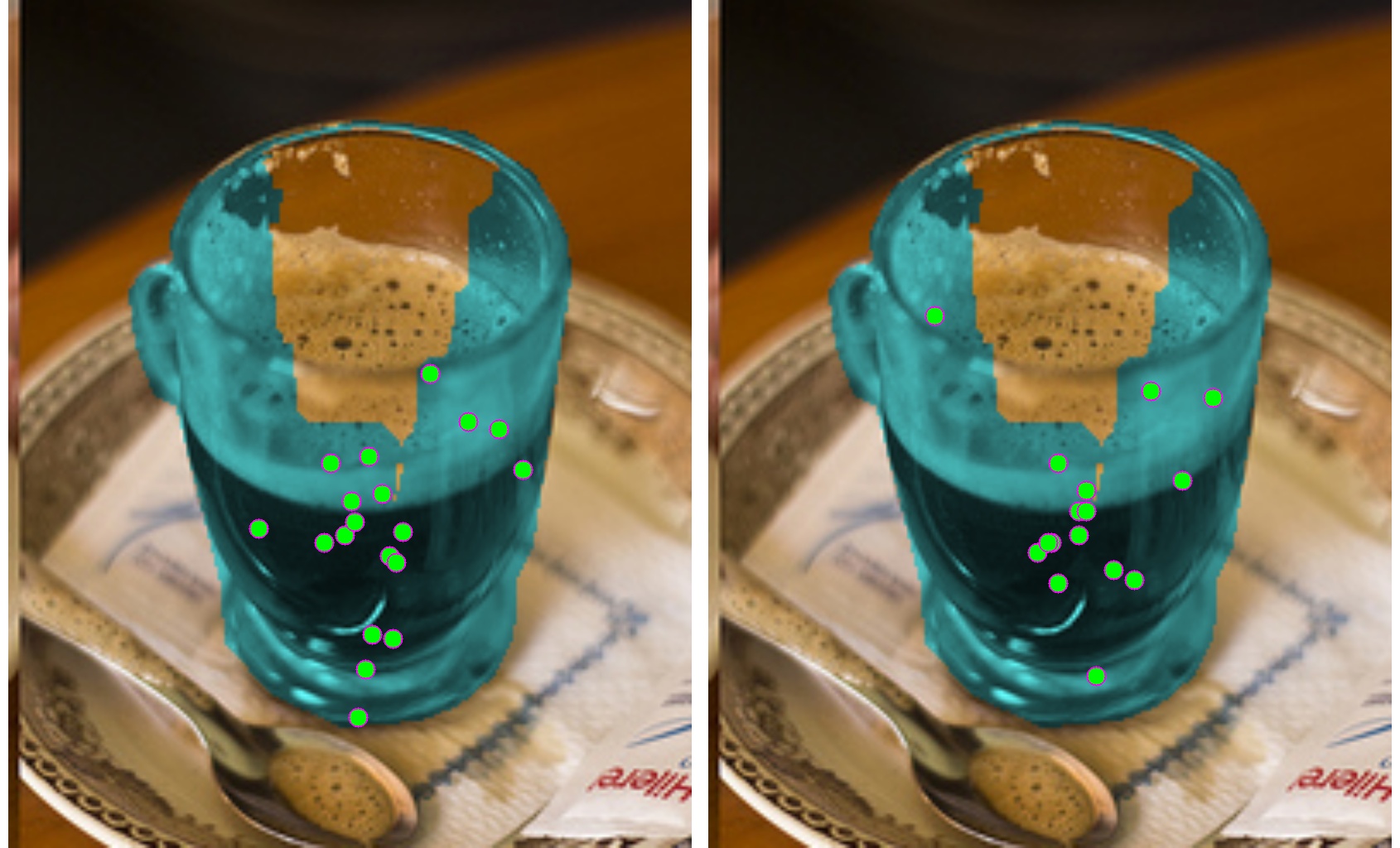}
    \end{subfigure}
    \begin{subfigure}[t]{0.45\textwidth}
        \centering
        \caption{Subsequent round (FN), KS-test -- False}
        \includegraphics[width=1\textwidth]{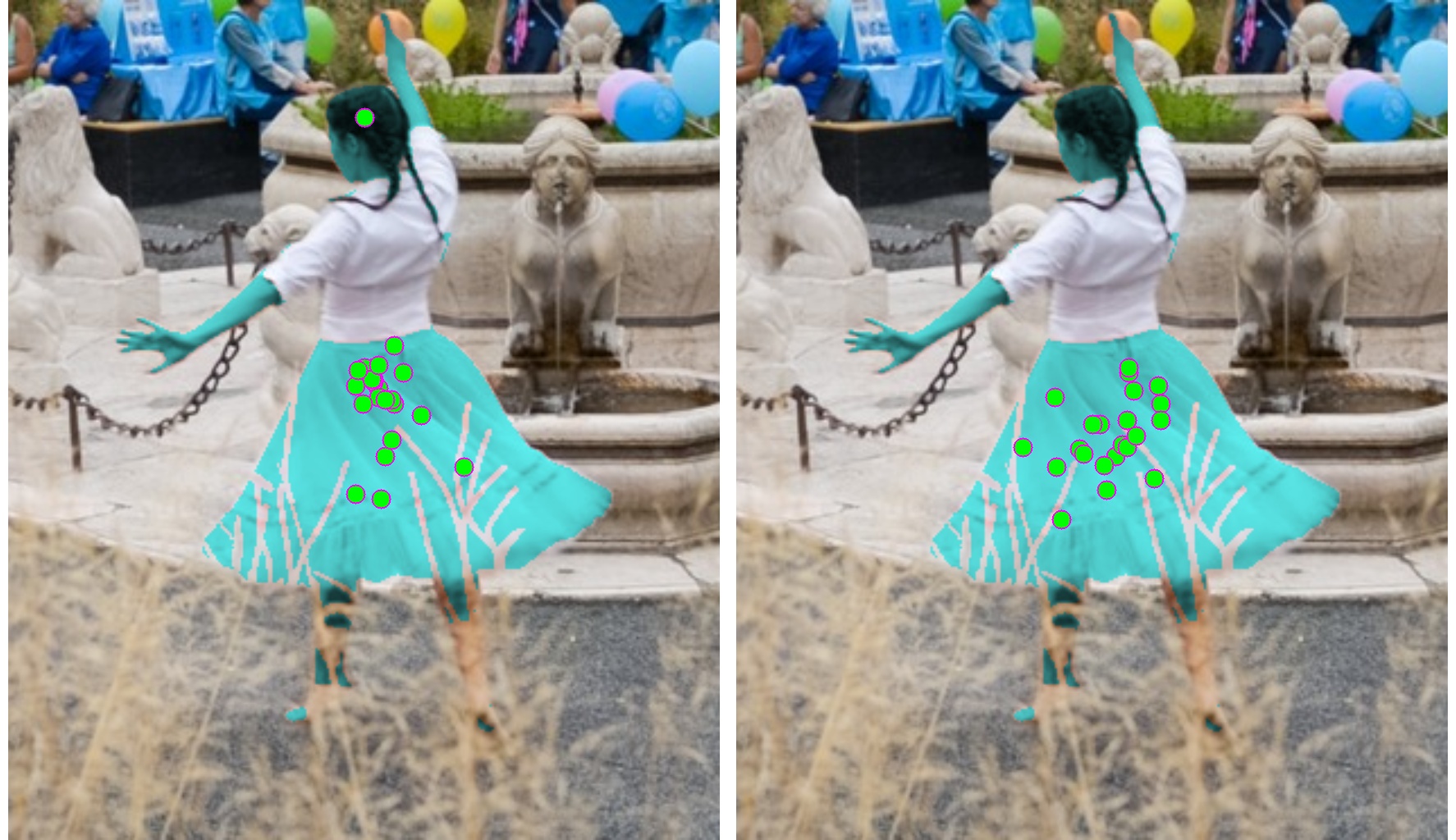}
    \end{subfigure}
    
    \caption{Examples of collected clicks for PC (left) and mobile (right) devices for the first and subsequent rounds.
    Green dots illustrate the first round and subsequent false-negative (FN) clicks, red dots -- the subsequent false-positive (FP) clicks.
    Teal masks represent target regions that should be segmented, red masks -- regions of segmentation errors.
    In the subcaptions the results of Kolmogorov-Smirnov test (True if p-value > 0.05, i.e. there are no significant differences between the distributions of clicks) are provided.
    }
    \label{fig:examples_pc_vs_mobile}
\end{figure}

\begin{wraptable}{r}{0.45\textwidth}
\vspace{-1.5em}
\caption{Comparison clicks collected from PC and mobile devices}
\label{tab:pc_mobile_clicks}
\fontsize{10pt}{14pt}\selectfont
\tabcolsep=3pt
\centering
\begin{tabular}{cccc}
\toprule
Dataset & \multicolumn{1}{l}{KS\,$\uparrow$} & \multicolumn{1}{l}{WD\,$\downarrow$} & \multicolumn{1}{l}{PL$_1$\,$\downarrow$} \\
\midrule
GrabCut  & 0.63 & 0.26 & 0.54 \\
Berkeley & 0.64 & 0.41 & 0.79 \\
DAVIS    & 0.62 & 0.58 & 1.08 \\
COCO-MV     & 0.68 & 1.15 & 2.22 \\
TETRIS   & 0.70 & 0.41 & 0.70 \\
\bottomrule
\end{tabular}
\vspace{-2em}
\end{wraptable}
Table~\ref{tab:pc_mobile_clicks} present the average comparison metrics clicks collected from PC and mobile devices.
We averaged only clicks for only instances that have more than 10 clicks for both devices.
According to KS column, for 60\% of the compared instances, there were no significant difference between PC and mobile distributions.
In Figure~\ref{fig:examples_pc_vs_mobile} we provide examples that illustrate both significant differences and cases without significant differences between PC and mobile.
\vspace{4em}

\FloatBarrier

\newpage

\section{Additional Benchmark Results}
\label{app:additional_results}

\subsection{Evaluation setup}
We benchmarked 11 methods with 33 checkpoints.
To do this, we spent 2400 GPU hours, which is equivalent to approximately 6 days of compute using 16 NVIDIA Tesla A100 GPUs.

\subsection{Additional evaluation results}
\label{app:additional_eval_results}

Tables~\ref{tab:full_noc},~\ref{tab:full_nof},~\ref{tab:full_iou} present evaluation results on simulated user clicks of various interactive segmentation algorithms on GrabCut, Berkeley, DAVIS, COCO-MVal and TETRIS datasets for NoC$_{20}$@90, NoF$_{20}$@90 and IoU-AuC$_{20}$ respectively.
Table~\ref{tab:full_iou_1} presents the evaluation results on the first round real and simulated user clicks.

\textbf{Base} statistics represent the performance of the methods according to baseline benchmark.
In addition to the statistics presented in the main paper, we calculated \textbf{$\Delta$HH} statistic -- this statistic is based on $\Delta$GR, but for $\Delta$HH we merge intervals \{G$_i$\}$^5_{i=1}$ and \{G$_i$\}$^{10}_{i=6}$ s.t. every interval has 50\% of total probability mass.
Moreover, we calculated noise-signal-ratio (\textbf{NSR}) of IoU averaged over the datasets instances.

\begin{figure}[ht!]
\includegraphics[width=1.0\textwidth]{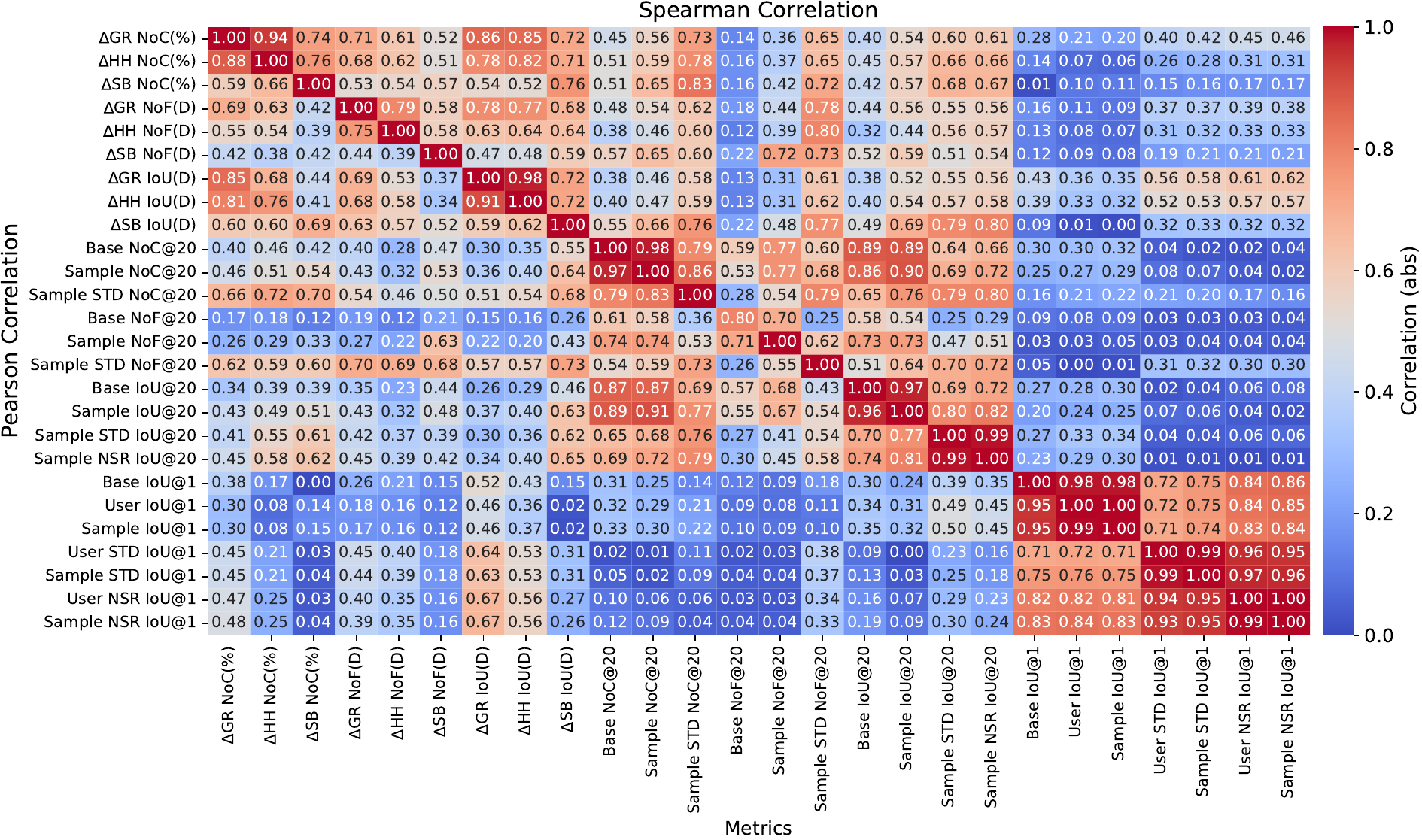}
\caption{Pearson\textbackslash Spearman correlations of considered averaged performance and robustness metrics statistics averaged over all instances in all datasets.}
\label{fig:corr}
\end{figure}

Figure~\ref{fig:corr} shows the correlations between different metrics. From this, we can draw several conclusions:
\begin{enumerate}
    \item There is a strong correlation between Sample and User performances on the first click, as seen in Sample IoU@1 and User IoU@1, as well as Sample NSR IoU@1 and User NSR IoU@1. This demonstrates that performance on real and simulated clicks is very similar.
    \item There is a weak correlation between performance on the first and twentieth clicks, as seen in the Sample IoU@1 with Sample IoU@20 and Base IoU@1 with Base IoU@20. This likely indicates that all models converge by the 20th round.
    \item There is a weak correlation between robustness metrics and average model performance. This is observed in: $\Delta$GR NoC with (Base NoC@20 and Sample NoC@20), $\Delta$HH NoC with (Base NoC@20 and Sample NoC@20), $\Delta$GR IoU with (Base IoU@20 and Sample IoU@20), and $\Delta$HH IoU with (Base IoU@20 and Sample IoU@20). This suggests that high-performance methods may not be robust.
    \item Performance measures have high correlations with each other, as seen in the cross-correlations between Sample NoC, Sample IoU, Sample NoF and Base NoC, Base IoU, Base NoF. 
    \item The same phenomenon is observed with robustness metrics, demonstrated by the cross-correlations between $\Delta$HH NoC, $\Delta$GR NoC and $\Delta$HH IoU, $\Delta$GR IoU.
\end{enumerate}

\begin{sidewaystable}
\centering
\caption{NoC evaluation results}
\label{tab:full_noc}
\fontsize{7pt}{12pt}\selectfont
\tabcolsep=1.2pt
\begin{tabular}{cccc|ccccccc|ccccccc|ccccccc|ccccccc|cccccc}
\toprule
\multirow{3}{*}{Method}                                                   & \multirow{3}{*}{Model} & \multirow{3}{*}{Data} &  &  & \multicolumn{5}{c}{GrabCut \citep{grabcut}}                                                                                                                                                                                                             &  &  & \multicolumn{5}{c}{Berkeley \citep{berkeley}}                                                                                                                                                                                                           &  &  & \multicolumn{5}{c}{DAVIS \citep{davis}}                                                                                                                                                                                                                 &  &  & \multicolumn{5}{c}{COCO-MVal \citep{coco}}                                                                                                                                                                                                              &  &  & \multicolumn{5}{c}{TETRIS \citep{moskalenko2024tetris}}                                                                                                                                                                                                 \\
                                                                          &                        &                       &  &  & \multicolumn{5}{c}{$\textrm{NoC}_{\textrm{20}}\textrm{@}\textrm{90}$}                                                                                                                                                                                                    &  &  & \multicolumn{5}{c}{$\textrm{NoC}_{\textrm{20}}\textrm{@}\textrm{90}$}                                                                                                                                                                                                    &  &  & \multicolumn{5}{c}{$\textrm{NoC}_{\textrm{20}}\textrm{@}\textrm{90}$}                                                                                                                                                                                                    &  &  & \multicolumn{5}{c}{$\textrm{NoC}_{\textrm{20}}\textrm{@}\textrm{90}$}                                                                                                                                                                                                    &  &  & \multicolumn{5}{c}{$\textrm{NoC}_{\textrm{20}}\textrm{@}\textrm{90}$}                                                                                                                                                                                                    \\
                                                                          &                        &                       &  &  & \begin{tabular}[c]{@{}c@{}}Sample\\ (±std)\end{tabular} & Base              & \begin{tabular}[c]{@{}c@{}}${\Delta}$SB\\ (+\%)\end{tabular} & \begin{tabular}[c]{@{}c@{}}${\Delta}$GR\\ (+\%)\end{tabular} & \begin{tabular}[c]{@{}c@{}}${\Delta}$HH\\ (+\%)\end{tabular} &  &  & \begin{tabular}[c]{@{}c@{}}Sample\\ (±std)\end{tabular} & Base              & \begin{tabular}[c]{@{}c@{}}${\Delta}$SB\\ (+\%)\end{tabular} & \begin{tabular}[c]{@{}c@{}}${\Delta}$GR\\ (+\%)\end{tabular} & \begin{tabular}[c]{@{}c@{}}${\Delta}$HH\\ (+\%)\end{tabular} &  &  & \begin{tabular}[c]{@{}c@{}}Sample\\ (±std)\end{tabular} & Base              & \begin{tabular}[c]{@{}c@{}}${\Delta}$SB\\ (+\%)\end{tabular} & \begin{tabular}[c]{@{}c@{}}${\Delta}$GR\\ (+\%)\end{tabular} & \begin{tabular}[c]{@{}c@{}}${\Delta}$HH\\ (+\%)\end{tabular} &  &  & \begin{tabular}[c]{@{}c@{}}Sample\\ (±std)\end{tabular} & Base              & \begin{tabular}[c]{@{}c@{}}${\Delta}$SB\\ (+\%)\end{tabular} & \begin{tabular}[c]{@{}c@{}}${\Delta}$GR\\ (+\%)\end{tabular} & \begin{tabular}[c]{@{}c@{}}${\Delta}$HH\\ (+\%)\end{tabular} &  &  & \begin{tabular}[c]{@{}c@{}}Sample\\ (±std)\end{tabular} & Base              & \begin{tabular}[c]{@{}c@{}}${\Delta}$SB\\ (+\%)\end{tabular} & \begin{tabular}[c]{@{}c@{}}${\Delta}$GR\\ (+\%)\end{tabular} & \begin{tabular}[c]{@{}c@{}}${\Delta}$HH\\ (+\%)\end{tabular} \\ \cline{1-3} \cline{6-10} \cline{13-17} \cline{20-24} \cline{27-31} \cline{34-38} 
GPCIS \citep{zhou2023interactivegpcis}                   & RN50                   & C+L                   &  &  & 2.04±0.49                                               & 1.80              & 13.58                                                        & 61.59                                                        & 20.38                                                        &  &  & 3.06±0.83                                               & 2.61              & 22.13                                                        & 57.13                                                        & 24.53                                                        &  &  & 6.44±0.85                                               & 5.94              & 16.88                                                        & 53.65                                                        & 19.31                                                        &  &  & 4.74±1.31                                               & 4.05              & 26.43                                                        & 79.00                                                        & 30.02                                                        &  &  & 3.87±0.79                                               & 3.39              & 19.55                                                        & 56.43                                                        & 22.90                                                        \\ \cline{1-3} \cline{6-10} \cline{13-17} \cline{20-24} \cline{27-31} \cline{34-38} 
\multirow{2}{*}{CDNet \citep{chen2021conditional}}       & RN34                   & C+L                   &  &  & 1.72±0.38                                               & 1.52              & 12.89                                                        & 52.57                                                        & 20.38                                                        &  &  & 2.33±0.60                                               & 2.06              & 10.02                                                        & 49.17                                                        & 19.03                                                        &  &  & 5.95±0.73                                               & 5.56              & 14.95                                                        & 45.88                                                        & 18.56                                                        &  &  & 4.13±0.85                                               & 3.88              & 15.15                                                        & 49.79                                                        & 22.56                                                        &  &  & 3.10±0.53                                               & 2.83              & 14.16                                                        & 44.06                                                        & 17.25                                                        \\
                                                                          & RN34                   & SBD                   &  &  & 2.32±0.46                                               & 2.18              & 11.57                                                        & 66.73                                                        & 21.89                                                        &  &  & 3.82±0.97                                               & 3.27              & 20.48                                                        & 55.40                                                        & 22.76                                                        &  &  & 7.87±1.25                                               & 6.89              & 23.39                                                        & 64.95                                                        & 26.09                                                        &  &  & 6.36±1.29                                               & 5.88              & 20.86                                                        & 51.58                                                        & 24.18                                                        &  &  & 4.51±0.76                                               & 4.18              & 17.08                                                        & 55.22                                                        & 22.27                                                        \\ \cline{1-3} \cline{6-10} \cline{13-17} \cline{20-24} \cline{27-31} \cline{34-38} 
\multirow{5}{*}{RITM \citep{ritm}}                       & HR18                   & C+L                   &  &  & 1.68±0.28                                               & 1.70              & 5.78                                                         & 30.83                                                        & 14.01                                                        &  &  & 2.81±0.65                                               & 2.48              & 13.86                                                        & 21.05                                                        & 12.66                                                        &  &  & 6.23±0.67                                               & 6.00              & \underline{6.92}                                            & 16.13                                                        & 6.59                                                         &  &  & 3.71±0.78                                               & 3.58              & 10.27                                                        & 20.22                                                        & 10.55                                                        &  &  & 3.69±0.52                                               & 3.57              & 7.02                                                         & 13.95                                                        & 7.76                                                         \\
                                                                          & HR18s-IT               & C+L                   &  &  & 1.82±0.40                                               & 1.68              & 12.98                                                        & 44.92                                                        & 18.28                                                        &  &  & 3.12±0.79                                               & 2.60              & 24.34                                                        & 31.75                                                        & 16.45                                                        &  &  & 6.71±0.99                                               & 5.98              & 20.88                                                        & 54.15                                                        & 19.71                                                        &  &  & 3.65±0.92                                               & 3.33              & 16.81                                                        & 33.89                                                        & 16.46                                                        &  &  & 3.80±0.67                                               & 3.46              & 15.79                                                        & 32.66                                                        & 14.86                                                        \\
                                                                          & HR18-IT                & C+L                   &  &  & 1.70±0.29                                               & 1.54              & 11.61                                                        & 31.40                                                        & 9.45                                                         &  &  & 2.61±0.53                                               & 2.26              & 14.76                                                        & 24.25                                                        & 7.05                                                         &  &  & 6.15±0.83                                               & 5.74              & 11.37                                                        & 31.14                                                        & 13.38                                                        &  &  & 3.22±0.83                                               & 2.98              & 15.84                                                        & 37.01                                                        & 17.10                                                        &  &  & 3.48±0.60                                               & 3.22              & 11.59                                                        & 23.99                                                        & 11.24                                                        \\
                                                                          & HR32-IT                & C+L                   &  &  & 1.64±0.36                                               & 1.56              & 9.83                                                         & 47.67                                                        & 14.10                                                        &  &  & 2.35±0.55                                               & 2.10              & 15.92                                                        & 27.52                                                        & 13.72                                                        &  &  & 5.90±0.89                                               & 5.34              & 18.34                                                        & 51.07                                                        & 21.96                                                        &  &  & 3.24±0.83                                               & 2.97              & 15.50                                                        & 37.31                                                        & 18.31                                                        &  &  & 3.44±0.65                                               & 3.06              & 17.47                                                        & 30.69                                                        & 13.08                                                        \\
                                                                          & HR18-IT                & SBD                   &  &  & 2.19±0.34                                               & 2.04              & 7.10                                                         & 21.61                                                        & 6.47                                                         &  &  & 3.71±0.79                                               & 3.22              & 15.80                                                        & 21.59                                                        & 11.99                                                        &  &  & 7.42±1.03                                               & 6.71              & 17.85                                                        & 38.39                                                        & 17.14                                                        &  &  & 4.81±1.24                                               & 4.39              & 17.17                                                        & 43.23                                                        & 19.47                                                        &  &  & 4.80±0.74                                               & 4.38              & 11.96                                                        & 25.31                                                        & 11.74                                                        \\ \cline{1-3} \cline{6-10} \cline{13-17} \cline{20-24} \cline{27-31} \cline{34-38} 
\multirow{2}{*}{AdaptClick \citep{lin2024adaptiveclick}} & ViT-B                  & C+L                   &  &  & \underline{1.40±0.16}                                  & 1.48              & \textbf{-2.03}                                               & 16.17                                                        & 4.43                                                         &  &  & 2.15±0.43                                               & 1.83              & 17.47                                                        & 12.21                                                        & 7.33                                                         &  &  & 4.97±0.40                                               & 4.81              & 8.60                                                         & 15.14                                                        & 8.55                                                         &  &  & 2.93±0.58                                               & 2.87              & 9.44                                                         & 19.75                                                        & 9.81                                                         &  &  & 2.62±0.37                                               & 2.51              & 6.99                                                         & 12.94                                                        & 6.82                                                         \\
                                                                          & ViT-B                  & SBD                   &  &  & \textit{1.42±0.22}                                      & 1.44              & 3.87                                                         & 12.80                                                        & \underline{3.06}                                            &  &  & 2.39±0.37                                               & 2.18              & 10.47                                                        & 12.63                                                        & 7.56                                                         &  &  & 5.37±0.49                                               & 5.15              & 8.69                                                         & 17.94                                                        & 9.94                                                         &  &  & 4.33±1.06                                               & 4.12              & 14.59                                                        & 35.07                                                        & 15.95                                                        &  &  & 3.49±0.50                                               & 3.31              & 8.40                                                         & 16.44                                                        & 8.28                                                         \\ \cline{1-3} \cline{6-10} \cline{13-17} \cline{20-24} \cline{27-31} \cline{34-38} 
\multirow{7}{*}{SimpleClick \citep{liu2022simpleclick}}  & ViT-B                  & C+L                   &  &  & 1.59±0.30                                               & 1.48              & 5.73                                                         & 22.50                                                        & 6.64                                                         &  &  & 2.01±0.40                                               & 1.97              & 7.33                                                         & 23.34                                                        & 10.55                                                        &  &  & 5.32±0.54                                               & 5.06              & 9.05                                                         & 26.33                                                        & 10.95                                                        &  &  & 3.07±0.70                                               & 2.92              & 11.72                                                        & 23.60                                                        & 11.91                                                        &  &  & 2.73±0.41                                               & 2.57              & 8.86                                                         & 16.64                                                        & 8.85                                                         \\
                                                                          & ViT-L                  & C+L                   &  &  & 1.45±0.24                                               & \underline{1.40} & 4.28                                                         & 22.13                                                        & 9.80                                                         &  &  & 1.83±0.29                                               & 1.89              & 5.12                                                         & 6.75                                                         & 6.28                                                         &  &  & 5.03±0.42                                               & 4.81              & 8.71                                                         & 16.67                                                        & 7.79                                                         &  &  & \underline{2.67±0.56}                                  & \textbf{2.63}     & 8.05                                                         & 20.88                                                        & 10.28                                                        &  &  & \textit{2.46±0.35}                                      & \textit{2.38}     & 7.11                                                         & 10.01                                                        & 7.06                                                         \\
                                                                          & ViT-H                  & C+L                   &  &  & 1.48±0.29                                               & 1.50              & \textit{2.93}                                                & 23.50                                                        & 8.16                                                         &  &  & 1.78±0.28                                               & 1.75              & \textit{4.60}                                                & 13.99                                                        & 6.17                                                         &  &  & 5.00±0.42                                               & 4.78              & \textit{7.06}                                                & 12.29                                                        & 6.51                                                         &  &  & \textbf{2.57±0.54}                                      & \underline{2.65} & \underline{6.14}                                            & 17.65                                                        & 10.04                                                        &  &  & \underline{2.36±0.33}                                  & \underline{2.26} & 6.94                                                         & 10.83                                                        & 6.64                                                         \\
                                                                          & ViT-XT                 & SBD                   &  &  & 2.26±0.49                                               & 2.28              & 4.68                                                         & 23.43                                                        & 9.79                                                         &  &  & 4.44±1.03                                               & 3.93              & 18.62                                                        & 41.23                                                        & 19.34                                                        &  &  & 8.35±1.36                                               & 7.54              & 18.67                                                        & 51.05                                                        & 20.95                                                        &  &  & 5.86±1.65                                               & 5.03              & 26.28                                                        & 61.63                                                        & 24.60                                                        &  &  & 5.49±1.22                                               & 4.66              & 28.57                                                        & 35.40                                                        & 18.65                                                        \\
                                                                          & ViT-B                  & SBD                   &  &  & 1.53±0.22                                               & 1.54              & 4.70                                                         & 12.83                                                        & 6.11                                                         &  &  & 2.61±0.46                                               & 2.46              & 9.90                                                         & 23.97                                                        & 13.03                                                        &  &  & 5.77±0.58                                               & 5.48              & 8.44                                                         & 25.72                                                        & 10.62                                                        &  &  & 4.52±1.07                                               & 4.07              & 17.58                                                        & 36.18                                                        & 15.13                                                        &  &  & 3.75±0.54                                               & 3.42              & 11.42                                                        & 19.50                                                        & 8.77                                                         \\
                                                                          & ViT-L                  & SBD                   &  &  & 1.55±0.22                                               & 1.46              & 8.91                                                         & 18.74                                                        & 5.35                                                         &  &  & 2.36±0.42                                               & 2.33              & 6.70                                                         & 13.55                                                        & 5.81                                                         &  &  & 5.56±0.53                                               & 5.38              & 7.26                                                         & 15.97                                                        & 8.05                                                         &  &  & 3.83±0.88                                               & 3.75              & 10.02                                                        & 33.06                                                        & 14.00                                                        &  &  & 3.40±0.43                                               & 3.24              & 7.32                                                         & 16.15                                                        & 8.01                                                         \\
                                                                          & ViT-H                  & SBD                   &  &  & 1.52±0.18                                               & \textit{1.44}     & 6.13                                                         & \textbf{1.58}                                                & \textbf{1.14}                                                &  &  & 2.22±0.37                                               & 2.09              & 9.44                                                         & 16.69                                                        & 8.82                                                         &  &  & 5.49±0.55                                               & 5.33              & 7.67                                                         & 23.69                                                        & 12.20                                                        &  &  & 3.74±0.86                                               & 3.59              & 10.46                                                        & 31.97                                                        & 13.17                                                        &  &  & 3.32±0.43                                               & 3.13              & 7.37                                                         & 16.88                                                        & 7.55                                                         \\ \cline{1-3} \cline{6-10} \cline{13-17} \cline{20-24} \cline{27-31} \cline{34-38} 
CFR-ICL \citep{sun2023cfricl}                            & ViT-H                  & C+L                   &  &  & 1.49±0.32                                               & 1.58              & \underline{0.32}                                            & 24.23                                                        & 11.26                                                        &  &  & \textbf{1.57±0.27}                                      & \textbf{1.46}     & 6.11                                                         & 23.58                                                        & 7.69                                                         &  &  & \textit{4.53±0.46}                                      & \textbf{4.24}     & 9.32                                                         & 18.47                                                        & 9.21                                                         &  &  & \textit{2.70±0.63}                                      & \textit{2.70}     & 9.58                                                         & 24.13                                                        & 12.83                                                        &  &  & \textbf{2.12±0.34}                                      & \textbf{2.01}     & 8.76                                                         & 14.33                                                        & 8.68                                                         \\ \cline{1-3} \cline{6-10} \cline{13-17} \cline{20-24} \cline{27-31} \cline{34-38} 
MobileSAM \citep{zhang2023faster}                        & ViT-Tiny               & SA-1B                 &  &  & 1.81±0.32                                               & 1.70              & 9.91                                                         & 11.50                                                        & 8.92                                                         &  &  & 2.80±0.58                                               & 2.73              & 11.94                                                        & 18.04                                                        & 11.88                                                        &  &  & 5.96±0.56                                               & 5.92              & 8.63                                                         & 15.39                                                        & 9.27                                                         &  &  & 5.25±0.78                                               & 5.24              & 9.79                                                         & 19.78                                                        & 9.43                                                         &  &  & 3.42±0.48                                               & 3.38              & 7.47                                                         & 12.69                                                        & 8.19                                                         \\ \cline{1-3} \cline{6-10} \cline{13-17} \cline{20-24} \cline{27-31} \cline{34-38} 
\multirow{3}{*}{SAM \citep{kirillov2023segment}}         & ViT-B                  & SA-1B                 &  &  & 1.91±0.38                                               & 1.88              & 6.67                                                         & 9.42                                                         & 17.86                                                        &  &  & 2.21±0.44                                               & 2.15              & 10.50                                                        & 11.35                                                        & 9.32                                                         &  &  & 5.30±0.53                                               & 5.15              & 8.26                                                         & 11.27                                                        & \textit{5.17}                                                &  &  & 4.91±0.79                                               & 4.86              & 9.88                                                         & 15.73                                                        & 9.69                                                         &  &  & 3.04±0.51                                               & 2.97              & 11.17                                                        & 10.06                                                        & 8.72                                                         \\
                                                                          & ViT-L                  & SA-1B                 &  &  & 1.70±0.29                                               & 1.64              & 5.06                                                         & \underline{4.79}                                            & 6.25                                                         &  &  & 1.89±0.32                                               & 1.89              & 5.58                                                         & 12.82                                                        & 8.52                                                         &  &  & 5.21±0.41                                               & 5.08              & 8.82                                                         & 11.59                                                        & 6.96                                                         &  &  & 4.81±0.63                                               & 4.77              & 8.89                                                         & 14.97                                                        & 7.96                                                         &  &  & 2.60±0.40                                               & 2.56              & 8.11                                                         & 7.08                                                         & 6.75                                                         \\
                                                                          & ViT-H                  & SA-1B                 &  &  & 1.82±0.28                                               & 1.78              & 3.92                                                         & 5.25                                                         & 7.38                                                         &  &  & 2.01±0.43                                               & 2.08              & 5.18                                                         & 10.60                                                        & 8.58                                                         &  &  & 5.42±0.49                                               & 5.32              & 8.00                                                         & 15.02                                                        & 7.89                                                         &  &  & 5.14±0.68                                               & 5.14              & 7.63                                                         & 15.61                                                        & 9.01                                                         &  &  & 2.66±0.38                                               & 2.66              & \textbf{5.95}                                                & 8.50                                                         & 6.76                                                         \\ \cline{1-3} \cline{6-10} \cline{13-17} \cline{20-24} \cline{27-31} \cline{34-38} 
\multirow{3}{*}{SAM-HQ \citep{ke2023segment}}            & ViT-B                  & SA-1B                 &  &  & 2.10±0.56                                               & 1.96              & 16.99                                                        & \textit{4.90}                                                & 6.12                                                         &  &  & 2.20±0.45                                               & 2.21              & 5.93                                                         & 16.78                                                        & 12.61                                                        &  &  & 5.32±0.50                                               & 5.26              & 7.45                                                         & 13.03                                                        & 6.08                                                         &  &  & 5.39±0.89                                               & 5.12              & 12.48                                                        & 16.68                                                        & 10.58                                                        &  &  & 3.35±0.67                                               & 3.13              & 16.41                                                        & 10.74                                                        & 8.46                                                         \\
                                                                          & ViT-L                  & SA-1B                 &  &  & 2.03±0.48                                               & 1.80              & 12.20                                                        & 15.93                                                        & 9.05                                                         &  &  & 2.03±0.38                                               & 2.11              & 6.78                                                         & \underline{5.86}                                            & 7.40                                                         &  &  & 5.19±0.48                                               & 5.08              & 8.58                                                         & 15.69                                                        & 9.05                                                         &  &  & 5.05±0.74                                               & 4.89              & 9.64                                                         & 13.50                                                        & 7.51                                                         &  &  & 2.81±0.51                                               & 2.72              & 11.02                                                        & 7.69                                                         & 6.87                                                         \\
                                                                          & ViT-H                  & SA-1B                 &  &  & 1.92±0.32                                               & 1.86              & 4.79                                                         & 17.70                                                        & 10.91                                                        &  &  & 1.99±0.31                                               & 2.10              & 4.60                                                         & \textbf{-0.10}                                               & \textbf{4.53}                                                &  &  & 5.16±0.44                                               & 5.06              & 8.15                                                         & 18.36                                                        & 9.50                                                         &  &  & 4.97±0.68                                               & 4.92              & 7.71                                                         & 12.36                                                        & 6.83                                                         &  &  & 2.75±0.41                                               & 2.72              & \textit{6.78}                                                & 7.95                                                         & 5.64                                                         \\ \cline{1-3} \cline{6-10} \cline{13-17} \cline{20-24} \cline{27-31} \cline{34-38} 
\multirow{4}{*}{SAM 2 \citep{ravi2024sam}}               & Hiera-T                & SA-V                  &  &  & 1.60±0.23                                               & 1.52              & 7.61                                                         & 11.95                                                        & 8.46                                                         &  &  & 1.88±0.27                                               & 1.92              & 6.48                                                         & 13.43                                                        & 8.68                                                         &  &  & 4.65±0.28                                               & 4.59              & \textbf{4.86}                                                & \textbf{7.46}                                                & \textbf{4.33}                                                &  &  & 3.86±0.64                                               & 3.76              & 7.79                                                         & 13.14                                                        & 7.83                                                         &  &  & 3.11±0.50                                               & 3.00              & 9.45                                                         & \underline{3.57}                                            & \textit{4.47}                                                \\
                                                                          & Hiera-S                & SA-V                  &  &  & 1.65±0.24                                               & 1.60              & 5.53                                                         & 7.67                                                         & 8.02                                                         &  &  & 1.80±0.26                                               & 1.90              & \underline{3.77}                                            & 7.14                                                         & 6.20                                                         &  &  & 4.67±0.33                                               & 4.56              & 8.49                                                         & 15.86                                                        & 7.62                                                         &  &  & 3.75±0.61                                               & 3.70              & 7.44                                                         & 12.67                                                        & 8.69                                                         &  &  & 3.02±0.47                                               & 2.91              & 9.51                                                         & 4.79                                                         & 4.64                                                         \\
                                                                          & Hiera-B+               & SA-V                  &  &  & 1.59±0.24                                               & 1.50              & 8.27                                                         & 20.29                                                        & 13.70                                                        &  &  & 1.69±0.22                                               & 1.70              & 5.13                                                         & 9.90                                                         & \underline{4.73}                                            &  &  & 4.61±0.29                                               & 4.49              & 9.51                                                         & 13.28                                                        & 7.24                                                         &  &  & 3.84±0.62                                               & 3.75              & 9.12                                                         & 12.35                                                        & 7.05                                                         &  &  & 2.83±0.41                                               & 2.76              & 7.46                                                         & \textit{4.10}                                                & 4.82                                                         \\
                                                                          & Hiera-L                & SA-V                  &  &  & \textbf{1.40±0.17}                                      & \textbf{1.36}     & 4.06                                                         & 14.63                                                        & 6.15                                                         &  &  & \underline{1.60±0.17}                                  & \textit{1.60}     & \textbf{2.77}                                                & 8.58                                                         & 6.71                                                         &  &  & \textbf{4.39±0.23}                                      & \textit{4.30}     & 7.55                                                         & 10.03                                                        & \underline{4.34}                                            &  &  & 3.42±0.51                                               & 3.40              & \textbf{6.12}                                                & \textbf{9.34}                                                & \textbf{5.61}                                                &  &  & 2.74±0.38                                               & 2.69              & \underline{6.51}                                            & 4.87                                                         & 4.68                                                         \\ \cline{1-3} \cline{6-10} \cline{13-17} \cline{20-24} \cline{27-31} \cline{34-38} 
\multirow{4}{*}{SAM 2.1 \citep{ravi2024sam}}             & Hiera-T                & SA-V                  &  &  & 1.60±0.21                                               & 1.54              & 9.06                                                         & 21.50                                                        & 10.29                                                        &  &  & 1.89±0.31                                               & 1.90              & 9.38                                                         & 16.70                                                        & 11.08                                                        &  &  & 4.67±0.32                                               & 4.58              & 7.08                                                         & \underline{8.99}                                            & 6.44                                                         &  &  & 3.91±0.68                                               & 3.83              & 8.45                                                         & \textit{11.88}                                               & 7.66                                                         &  &  & 3.11±0.50                                               & 2.99              & 9.75                                                         & \textbf{3.35}                                                & \underline{4.45}                                            \\
                                                                          & Hiera-S                & SA-V                  &  &  & 1.64±0.22                                               & 1.50              & 11.73                                                        & 9.20                                                         & \textit{3.27}                                                &  &  & 1.76±0.28                                               & 1.71              & 8.59                                                         & 7.47                                                         & 5.95                                                         &  &  & 4.63±0.32                                               & 4.51              & 9.72                                                         & 14.30                                                        & 6.82                                                         &  &  & 3.76±0.62                                               & 3.69              & 8.16                                                         & 12.35                                                        & \textit{6.43}                                                &  &  & 3.04±0.49                                               & 2.91              & 9.59                                                         & 4.70                                                         & 4.76                                                         \\
                                                                          & Hiera-B+               & SA-V                  &  &  & 1.69±0.27                                               & 1.60              & 5.75                                                         & 28.50                                                        & 13.74                                                        &  &  & 1.72±0.30                                               & 1.66              & 10.55                                                        & 10.24                                                        & 8.53                                                         &  &  & 4.67±0.32                                               & 4.53              & 11.75                                                        & 15.39                                                        & 8.73                                                         &  &  & 3.88±0.62                                               & 3.87              & 7.47                                                         & 11.95                                                        & 7.37                                                         &  &  & 2.87±0.43                                               & 2.78              & 8.35                                                         & 4.51                                                         & 5.01                                                         \\
                                                                          & Hiera-L                & SA-V                  &  &  & 1.58±0.25                                               & 1.52              & 7.07                                                         & 21.50                                                        & 9.87                                                         &  &  & \textit{1.62±0.20}                                      & \underline{1.60} & 4.90                                                         & \textit{6.17}                                                & \textit{5.25}                                                &  &  & \underline{4.44±0.25}                                  & \underline{4.28} & 10.35                                                        & \textit{9.48}                                                & 5.50                                                         &  &  & 3.51±0.52                                               & 3.44              & \textit{6.78}                                                & \underline{9.91}                                            & \underline{5.65}                                            &  &  & 2.81±0.39                                               & 2.73              & 7.41                                                         & 4.50                                                         & \textbf{4.27}                                               

\\
\bottomrule
\end{tabular}
\end{sidewaystable}

\begin{sidewaystable}
\centering
\caption{NoF evaluation results.}
\label{tab:full_nof}
\fontsize{7pt}{12pt}\selectfont
\tabcolsep=1.5pt
\begin{tabular}{cccc|ccccccc|ccccccc|ccccccc|ccccccc|cccccc}
\toprule
\multirow{3}{*}{Method}                                                   & \multirow{3}{*}{Model} & \multirow{3}{*}{Data} &  &  & \multicolumn{5}{c}{GrabCut \citep{grabcut}}                                                               &  &  & \multicolumn{5}{c}{Berkeley \citep{berkeley}}                                                                     &  &  & \multicolumn{5}{c}{DAVIS \citep{davis}}                                                                         &  &  & \multicolumn{5}{c}{COCO-MVal \citep{coco}}                                                                              &  &  & \multicolumn{5}{c}{TETRIS \citep{moskalenko2024tetris}}                                                           \\
                                                                          &                        &                       &  &  & \multicolumn{5}{c}{$\textrm{NoF}_{\textrm{20}}\textrm{@}\textrm{90}$}                                                      &  &  & \multicolumn{5}{c}{$\textrm{NoF}_{\textrm{20}}\textrm{@}\textrm{90}$}                                                              &  &  & \multicolumn{5}{c}{$\textrm{NoF}_{\textrm{20}}\textrm{@}\textrm{90}$}                                                            &  &  & \multicolumn{5}{c}{$\textrm{NoF}_{\textrm{20}}\textrm{@}\textrm{90}$}                                                               &  &  & \multicolumn{5}{c}{$\textrm{NoF}_{\textrm{20}}\textrm{@}\textrm{90}$}                                                              \\
                                                                          &                        &                       &  &  & \begin{tabular}[c]{@{}c@{}}Sample\\ (±std)\end{tabular} & Base           & ${\Delta}$SB     & ${\Delta}$GR & ${\Delta}$HH  &  &  & \begin{tabular}[c]{@{}c@{}}Sample\\ (±std)\end{tabular} & Base           & ${\Delta}$SB      & ${\Delta}$GR    & ${\Delta}$HH      &  &  & \begin{tabular}[c]{@{}c@{}}Sample\\ (±std)\end{tabular} & Base            & ${\Delta}$SB     & ${\Delta}$GR   & ${\Delta}$HH     &  &  & \begin{tabular}[c]{@{}c@{}}Sample\\ (±std)\end{tabular} & Base            & ${\Delta}$SB       & ${\Delta}$GR   & ${\Delta}$HH      &  &  & \begin{tabular}[c]{@{}c@{}}Sample\\ (±std)\end{tabular} & Base            & ${\Delta}$SB      & ${\Delta}$GR   & ${\Delta}$HH      \\ \cline{1-3} \cline{6-10} \cline{13-17} \cline{20-24} \cline{27-31} \cline{34-38} 
GPCIS \citep{zhou2023interactivegpcis}                   & RN50                   & C+L                   &  &  & 0.0±0.00                                                & 0              & 0.0              & 0            & 0.0           &  &  & 1.5±0.81                                                & \textit{1}     & 0.5               & 1               & 0.6               &  &  & 61.0±3.82                                               & 59              & 2.0              & 11             & 4.8              &  &  & 46.2±15.89                                              & 48              & -1.8               & 48             & 15.2              &  &  & 148.7±27.31                                             & 126             & 22.7              & 93             & 28.6              \\ \cline{1-3} \cline{6-10} \cline{13-17} \cline{20-24} \cline{27-31} \cline{34-38} 
\multirow{2}{*}{CDNet \citep{chen2021conditional}}       & RN34                   & C+L                   &  &  & 0.0±0.00                                                & 0              & 0.0              & 0            & 0.0           &  &  & \textit{0.9±0.94}                                       & \textbf{0}     & 0.9               & 3               & 1.4               &  &  & 61.8±5.10                                               & 56              & 5.8              & 16             & 6.0              &  &  & 58.0±5.85                                               & 58              & 0.0                & 15             & 6.4               &  &  & 144.6±14.85                                             & 126             & 18.6              & 48             & 21.6              \\
                                                                          & RN34                   & SBD                   &  &  & 0.6±0.49                                                & 0              & 0.6              & 1            & 0.0           &  &  & 3.4±1.91                                                & 1              & 2.4               & 5               & 2.8               &  &  & 77.6±18.11                                              & 62              & 15.6             & 59             & 23.2             &  &  & 82.3±12.81                                              & 69              & 13.3               & 35             & 13.0              &  &  & 229.3±28.18                                             & 195             & 34.3              & 95             & 38.2              \\ \cline{1-3} \cline{6-10} \cline{13-17} \cline{20-24} \cline{27-31} \cline{34-38} 
\multirow{5}{*}{RITM \citep{ritm}}                       & HR18                   & C+L                   &  &  & 0.0±0.00                                                & 0              & 0.0              & 0            & 0.0           &  &  & 3.4±1.36                                                & 2              & 1.4               & 4               & 0.8               &  &  & 57.9±3.05                                               & 54              & 3.9              & 10             & 3.8              &  &  & 40.8±7.74                                               & 40              & 0.8                & 29             & 11.6              &  &  & 199.8±11.33                                             & 176             & 23.8              & 42             & 15.2              \\
                                                                          & HR18s-IT               & C+L                   &  &  & 0.0±0.00                                                & 0              & 0.0              & 0            & 0.0           &  &  & 2.3±0.64                                                & 1              & 1.3               & 1               & 0.2               &  &  & 59.6±4.20                                               & 53              & 6.6              & 12             & 5.2              &  &  & 23.2±8.28                                               & \textbf{24}     & -0.8               & 27             & 9.2               &  &  & 171.7±14.63                                             & 150             & 21.7              & 48             & 17.8              \\
                                                                          & HR18-IT                & C+L                   &  &  & 0.0±0.00                                                & 0              & 0.0              & 0            & 0.0           &  &  & 2.3±1.10                                                & 2              & 0.3               & 2               & \textbf{-0.6}     &  &  & 54.4±2.62                                               & 52              & 2.4              & 8              & 1.6              &  &  & 20.7±7.01                                               & 29              & -8.3               & 20             & 9.0               &  &  & 167.4±14.54                                             & 150             & 17.4              & 43             & 12.8              \\
                                                                          & HR32-IT                & C+L                   &  &  & 0.0±0.00                                                & 0              & 0.0              & 0            & 0.0           &  &  & 1.7±1.00                                                & 2              & -0.3              & 2               & 1.0               &  &  & 52.5±1.86                                               & 51              & 1.5              & 6              & 3.0              &  &  & 27.1±5.70                                               & 31              & -3.9               & 19             & 3.8               &  &  & 160.4±7.96                                              & 135             & 25.4              & 16             & 1.6               \\
                                                                          & HR18-IT                & SBD                   &  &  & 0.0±0.00                                                & 0              & 0.0              & 0            & 0.0           &  &  & 4.4±0.66                                                & 3              & 1.4               & 1               & \textit{-0.4}     &  &  & 66.0±7.24                                               & 57              & 9.0              & 23             & 8.8              &  &  & 33.5±14.14                                              & 36              & -2.5               & 45             & 11.8              &  &  & 256.3±30.36                                             & 211             & 45.3              & 89             & 25.4              \\ \cline{1-3} \cline{6-10} \cline{13-17} \cline{20-24} \cline{27-31} \cline{34-38} 
\multirow{2}{*}{AdaptClick \citep{lin2024adaptiveclick}} & ViT-B                  & C+L                   &  &  & 0.0±0.00                                                & 0              & 0.0              & 0            & 0.0           &  &  & 2.4±0.66                                                & 1              & 1.4               & \textbf{-2}     & -0.4              &  &  & 50.9±0.70                                               & 50              & 0.9              & 1              & 0.6              &  &  & 23.6±5.08                                               & 29              & -5.4               & 14             & 3.2               &  &  & 109.4±6.58                                              & 97              & 12.4              & 19             & 5.2               \\
                                                                          & ViT-B                  & SBD                   &  &  & 0.0±0.00                                                & 0              & 0.0              & 0            & 0.0           &  &  & 2.1±0.54                                                & 2              & 0.1               & \underline{-1} & 0.2               &  &  & 53.3±1.90                                               & 51              & 2.3              & 7              & 2.6              &  &  & 33.4±11.09                                              & 42              & -8.6               & 31             & 10.4              &  &  & 155.7±15.24                                             & 139             & 16.7              & 50             & 12.6              \\ \cline{1-3} \cline{6-10} \cline{13-17} \cline{20-24} \cline{27-31} \cline{34-38} 
\multirow{7}{*}{SimpleClick \citep{liu2022simpleclick}}  & ViT-B                  & C+L                   &  &  & 0.0±0.00                                                & 0              & 0.0              & 0            & 0.0           &  &  & 1.2±0.40                                                & 2              & -0.8              & 0               & 0.4               &  &  & \textit{49.6±0.92}                                      & 49              & 0.6              & 2              & 0.8              &  &  & 22.0±8.99                                               & 29              & -7.0               & 27             & 9.6               &  &  & 107.6±6.74                                              & 101             & 6.6               & 22             & 8.0               \\
                                                                          & ViT-L                  & C+L                   &  &  & 0.0±0.00                                                & 0              & 0.0              & 0            & 0.0           &  &  & 1.1±0.30                                                & 2              & -0.9              & 0               & 0.2               &  &  & 49.6±1.11                                               & \textit{49}     & 0.6              & 3              & 1.2              &  &  & \textit{19.3±5.27}                                      & \textit{25}     & -5.7               & 19             & 6.6               &  &  & 95.2±5.06                                               & 89              & 6.2               & 11             & 7.2               \\
                                                                          & ViT-H                  & C+L                   &  &  & 0.0±0.00                                                & 0              & 0.0              & 0            & 0.0           &  &  & \underline{0.6±0.49}                                   & 1              & -0.4              & 0               & 0.8               &  &  & \underline{48.6±1.02}                                  & \underline{44} & 4.6              & 3              & 1.6              &  &  & \textbf{14.5±4.63}                                      & \underline{24} & -9.5               & 10             & 6.6               &  &  & \underline{85.6±5.10}                                  & \underline{81} & 4.6               & 13             & 6.8               \\
                                                                          & ViT-XT                 & SBD                   &  &  & 0.0±0.00                                                & 0              & 0.0              & 0            & 0.0           &  &  & 3.0±1.00                                                & 3              & 0.0               & 2               & 1.2               &  &  & 63.5±9.72                                               & 57              & 6.5              & 33             & 10.2             &  &  & 42.4±23.45                                              & 40              & 2.4                & 72             & 21.6              &  &  & 203.1±43.73                                             & 163             & 40.1              & 148            & 41.8              \\
                                                                          & ViT-B                  & SBD                   &  &  & 0.0±0.00                                                & 0              & 0.0              & 0            & 0.0           &  &  & 2.8±1.17                                                & 3              & -0.2              & 3               & 1.2               &  &  & 56.2±2.82                                               & 54              & 2.2              & 9              & 3.2              &  &  & 33.1±14.65                                              & 39              & -5.9               & 40             & 13.0              &  &  & 164.7±14.23                                             & 144             & 20.7              & 48             & 16.2              \\
                                                                          & ViT-L                  & SBD                   &  &  & 0.0±0.00                                                & 0              & 0.0              & 0            & 0.0           &  &  & 2.3±0.64                                                & 2              & 0.3               & 0               & \underline{-0.6} &  &  & 55.7±2.10                                               & 54              & 1.7              & 6              & 2.6              &  &  & 22.8±7.56                                               & 38              & \textbf{-15.2}     & 26             & 5.6               &  &  & 155.8±6.45                                              & 139             & 16.8              & 23             & 5.6               \\
                                                                          & ViT-H                  & SBD                   &  &  & 0.0±0.00                                                & 0              & 0.0              & 0            & 0.0           &  &  & 1.0±0.00                                                & 1              & 0.0               & \textit{0}      & 0.0               &  &  & 53.5±0.81                                               & 52              & 1.5              & 1              & \textbf{-0.6}    &  &  & 20.9±6.80                                               & 31              & \textit{-10.1}     & 27             & 8.6               &  &  & 146.9±8.25                                              & 136             & 10.9              & 28             & 9.0               \\ \cline{1-3} \cline{6-10} \cline{13-17} \cline{20-24} \cline{27-31} \cline{34-38} 
CFR-ICL \citep{sun2023cfricl}                            & ViT-H                  & C+L                   &  &  & 0.0±0.00                                                & 0              & 0.0              & 0            & 0.0           &  &  & \textbf{0.2±0.40}                                       & \underline{0} & 0.2               & 1               & 0.4               &  &  & \textbf{41.1±2.21}                                      & \textbf{35}     & 6.1              & 6              & 3.0              &  &  & \underline{15.1±6.62}                                  & 26              & \underline{-10.9} & 17             & 7.0               &  &  & \textbf{62.4±4.29}                                      & \textbf{58}     & 4.4               & 15             & 5.6               \\ \cline{1-3} \cline{6-10} \cline{13-17} \cline{20-24} \cline{27-31} \cline{34-38} 
MobileSAM \citep{zhang2023faster}                        & ViT-Tiny               & SA-1B                 &  &  & 0.1±0.30                                                & 0              & 0.1              & 0            & 0.2           &  &  & 2.9±0.70                                                & 3              & -0.1              & 0               & 0.2               &  &  & 65.3±1.00                                               & 67              & \textbf{-1.7}    & 1              & 1.0              &  &  & 92.0±5.12                                               & 101             & -9.0               & 15             & 1.2               &  &  & 181.4±5.82                                              & 178             & 3.4               & 13             & 1.2               \\ \cline{1-3} \cline{6-10} \cline{13-17} \cline{20-24} \cline{27-31} \cline{34-38} 
\multirow{3}{*}{SAM \citep{kirillov2023segment}}         & ViT-B                  & SA-1B                 &  &  & 0.0±0.00                                                & 0              & 0.0              & 0            & 0.0           &  &  & 0.9±0.30                                                & 1              & -0.1              & 1               & 0.2               &  &  & 53.0±1.48                                               & 51              & 2.0              & 4              & 0.8              &  &  & 80.1±2.51                                               & 88              & -7.9               & \textbf{0}     & \underline{-1.4} &  &  & 111.4±1.91                                              & 113             & \textbf{-1.6}     & \underline{2} & \textit{-0.4}     \\
                                                                          & ViT-L                  & SA-1B                 &  &  & 0.0±0.00                                                & 0              & 0.0              & 0            & 0.0           &  &  & 1.0±0.00                                                & 1              & 0.0               & 0               & 0.0               &  &  & 53.1±0.83                                               & 53              & 0.1              & 2              & 1.0              &  &  & 89.9±4.01                                               & 92              & -2.1               & 12             & 3.4               &  &  & 99.2±2.36                                               & 99              & \textit{0.2}      & 5              & -0.4              \\
                                                                          & ViT-H                  & SA-1B                 &  &  & 0.0±0.00                                                & 0              & 0.0              & 0            & 0.0           &  &  & 0.9±0.54                                                & 2              & \textbf{-1.1}     & 0               & 0.2               &  &  & 55.7±1.79                                               & 55              & 0.7              & 5              & 2.6              &  &  & 106.5±4.08                                              & 109             & -2.5               & 9              & \textbf{-1.4}     &  &  & 113.0±1.61                                              & 111             & 2.0               & \textit{3}     & 0.4               \\ \cline{1-3} \cline{6-10} \cline{13-17} \cline{20-24} \cline{27-31} \cline{34-38} 
\multirow{3}{*}{SAM-HQ \citep{ke2023segment}}            & ViT-B                  & SA-1B                 &  &  & 0.1±0.30                                                & 0              & 0.1              & \textbf{-1}  & \textbf{-0.2} &  &  & 1.0±0.00                                                & 1              & 0.0               & 0               & 0.0               &  &  & 52.7±1.27                                               & 52              & 0.7              & 3              & 1.0              &  &  & 92.6±6.09                                               & 88              & 4.6                & 8              & 0.8               &  &  & 104.0±6.75                                              & 97              & 7.0               & 15             & 6.4               \\
                                                                          & ViT-L                  & SA-1B                 &  &  & 0.6±0.49                                                & 0              & 0.6              & 0            & 0.4           &  &  & 1.1±0.30                                                & 2              & \textit{-0.9}     & 1               & 0.2               &  &  & 52.6±0.66                                               & 52              & 0.6              & \textbf{-1}    & 0.0              &  &  & 98.6±6.23                                               & 94              & 4.6                & 4              & 1.6               &  &  & 90.9±7.12                                               & 88              & 2.9               & \textbf{-7}    & \textbf{-1.8}     \\
                                                                          & ViT-H                  & SA-1B                 &  &  & 0.0±0.00                                                & 0              & 0.0              & 0            & 0.0           &  &  & 1.0±0.00                                                & 2              & \underline{-1.0} & 0               & 0.0               &  &  & 53.5±0.67                                               & 53              & 0.5              & 2              & 0.6              &  &  & 95.1±3.73                                               & 102             & -6.9               & 5              & \textit{-0.6}     &  &  & 93.9±1.92                                               & 93              & 0.9               & 4              & 1.8               \\ \cline{1-3} \cline{6-10} \cline{13-17} \cline{20-24} \cline{27-31} \cline{34-38} 
\multirow{4}{*}{SAM 2 \citep{ravi2024sam}}               & Hiera-T                & SA-V                  &  &  & 0.0±0.00                                                & 0              & 0.0              & 0            & 0.0           &  &  & 1.0±0.00                                                & 1              & 0.0               & 0               & 0.0               &  &  & 51.0±0.00                                               & 51              & \textit{0.0}     & 0              & 0.0              &  &  & 42.7±3.03                                               & 48              & -5.3               & 7              & 0.6               &  &  & 93.8±3.89                                               & 92              & 1.8               & 12             & 3.6               \\
                                                                          & Hiera-S                & SA-V                  &  &  & 0.0±0.00                                                & 0              & 0.0              & 0            & 0.0           &  &  & 1.0±0.00                                                & 1              & 0.0               & 0               & 0.0               &  &  & 50.0±0.00                                               & 50              & 0.0              & 0              & 0.0              &  &  & 39.5±4.18                                               & 45              & -5.5               & \underline{2} & -0.2              &  &  & 90.6±3.41                                               & 89              & 1.6               & 14             & 2.8               \\
                                                                          & Hiera-B+               & SA-V                  &  &  & 0.0±0.00                                                & 0              & 0.0              & 0            & 0.0           &  &  & 1.0±0.00                                                & 1              & 0.0               & 0               & 0.0               &  &  & 50.0±0.00                                               & 50              & 0.0              & 0              & 0.0              &  &  & 50.6±2.80                                               & 54              & -3.4               & 6              & 1.6               &  &  & 93.7±2.15                                               & 93              & 0.7               & 5              & 2.2               \\
                                                                          & Hiera-L                & SA-V                  &  &  & 0.0±0.00                                                & 0              & 0.0              & 0            & 0.0           &  &  & 1.0±0.00                                                & 1              & 0.0               & 0               & 0.0               &  &  & 50.0±0.00                                               & 50              & 0.0              & 0              & \textit{0.0}     &  &  & 40.3±2.93                                               & 47              & -6.7               & 7              & 1.8               &  &  & 88.9±2.51                                               & \textit{87}     & 1.9               & 5              & \underline{-1.4} \\ \cline{1-3} \cline{6-10} \cline{13-17} \cline{20-24} \cline{27-31} \cline{34-38} 
\multirow{4}{*}{SAM 2.1 \citep{ravi2024sam}}             & Hiera-T                & SA-V                  &  &  & \textit{0.0±0.00}                                       & \textit{0}     & \textit{0.0}     & 0            & 0.0           &  &  & 1.0±0.00                                                & 1              & 0.0               & 0               & 0.0               &  &  & 51.1±0.30                                               & 51              & 0.1              & 1              & 0.2              &  &  & 41.3±3.10                                               & 49              & -7.7               & 7              & 1.4               &  &  & 94.8±3.63                                               & 92              & 2.8               & 10             & 4.8               \\
                                                                          & Hiera-S                & SA-V                  &  &  & 0.0±0.00                                   & 0 & 0.0 & 0            & 0.0           &  &  & 1.0±0.00                                                & 1              & 0.0               & 0               & 0.0               &  &  & 50.0±0.00                                               & 50              & 0.0              & \textit{0}     & 0.0 &  &  & 38.1±2.84                                               & 41              & -2.9               & 6              & 2.6               &  &  & 93.4±2.91                                               & 91              & 2.4               & 8              & 0.8               \\
                                                                          & Hiera-B+               & SA-V                  &  &  & 0.0±0.00                                                & 0              & 0.0              & 0            & 0.0           &  &  & 0.9±0.30                                                & 1              & -0.1              & 0               & 0.2               &  &  & 50.0±0.00                                               & 50              & 0.0              & 0              & 0.0              &  &  & 48.1±2.21                                               & 57              & -8.9               & \textit{3}     & 0.2               &  &  & 90.8±2.52                                               & 88              & 2.8               & 4              & 1.2               \\
                                                                          & Hiera-L                & SA-V                  &  &  & 0.0±0.00                                                & 0              & 0.0              & 0            & 0.0           &  &  & 1.0±0.00                                                & 1              & 0.0               & 0               & 0.0               &  &  & 50.0±0.00                                               & 50              & 0.0 & \underline{0} & 0.0              &  &  & 40.3±2.97                                               & 44              & -3.7               & 8              & 2.2               &  &  & \textit{88.7±2.76}                                      & 89              & \underline{-0.3} & 8              & 3.4         \\ \bottomrule     
\end{tabular}
\end{sidewaystable}

\begin{sidewaystable}
\centering
\caption{IoU-AuC evaluation results}
\label{tab:full_iou}
\fontsize{7pt}{12pt}\selectfont
\tabcolsep=1pt
\begin{tabular}{cccc|ccccccc|ccccccc|ccccccc|ccccccc|cccccc}
\toprule
\multirow{3}{*}{Method}                                                   & \multirow{3}{*}{Model} & \multirow{3}{*}{Data} &  &  & \multicolumn{5}{c}{GrabCut \citep{grabcut}}                                                                             &  &  & \multicolumn{5}{c}{Berkeley \citep{berkeley}}                                                                           &  &  & \multicolumn{5}{c}{DAVIS \citep{davis}}                                                                                    &  &  & \multicolumn{5}{c}{COCO-MVal \citep{coco}}                                                                                    &  &  & \multicolumn{5}{c}{TETRIS \citep{moskalenko2024tetris}}                                                                  \\
                                                                          &                        &                       &  &  & \multicolumn{5}{c}{$\textrm{IoU-AuC}_{\textrm{20}}$}                                                                                     &  &  & \multicolumn{5}{c}{$\textrm{IoU-AuC}_{\textrm{20}}$}                                                                                     &  &  & \multicolumn{5}{c}{$\textrm{IoU-AuC}_{\textrm{20}}$}                                                                                        &  &  & \multicolumn{5}{c}{$\textrm{IoU-AuC}_{\textrm{20}}$}                                                                                      &  &  & \multicolumn{5}{c}{$\textrm{IoU-AuC}_{\textrm{20}}$}                                                                                      \\
                                                                          &                        &                       &  &  & \begin{tabular}[c]{@{}c@{}}Sample\\ (±std)\end{tabular} & Base               & ${\Delta}$SB      & ${\Delta}$GR      & ${\Delta}$HH      &  &  & \begin{tabular}[c]{@{}c@{}}Sample\\ (±std)\end{tabular} & Base               & ${\Delta}$SB      & ${\Delta}$GR      & ${\Delta}$HH      &  &  & \begin{tabular}[c]{@{}c@{}}Sample\\ (±std)\end{tabular} & Base               & ${\Delta}$SB       & ${\Delta}$GR       & ${\Delta}$HH       &  &  & \begin{tabular}[c]{@{}c@{}}Sample\\ (±std)\end{tabular} & Base               & ${\Delta}$SB       & ${\Delta}$GR      & ${\Delta}$HH      &  &  & \begin{tabular}[c]{@{}c@{}}Sample\\ (±std)\end{tabular} & Base               & ${\Delta}$SB       & ${\Delta}$GR      & ${\Delta}$HH      \\ \cline{1-3} \cline{6-10} \cline{13-17} \cline{20-24} \cline{27-31} \cline{34-38} 
GPCIS \citep{zhou2023interactivegpcis}                   & RN50                   & C+L                   &  &  & 97.16±0.83                                              & 97.29              & -0.13             & -1.71             & -0.56             &  &  & 94.62±0.92                                              & 95.17              & -0.55             & -1.73             & -0.61             &  &  & 89.30±1.27                                              & 90.39              & -1.09              & -2.86              & -1.08              &  &  & 90.89±2.91                                              & 92.62              & -1.73              & -6.48             & -2.30             &  &  & 92.73±1.83                                              & 93.98              & -1.25              & -3.69             & -1.29             \\ \cline{1-3} \cline{6-10} \cline{13-17} \cline{20-24} \cline{27-31} \cline{34-38} 
\multirow{2}{*}{CDNet \citep{chen2021conditional}}       & RN34                   & C+L                   &  &  & 96.58±1.59                                              & 97.14              & -0.56             & -3.80             & -1.07             &  &  & 93.87±1.80                                              & 94.78              & -0.91             & -3.70             & -1.39             &  &  & 88.05±2.05                                              & 89.71              & -1.65              & -5.18              & -1.92              &  &  & 90.74±2.17                                              & 91.81              & -1.08              & -3.57             & -1.43             &  &  & 92.98±1.97                                              & 94.19              & -1.22              & -3.97             & -1.54             \\
                                                                          & RN34                   & SBD                   &  &  & 95.44±1.25                                              & 95.90              & -0.47             & -2.99             & -1.12             &  &  & 92.95±1.36                                              & 93.77              & -0.82             & -3.59             & -1.29             &  &  & 86.59±2.07                                              & 88.21              & -1.62              & -5.97              & -2.10              &  &  & 87.10±2.89                                              & 88.14              & -1.05              & -4.73             & -1.92             &  &  & 90.52±2.12                                              & 91.49              & -0.97              & -4.66             & -1.76             \\ \cline{1-3} \cline{6-10} \cline{13-17} \cline{20-24} \cline{27-31} \cline{34-38} 
\multirow{5}{*}{RITM \citep{ritm}}                       & HR18                   & C+L                   &  &  & 96.18±1.22                                              & 96.32              & -0.15             & -1.01             & -0.45             &  &  & 93.22±1.38                                              & 93.92              & -0.69             & -1.73             & -0.69             &  &  & 87.75±1.65                                              & 88.62              & -0.87              & -2.40              & -0.81              &  &  & 91.05±2.04                                              & 91.99              & -0.94              & -2.53             & -0.95             &  &  & 91.77±1.68                                              & 92.45              & -0.68              & -1.76             & -0.70             \\
                                                                          & HR18s-IT               & C+L                   &  &  & 97.27±0.76                                              & 97.15              & 0.12              & -0.77             & -0.38             &  &  & 94.61±0.82                                              & 94.90              & -0.28             & -0.75             & -0.33             &  &  & 88.19±1.96                                              & 89.72              & -1.52              & -3.05              & -1.02              &  &  & 93.00±1.64                                              & 93.74              & -0.74              & -1.77             & -0.59             &  &  & 93.11±1.26                                              & 93.88              & -0.77              & -1.40             & -0.45             \\
                                                                          & HR18-IT                & C+L                   &  &  & 97.24±0.76                                              & 97.33              & -0.09             & -0.76             & -0.32             &  &  & 95.12±0.71                                              & 95.40              & -0.27             & -1.05             & -0.25             &  &  & 89.03±1.61                                              & 89.99              & -0.96              & -2.67              & -0.90              &  &  & 93.60±1.54                                              & 94.17              & -0.58              & -2.09             & -0.73             &  &  & 93.49±1.19                                              & 94.20              & -0.71              & -1.29             & -0.43             \\
                                                                          & HR32-IT                & C+L                   &  &  & 97.39±0.88                                              & 97.26              & 0.12              & -1.07             & -0.24             &  &  & 95.16±0.83                                              & 95.54              & -0.37             & -0.96             & -0.38             &  &  & 89.32±1.89                                              & 90.67              & -1.36              & -3.34              & -1.14              &  &  & 92.95±1.64                                              & 93.90              & -0.95              & -1.38             & -0.57             &  &  & 93.33±1.53                                              & 94.43              & -1.10              & -1.25             & -0.21             \\
                                                                          & HR18-IT                & SBD                   &  &  & 96.36±0.69                                              & 96.38              & -0.01             & -0.57             & -0.14             &  &  & 93.69±0.85                                              & 94.01              & -0.32             & -1.05             & -0.28             &  &  & 87.64±1.61                                              & 88.72              & -1.08              & -2.56              & -0.98              &  &  & 89.87±2.97                                              & 91.48              & -1.60              & -4.75             & -1.55             &  &  & 89.69±2.05                                              & 91.27              & -1.58              & -2.77             & -0.82             \\ \cline{1-3} \cline{6-10} \cline{13-17} \cline{20-24} \cline{27-31} \cline{34-38} 
\multirow{2}{*}{AdaptClick \citep{lin2024adaptiveclick}} & ViT-B                  & C+L                   &  &  & 98.27±0.37                                              & 98.11              & 0.15              & -0.82             & -0.25             &  &  & 95.69±0.62                                              & 95.74              & -0.05             & -0.95             & -0.32             &  &  & 91.30±1.05                                              & 92.05              & -0.74              & -1.91              & -0.72              &  &  & 93.97±1.18                                              & 94.30              & -0.33              & -1.58             & -0.52             &  &  & 95.07±0.82                                              & 95.44              & -0.37              & -0.67             & -0.22             \\
                                                                          & ViT-B                  & SBD                   &  &  & 97.94±0.34                                              & 97.73              & \underline{0.21} & -0.45             & -0.10             &  &  & 94.97±0.59                                              & 95.24              & -0.27             & -0.77             & -0.38             &  &  & 90.54±0.88                                              & 91.21              & -0.67              & -1.45              & -0.59              &  &  & 90.61±2.86                                              & 91.43              & -0.82              & -4.59             & -1.54             &  &  & 92.61±1.52                                              & 93.26              & -0.65              & -2.10             & -0.67             \\ \cline{1-3} \cline{6-10} \cline{13-17} \cline{20-24} \cline{27-31} \cline{34-38} 
\multirow{7}{*}{SimpleClick \citep{liu2022simpleclick}}  & ViT-B                  & C+L                   &  &  & 98.17±0.44                                              & 98.12              & 0.05              & -0.98             & -0.21             &  &  & 95.86±0.56                                              & 95.99              & -0.13             & -0.93             & -0.34             &  &  & 91.00±1.05                                              & 91.51              & -0.52              & -2.08              & -0.78              &  &  & 93.95±1.42                                              & 94.42              & -0.47              & -1.98             & -0.75             &  &  & 95.23±0.83                                              & 95.63              & -0.41              & -0.82             & -0.30             \\
                                                                          & ViT-L                  & C+L                   &  &  & \textit{98.29±0.52}                                     & \underline{98.29} & 0.01              & -0.78             & -0.28             &  &  & 96.02±0.58                                              & \textit{96.08}     & -0.06             & -0.34             & -0.14             &  &  & 91.44±0.94                                              & 92.00              & -0.56              & -1.72              & -0.55              &  &  & \underline{94.58±1.10}                                 & \textit{94.79}     & -0.21              & -1.61             & -0.58             &  &  & \textit{95.62±0.76}                                     & \textit{95.90}     & -0.29              & -0.42             & -0.18             \\
                                                                          & ViT-H                  & C+L                   &  &  & \underline{98.31±0.59}                                 & 98.23              & 0.08              & -0.84             & -0.20             &  &  & \textit{96.11±0.54}                                     & \underline{96.21} & -0.10             & -0.64             & -0.29             &  &  & 91.15±1.11                                              & 91.81              & -0.65              & -1.60              & -0.59              &  &  & \textbf{94.81±1.08}                                     & \textbf{94.98}     & -0.17              & -1.58             & -0.63             &  &  & \underline{95.73±0.76}                                 & \underline{96.05} & -0.32              & -0.53             & -0.18             \\
                                                                          & ViT-XT                 & SBD                   &  &  & 96.62±0.76                                              & 96.63              & -0.01             & -1.29             & -0.52             &  &  & 93.10±1.24                                              & 93.95              & -0.84             & -2.11             & -0.91             &  &  & 86.51±2.25                                              & 87.95              & -1.44              & -3.38              & -1.37              &  &  & 88.88±3.72                                              & 91.16              & -2.28              & -6.27             & -1.87             &  &  & 89.76±2.77                                              & 91.78              & -2.02              & -3.17             & -1.05             \\
                                                                          & ViT-B                  & SBD                   &  &  & 97.74±0.32                                              & 97.59              & 0.15              & -0.48             & -0.10             &  &  & 95.01±0.56                                              & 95.27              & -0.26             & -0.90             & -0.38             &  &  & 90.12±0.87                                              & 90.81              & -0.69              & -1.65              & -0.65              &  &  & 90.05±2.80                                              & 91.91              & -1.86              & -3.59             & -1.09             &  &  & 92.20±1.47                                              & 93.30              & -1.10              & -1.46             & -0.34             \\
                                                                          & ViT-L                  & SBD                   &  &  & 97.36±0.62                                              & 97.64              & -0.28             & -0.47             & -0.00             &  &  & 95.01±0.53                                              & 95.25              & -0.25             & -0.89             & -0.30             &  &  & 90.25±0.90                                              & 90.83              & -0.57              & -1.50              & -0.55              &  &  & 92.02±1.94                                              & 92.65              & -0.63              & -3.13             & -1.01             &  &  & 93.12±1.10                                              & 93.62              & -0.50              & -1.47             & -0.49             \\
                                                                          & ViT-H                  & SBD                   &  &  & 97.62±0.39                                              & 97.70              & -0.08             & -0.15             & \textit{-0.00}    &  &  & 94.97±0.61                                              & 95.38              & -0.42             & -0.80             & -0.37             &  &  & 90.16±1.07                                              & 90.69              & -0.53              & -1.94              & -0.69              &  &  & 91.98±2.12                                              & 92.88              & -0.90              & -3.42             & -1.10             &  &  & 93.02±1.20                                              & 93.68              & -0.66              & -1.67             & -0.57             \\ \cline{1-3} \cline{6-10} \cline{13-17} \cline{20-24} \cline{27-31} \cline{34-38} 
CFR-ICL \citep{sun2023cfricl}                            & ViT-H                  & C+L                   &  &  & \textbf{98.66±0.57}                                     & \textbf{98.35}     & \textbf{0.31}     & -0.80             & -0.38             &  &  & \textbf{96.57±0.60}                                     & \textbf{96.70}     & -0.13             & -0.93             & -0.31             &  &  & \textbf{92.09±1.09}                                     & \textbf{92.71}     & -0.62              & -1.61              & -0.66              &  &  & \textit{94.54±1.26}                                     & \underline{94.84} & -0.29              & -1.71             & -0.60             &  &  & \textbf{96.25±0.84}                                     & \textbf{96.60}     & -0.36              & -0.52             & -0.19             \\ \cline{1-3}
MobileSAM \citep{zhang2023faster}                        & ViT-Tiny               & SA-1B                 &  &  & 96.90±0.98                                              & 96.92              & -0.02             & -0.17             & -0.12             &  &  & 93.39±1.36                                              & 93.56              & -0.17             & -0.11             & -0.00             &  &  & 88.10±1.56                                              & 88.79              & -0.69              & -1.69              & -0.64              &  &  & 88.64±2.58                                              & 89.26              & -0.61              & -0.79             & -0.19             &  &  & 92.61±1.38                                              & 92.84              & -0.23              & -0.49             & -0.14             \\ \cline{1-3}
\multirow{3}{*}{SAM \citep{kirillov2023segment}}         & ViT-B                  & SA-1B                 &  &  & 96.97±0.93                                              & 96.99              & -0.02             & \textit{0.29}     & -0.19             &  &  & 94.52±0.86                                              & 94.57              & -0.05             & -0.02             & -0.18             &  &  & 89.46±1.30                                              & 89.76              & -0.30              & -0.53              & -0.14              &  &  & 90.70±1.53                                              & 91.03              & -0.32              & -0.23             & -0.04             &  &  & 93.92±1.18                                              & 94.18              & -0.26              & 0.09              & 0.04              \\
                                                                          & ViT-L                  & SA-1B                 &  &  & 97.54±0.77                                              & 97.58              & -0.04             & 0.00              & -0.09             &  &  & 95.11±0.88                                              & 94.99              & 0.12              & -0.11             & -0.05             &  &  & 89.71±1.27                                              & 90.17              & -0.47              & -0.89              & -0.36              &  &  & 90.73±1.42                                              & 91.09              & -0.36              & -0.11             & 0.01              &  &  & 94.65±1.07                                              & 94.85              & -0.20              & 0.35              & 0.09              \\
                                                                          & ViT-H                  & SA-1B                 &  &  & 97.37±0.79                                              & 97.28              & 0.09              & -0.18             & -0.17             &  &  & 94.65±1.09                                              & 94.82              & -0.17             & 0.09              & -0.33             &  &  & 89.14±1.55                                              & 89.85              & -0.70              & -1.31              & -0.61              &  &  & 89.24±2.30                                              & 89.64              & -0.41              & -0.53             & -0.08             &  &  & 94.32±1.09                                              & 94.50              & \textit{-0.18}     & 0.11              & 0.03              \\ \cline{1-3} \cline{6-10} \cline{13-17} \cline{20-24} \cline{27-31} \cline{34-38} 
\multirow{3}{*}{SAM-HQ \citep{ke2023segment}}            & ViT-B                  & SA-1B                 &  &  & 96.85±1.61                                              & 97.15              & -0.30             & \textbf{2.30}     & \textbf{0.57}     &  &  & 94.87±0.99                                              & 95.06              & -0.19             & 0.11              & -0.05             &  &  & 90.07±1.31                                              & 90.31              & \textit{-0.25}     & \textbf{-0.06}     & \textbf{0.04}      &  &  & 90.44±1.62                                              & 90.86              & -0.43              & \underline{0.36} & \textit{0.15}     &  &  & 93.73±1.77                                              & 94.14              & -0.41              & \textbf{0.89}     & \textbf{0.36}     \\
                                                                          & ViT-L                  & SA-1B                 &  &  & 97.29±1.43                                              & 97.64              & -0.35             & \underline{1.93} & \underline{0.37} &  &  & 95.47±0.85                                              & 95.54              & -0.08             & \underline{0.15} & \textbf{0.15}     &  &  & 90.22±1.32                                              & 90.41              & \textbf{-0.20}     & \textit{-0.27}     & \underline{-0.06} &  &  & 90.79±1.46                                              & 91.17              & -0.39              & \textbf{0.58}     & \textbf{0.31}     &  &  & 94.85±1.36                                              & 95.00              & \underline{-0.14} & \underline{0.88} & \underline{0.31} \\
                                                                          & ViT-H                  & SA-1B                 &  &  & 97.55±0.72                                              & 97.58              & -0.03             & -0.01             & -0.17             &  &  & 95.54±0.71                                              & 95.51              & 0.03              & \textbf{0.16}     & 0.06 &  &  & 90.36±1.22                                              & 90.59              & \underline{-0.23} & \underline{-0.19} & \textit{-0.11}     &  &  & 90.81±1.49                                              & 91.17              & -0.36              & 0.12              & 0.12              &  &  & 95.04±1.22                                              & 95.14              & \textbf{-0.10}     & 0.53              & 0.17              \\ \cline{1-3} \cline{6-10} \cline{13-17} \cline{20-24} \cline{27-31} \cline{34-38} 
\multirow{4}{*}{SAM 2 \citep{ravi2024sam}}               & Hiera-T                & SA-V                  &  &  & 97.99±0.69                                              & 98.21              & -0.22             & -0.19             & -0.18             &  &  & 95.88±0.67                                              & 95.82              & 0.06              & -0.18             & -0.05             &  &  & 91.30±0.96                                              & 91.70              & -0.40              & -0.55              & -0.17              &  &  & 92.34±1.29                                              & 92.45              & -0.10              & 0.17              & \underline{0.17} &  &  & 93.97±1.43                                              & 94.38              & -0.41              & \textit{0.84}     & \textit{0.30}     \\
                                                                          & Hiera-S                & SA-V                  &  &  & 97.82±0.84                                              & 97.98              & -0.16             & -0.44             & -0.40             &  &  & 95.87±0.64                                              & 95.59              & \textbf{0.28}     & \textit{0.11}     & 0.00              &  &  & 91.16±1.02                                              & 91.49              & -0.33              & -1.04              & -0.31              &  &  & 92.38±1.26                                              & 92.47              & \textit{-0.08}     & -0.19             & -0.01             &  &  & 94.07±1.37                                              & 94.47              & -0.40              & 0.70              & 0.23              \\
                                                                          & Hiera-B+               & SA-V                  &  &  & 97.97±0.61                                              & 97.98              & -0.01             & -0.43             & -0.27             &  &  & 96.10±0.57                                              & 95.96              & \textit{0.14}     & -0.09             & 0.01              &  &  & 91.33±1.09                                              & 91.78              & -0.45              & -0.78              & -0.42              &  &  & 91.97±1.33                                              & 92.15              & -0.18              & 0.19              & 0.14              &  &  & 94.26±1.26                                              & 94.58              & -0.33              & 0.60              & 0.17              \\
                                                                          & Hiera-L                & SA-V                  &  &  & 98.22±0.55                                              & \textit{98.24}     & -0.02             & -0.25             & -0.16             &  &  & \underline{96.22±0.55}                                 & 95.96              & \underline{0.26} & -0.07             & -0.04             &  &  & \underline{92.02±0.89}                                 & \underline{92.37} & -0.35              & -0.54              & -0.18              &  &  & 93.04±1.02                                              & 93.08              & \underline{-0.04} & -0.30             & 0.03              &  &  & 94.50±1.16                                              & 94.74              & -0.24              & 0.56              & 0.16              \\ \cline{1-3} \cline{6-10} \cline{13-17} \cline{20-24} \cline{27-31} \cline{34-38} 
\multirow{4}{*}{SAM 2.1 \citep{ravi2024sam}}             & Hiera-T                & SA-V                  &  &  & 98.05±0.66                                              & 98.13              & -0.08             & -0.56             & -0.15             &  &  & 95.73±0.77                                              & 95.69              & 0.04              & -0.34             & -0.12             &  &  & 91.16±1.14                                              & 91.56              & -0.39              & -0.80              & -0.37              &  &  & 92.38±1.33                                              & 92.51              & -0.14              & -0.03             & 0.11              &  &  & 94.07±1.40                                              & 94.48              & -0.41              & 0.78              & 0.28              \\
                                                                          & Hiera-S                & SA-V                  &  &  & 98.01±0.69                                              & 98.20              & -0.19             & -0.23             & -0.07             &  &  & 95.82±0.66                                              & 95.81              & 0.01              & -0.07             & \textit{0.02}     &  &  & 91.02±1.15                                              & 91.56              & -0.54              & -1.20              & -0.35              &  &  & 92.46±1.25                                              & 92.56              & -0.11              & 0.14              & 0.12              &  &  & 93.94±1.45                                              & 94.41              & -0.48              & 0.75              & 0.27              \\
                                                                          & Hiera-B+               & SA-V                  &  &  & 97.79±0.74                                              & 97.74              & 0.05              & -0.45             & -0.22             &  &  & 95.90±0.68                                              & 95.97              & -0.07             & -0.03             & 0.02              &  &  & 91.22±1.11                                              & 91.61              & -0.39              & -0.71              & -0.25              &  &  & 92.05±1.27                                              & 92.28              & -0.22              & \textit{0.20}     & 0.15              &  &  & 94.27±1.31                                              & 94.67              & -0.40              & 0.66              & 0.23              \\
                                                                          & Hiera-L                & SA-V                  &  &  & 98.01±0.69                                              & 97.85              & \textit{0.15}     & -0.43             & -0.15             &  &  & 96.05±0.64                                              & 95.97              & 0.08              & 0.02              & -0.05             &  &  & \textit{91.92±1.01}                                     & \textit{92.31}     & -0.38              & -0.57              & -0.22              &  &  & 92.97±0.99                                              & 92.96              & \textbf{0.01}      & -0.37             & -0.07             &  &  & 94.44±1.19                                              & 94.74              & -0.29              & 0.53              & 0.15          \\
                                                                          \bottomrule
\end{tabular}
\end{sidewaystable}

\begin{sidewaystable}
\centering
\caption{Evaluation results on real-user clicks and NSR@20 for simulated clicks.}
\label{tab:full_iou_1}
\fontsize{6.5pt}{12pt}\selectfont
\tabcolsep=0.7pt
\begin{tabular}{cccc|ccccccc|ccccccc|ccccccc|ccccccc|cccccc}
\toprule
\multirow{3}{*}{Method}                                                   & \multirow{3}{*}{Model} & \multirow{3}{*}{Data} &  &  & \multicolumn{5}{c}{GrabCut \citep{grabcut}}                                                                                                                                                                                                                   &  &  & \multicolumn{5}{c}{Berkeley \citep{berkeley}}                                                                                                                                                                                                                 &  &  & \multicolumn{5}{c}{DAVIS \citep{davis}}                                                                                                                                                                                                                       &  &  & \multicolumn{5}{c}{COCO-MVal \citep{coco}}                                                                                                                                                                                                                         &  &  & \multicolumn{5}{c}{TETRIS \citep{moskalenko2024tetris}}                                                                                                                                                                                                       \\
                                                                          &                        &                       &  &  & \multicolumn{5}{c}{IoU}                                                                                                                                                                                                                                                        &  &  & \multicolumn{5}{c}{IoU}                                                                                                                                                                                                                                                        &  &  & \multicolumn{5}{c}{IoU}                                                                                                                                                                                                                                                        &  &  & \multicolumn{5}{c}{IoU}                                                                                                                                                                                                                                                        &  &  & \multicolumn{5}{c}{IoU}                                                                                                                                                                                                                                                        \\
                                                                          &                        &                       &  &  & \begin{tabular}[c]{@{}c@{}}User@1\\ (±std)\end{tabular} & \begin{tabular}[c]{@{}c@{}}Sample@1\\ (±std)\end{tabular} & \begin{tabular}[c]{@{}c@{}}Base\\ @1\end{tabular} & \begin{tabular}[c]{@{}c@{}}NSR\\ @1\end{tabular} & \begin{tabular}[c]{@{}c@{}}NSR\\ @20\end{tabular} &  &  & \begin{tabular}[c]{@{}c@{}}User@1\\ (±std)\end{tabular} & \begin{tabular}[c]{@{}c@{}}Sample@1\\ (±std)\end{tabular} & \begin{tabular}[c]{@{}c@{}}Base\\ @1\end{tabular} & \begin{tabular}[c]{@{}c@{}}NSR\\ @1\end{tabular} & \begin{tabular}[c]{@{}c@{}}NSR\\ @20\end{tabular} &  &  & \begin{tabular}[c]{@{}c@{}}User@1\\ (±std)\end{tabular} & \begin{tabular}[c]{@{}c@{}}Sample@1\\ (±std)\end{tabular} & \begin{tabular}[c]{@{}c@{}}Base\\ @1\end{tabular} & \begin{tabular}[c]{@{}c@{}}NSR\\ @1\end{tabular} & \begin{tabular}[c]{@{}c@{}}NSR\\ @20\end{tabular} &  &  & \begin{tabular}[c]{@{}c@{}}User@1\\ (±std)\end{tabular} & \begin{tabular}[c]{@{}c@{}}Sample@1\\ (±std)\end{tabular} & \begin{tabular}[c]{@{}c@{}}Base\\ @1\end{tabular} & \begin{tabular}[c]{@{}c@{}}NSR\\ @1\end{tabular} & \begin{tabular}[c]{@{}c@{}}NSR\\ @20\end{tabular} &  &  & \begin{tabular}[c]{@{}c@{}}User@1\\ (±std)\end{tabular} & \begin{tabular}[c]{@{}c@{}}Sample@1\\ (±std)\end{tabular} & \begin{tabular}[c]{@{}c@{}}Base\\ @1\end{tabular} & \begin{tabular}[c]{@{}c@{}}NSR\\ @1\end{tabular} & \begin{tabular}[c]{@{}c@{}}NSR\\ @20\end{tabular} \\ \cline{1-3} \cline{6-10} \cline{13-17} \cline{20-24} \cline{27-31} \cline{34-38} 
GPCIS \citep{zhou2023interactivegpcis}                   & RN50                   & C+L                   &  &  & 83.41±8.67                                              & 83.49±7.68                                                & 84.46                                             & 11.63                                            & 0.89                                              &  &  & 78.79±8.30                                              & 79.15±8.04                                                & 79.61                                             & 12.24                                            & 0.99                                              &  &  & 72.53±9.71                                              & 73.66±7.39                                                & 76.03                                             & 16.30                                            & 1.62                                              &  &  & 64.04±9.06                                              & 64.43±7.95                                                & 66.30                                             & 18.21                                            & 3.81                                              &  &  & 71.99±10.13                                             & 71.75±9.73                                                & 74.40                                             & 17.24                                            & 2.36                                              \\ \cline{1-3}
\multirow{2}{*}{CDNet \citep{chen2021conditional}}       & RN34                   & C+L                   &  &  & 87.85±10.22                                             & 88.27±8.19                                                & \underline{90.80}                                & 12.50                                            & 1.66                                              &  &  & 81.92±10.42                                             & 82.47±9.45                                                & \underline{84.64}                                & 14.69                                            & 1.96                                              &  &  & 71.97±11.50                                             & 73.41±8.24                                                & 76.70                                             & 19.72                                            & 2.53                                              &  &  & 76.05±10.28                                             & 76.80±8.00                                                & 79.22                                             & 16.32                                            & 2.65                                              &  &  & 79.56±11.38                                             & 79.61±10.84                                               & 82.47                                             & 17.32                                            & 2.31                                              \\
                                                                          & RN34                   & SBD                   &  &  & 85.27±10.06                                             & 84.88±9.36                                                & 88.17                                             & 13.39                                            & 1.37                                              &  &  & 81.04±9.06                                              & 81.58±7.75                                                & 83.34                                             & 13.07                                            & 1.51                                              &  &  & 69.16±11.18                                             & 70.50±7.68                                                & 72.84                                             & 19.96                                            & 2.67                                              &  &  & 65.20±9.74                                              & 65.72±8.23                                                & 68.11                                             & 19.07                                            & 4.24                                              &  &  & 73.45±10.97                                             & 73.22±10.69                                               & 76.92                                             & 18.51                                            & 2.86                                              \\ \cline{1-3}
\multirow{5}{*}{RITM \citep{ritm}}                       & HR18                   & C+L                   &  &  & 86.10±7.57                                              & 86.11±7.34                                                & 86.78                                             & 10.53                                            & 1.29                                              &  &  & 82.55±6.09                                              & 82.33±5.75                                                & 83.16                                             & 9.37                                             & 1.50                                              &  &  & 67.30±8.63                                              & 67.80±7.99                                                & 68.18                                             & 20.88                                            & 2.05                                              &  &  & 73.81±8.34                                              & 74.09±6.86                                                & 73.80                                             & 17.44                                            & 2.51                                              &  &  & 76.99±7.11                                              & 76.99±7.09                                                & 76.23                                             & 12.89                                            & 1.96                                              \\
                                                                          & HR18s-IT               & C+L                   &  &  & 83.60±10.15                                             & 83.12±9.83                                                & 86.20                                             & 13.68                                            & 0.81                                              &  &  & 77.51±9.11                                              & 77.55±8.45                                                & 78.69                                             & 13.57                                            & 0.88                                              &  &  & 66.50±11.70                                             & 67.86±8.66                                                & 70.86                                             & 24.22                                            & 2.51                                              &  &  & 71.98±10.05                                             & 72.37±8.03                                                & 73.98                                             & 19.00                                            & 1.98                                              &  &  & 73.25±9.88                                              & 73.29±9.19                                                & 75.96                                             & 16.66                                            & 1.53                                              \\
                                                                          & HR18-IT                & C+L                   &  &  & 85.85±7.59                                              & 85.25±7.28                                                & 88.33                                             & 10.18                                            & 0.82                                              &  &  & 82.08±6.48                                              & 82.43±5.20                                                & 83.37                                             & 9.42                                             & 0.77                                              &  &  & 68.72±10.38                                             & 69.69±8.44                                                & 71.35                                             & 21.86                                            & 2.01                                              &  &  & 75.55±9.02                                              & 76.04±7.23                                                & 76.67                                             & 17.00                                            & 1.89                                              &  &  & 77.13±7.13                                              & 77.08±6.79                                                & 78.06                                             & 11.96                                            & 1.48                                              \\
                                                                          & HR32-IT                & C+L                   &  &  & 86.20±7.99                                              & 86.12±7.16                                                & 87.68                                             & 10.79                                            & 0.94                                              &  &  & 82.01±7.11                                              & 82.38±6.13                                                & 83.14                                             & 10.66                                            & 0.89                                              &  &  & 69.13±10.96                                             & 70.35±8.23                                                & 72.53                                             & 22.71                                            & 2.31                                              &  &  & 74.91±10.20                                             & 75.34±7.97                                                & 76.70                                             & 18.40                                            & 2.28                                              &  &  & 76.98±8.18                                              & 76.95±7.66                                                & 78.80                                             & 13.93                                            & 2.06                                              \\
                                                                          & HR18-IT                & SBD                   &  &  & 79.40±3.86                                              & 79.28±4.38                                                & 80.92                                             & 7.35                                             & 0.76                                              &  &  & 76.34±4.90                                              & 76.50±4.06                                                & 77.71                                             & 8.34                                             & 0.96                                              &  &  & 64.47±9.05                                              & 65.14±7.61                                                & 65.98                                             & 21.64                                            & 2.14                                              &  &  & 61.59±7.02                                              & 61.72±6.26                                                & 62.04                                             & 15.19                                            & 4.13                                              &  &  & 71.19±4.85                                              & 71.17±4.60                                                & 71.94                                             & 9.92                                             & 3.67                                              \\ \cline{1-3} \cline{6-10} \cline{13-17} \cline{20-24} \cline{27-31} \cline{34-38} 
\multirow{2}{*}{AdaptClick \citep{lin2024adaptiveclick}} & ViT-B                  & C+L                   &  &  & \underline{89.26±4.47}                                 & \textit{89.19±4.87}                                       & 87.07                                             & 6.05                                             & \textit{0.38}                                     &  &  & 83.45±5.56                                              & 84.10±4.16                                                & 84.05                                             & 8.87                                             & 0.67                                              &  &  & 74.92±7.09                                              & 76.18±5.94                                                & 76.94                                             & 13.45                                            & 1.27                                              &  &  & 77.17±6.49                                              & 77.53±5.26                                                & 78.21                                             & 12.87                                            & 1.55                                              &  &  & 82.16±6.15                                              & 82.18±6.07                                                & 81.97                                             & 10.82                                            & \textit{0.98}                                     \\
                                                                          & ViT-B                  & SBD                   &  &  & \textbf{90.31±2.38}                                     & \textbf{90.41±2.95}                                       & \textbf{91.11}                                    & \underline{3.61}                                & \underline{0.35}                                 &  &  & \textit{83.73±3.56}                                     & \textit{84.28±2.94}                                       & \textbf{85.02}                                    & \textit{5.41}                                    & 0.66                                              &  &  & \textbf{76.57±6.21}                                     & \textbf{77.91±4.91}                                       & \textbf{78.93}                                    & \underline{11.28}                               & \underline{1.08}                                 &  &  & 61.85±6.86                                              & 62.03±6.41                                                & 62.16                                             & 16.10                                            & 3.87                                              &  &  & 76.49±4.36                                              & 76.43±4.32                                                & 77.08                                             & 8.98                                             & 2.20                                              \\ \cline{1-3} \cline{6-10} \cline{13-17} \cline{20-24} \cline{27-31} \cline{34-38} 
\multirow{7}{*}{SimpleClick \citep{liu2022simpleclick}}  & ViT-B                  & C+L                   &  &  & \textit{89.20±3.78}                                     & 89.12±4.55                                                & 88.99                                             & 4.89                                             & 0.46                                              &  &  & \underline{84.13±5.73}                                 & \underline{85.02±4.22}                                   & 84.30                                             & 8.50                                             & 0.60                                              &  &  & 72.08±7.34                                              & 73.11±6.28                                                & 72.91                                             & 14.67                                            & 1.28                                              &  &  & 76.48±6.98                                              & 76.68±5.98                                                & 77.46                                             & 13.23                                            & 1.80                                              &  &  & 81.03±6.58                                              & 80.96±6.42                                                & 80.88                                             & 11.36                                            & 0.99                                              \\
                                                                          & ViT-L                  & C+L                   &  &  & 88.51±4.48                                              & 88.79±4.78                                                & 87.51                                             & 6.34                                             & 0.54                                              &  &  & 83.66±4.88                                              & 84.25±3.76                                                & 82.18                                             & 7.72                                             & 0.62                                              &  &  & 74.35±6.71                                              & 75.16±5.59                                                & 74.31                                             & 14.34                                            & 1.15                                              &  &  & \textit{80.34±6.16}                                     & \textit{80.62±5.17}                                       & \textit{80.62}                                    & \underline{11.74}                               & 1.38                                              &  &  & \textit{83.47±6.98}                                     & \textit{83.48±6.83}                                       & \textit{83.16}                                    & 11.16                                            & \underline{0.90}                                 \\
                                                                          & ViT-H                  & C+L                   &  &  & 87.91±4.72                                              & 88.49±4.25                                                & 86.29                                             & 7.30                                             & 0.61                                              &  &  & 83.16±4.71                                              & 83.75±3.52                                                & 83.15                                             & 7.49                                             & \textit{0.58}                                     &  &  & 72.44±7.56                                              & 73.22±6.19                                                & 73.43                                             & 16.76                                            & 1.36                                              &  &  & \underline{80.48±5.98}                                 & \underline{80.74±4.65}                                   & \underline{81.01}                                & \textbf{10.85}                                   & \textit{1.32}                                     &  &  & \underline{83.88±6.01}                                 & \underline{83.79±5.78}                                   & \underline{83.67}                                & 10.47                                            & \textbf{0.90}                                     \\
                                                                          & ViT-XT                 & SBD                   &  &  & 78.01±7.50                                              & 77.48±8.28                                                & 80.22                                             & 12.15                                            & 0.82                                              &  &  & 69.47±8.93                                              & 69.26±8.88                                                & 70.98                                             & 16.24                                            & 1.39                                              &  &  & 52.56±10.30                                             & 53.08±9.59                                                & 55.10                                             & 24.75                                            & 2.85                                              &  &  & 50.43±9.13                                              & 50.21±8.74                                                & 50.62                                             & 22.41                                            & 4.79                                              &  &  & 58.03±10.06                                             & 57.71±10.01                                               & 60.98                                             & 21.58                                            & 3.60                                              \\
                                                                          & ViT-B                  & SBD                   &  &  & 88.84±2.67                                              & 88.94±2.86                                                & 89.41                                             & \textit{3.74}                                    & \textbf{0.33}                                     &  &  & 83.10±4.24                                              & 83.67±3.87                                                & 84.41                                             & 6.50                                             & 0.60                                              &  &  & 74.22±6.11                                              & 75.56±4.87                                                & 76.02                                             & \textbf{11.20}                                   & \textit{1.08}                                     &  &  & 60.56±6.59                                              & 60.52±6.08                                                & 61.03                                             & 15.56                                            & 3.90                                              &  &  & 75.05±4.45                                              & 74.99±4.45                                                & 75.78                                             & \underline{8.69}                                & 2.08                                              \\
                                                                          & ViT-L                  & SBD                   &  &  & 88.48±3.56                                              & 88.49±3.75                                                & \textit{90.61}                                    & 6.12                                             & 0.69                                              &  &  & 83.41±3.29                                              & 84.01±2.98                                                & 84.23                                             & \underline{5.18}                                & \underline{0.58}                                 &  &  & \underline{76.25±6.21}                                 & \underline{77.67±4.45}                                   & \underline{77.96}                                & \textit{11.59}                                   & 1.11                                              &  &  & 64.97±6.71                                              & 64.96±5.96                                                & 65.10                                             & 15.57                                            & 2.40                                              &  &  & 77.76±4.14                                              & 77.65±4.18                                                & 78.18                                             & \textit{8.85}                                    & 1.43                                              \\
                                                                          & ViT-H                  & SBD                   &  &  & 88.54±2.09                                              & 88.32±2.47                                                & 89.38                                             & \textbf{3.01}                                    & 0.42                                              &  &  & 82.46±3.08                                              & 83.03±2.42                                                & 83.40                                             & \textbf{4.92}                                    & 0.67                                              &  &  & 74.96±7.10                                              & \textit{76.44±5.64}                                       & 75.56                                             & 13.14                                            & 1.31                                              &  &  & 65.52±6.39                                              & 65.71±6.01                                                & 65.69                                             & 14.60                                            & 2.68                                              &  &  & 77.72±3.11                                              & 77.72±3.16                                                & 78.22                                             & \textbf{6.69}                                    & 1.58                                              \\ \cline{1-3} \cline{6-10} \cline{13-17} \cline{20-24} \cline{27-31} \cline{34-38} 
CFR-ICL \citep{sun2023cfricl}                            & ViT-H                  & C+L                   &  &  & 88.95±4.72                                              & \underline{89.33±4.86}                                   & 87.61                                             & 7.29                                             & 0.59                                              &  &  & \textbf{85.05±5.88}                                     & \textbf{85.75±4.59}                                       & \textit{84.57}                                    & 9.37                                             & 0.64                                              &  &  & \textit{75.48±8.23}                                     & 76.43±6.71                                                & 76.84                                             & 17.09                                            & 1.31                                              &  &  & \textbf{81.18±6.55}                                     & \textbf{81.47±5.28}                                       & \textbf{81.64}                                    & \textit{12.34}                                   & 1.53                                              &  &  & \textbf{85.92±7.02}                                     & \textbf{85.86±6.68}                                       & \textbf{85.64}                                    & 11.47                                            & 0.98                                              \\ \cline{1-3} \cline{6-10} \cline{13-17} \cline{20-24} \cline{27-31} \cline{34-38} 
MobileSAM \citep{zhang2023faster}                        & ViT-Tiny               & SA-1B                 &  &  & 76.61±14.67                                             & 77.02±13.49                                               & 79.14                                             & 24.45                                            & 1.04                                              &  &  & 75.30±15.97                                             & 76.03±14.91                                               & 76.83                                             & 27.28                                            & 1.55                                              &  &  & 58.17±16.01                                             & 59.48±14.55                                               & 61.87                                             & 40.95                                            & 2.08                                              &  &  & 66.65±15.01                                             & 66.85±13.09                                               & 65.53                                             & 32.55                                            & 3.30                                              &  &  & 66.81±18.77                                             & 67.12±18.60                                               & 65.72                                             & 36.21                                            & 1.78                                              \\ \cline{1-3} \cline{6-10} \cline{13-17} \cline{20-24} \cline{27-31} \cline{34-38} 
\multirow{3}{*}{SAM \citep{kirillov2023segment}}         & ViT-B                  & SA-1B                 &  &  & 64.65±21.07                                             & 63.52±20.47                                               & 62.43                                             & 42.61                                            & 0.97                                              &  &  & 66.02±20.18                                             & 66.81±18.58                                               & 68.47                                             & 41.66                                            & 0.92                                              &  &  & 45.27±17.48                                             & 46.55±16.41                                               & 48.55                                             & 57.06                                            & 1.59                                              &  &  & 60.90±18.20                                             & 61.20±15.88                                               & 58.29                                             & 44.16                                            & 1.84                                              &  &  & 55.05±21.59                                             & 54.99±21.25                                               & 55.88                                             & 51.19                                            & 1.38                                              \\
                                                                          & ViT-L                  & SA-1B                 &  &  & 75.06±17.92                                             & 74.27±18.13                                               & 74.25                                             & 31.16                                            & 0.80                                              &  &  & 75.48±17.33                                             & 76.87±15.75                                               & 75.95                                             & 31.29                                            & 0.93                                              &  &  & 57.47±16.86                                             & 58.75±15.67                                               & 58.88                                             & 45.10                                            & 1.63                                              &  &  & 66.90±16.85                                             & 66.80±14.75                                               & 64.46                                             & 36.12                                            & 1.77                                              &  &  & 66.98±20.15                                             & 67.18±19.78                                               & 66.50                                             & 41.32                                            & 1.29                                              \\
                                                                          & ViT-H                  & SA-1B                 &  &  & 75.19±16.43                                             & 72.94±17.31                                               & 73.39                                             & 28.46                                            & 0.82                                              &  &  & 74.13±17.30                                             & 75.28±15.82                                               & 73.72                                             & 32.63                                            & 1.21                                              &  &  & 54.88±16.88                                             & 56.15±15.34                                               & 59.33                                             & 47.70                                            & 2.07                                              &  &  & 67.26±17.00                                             & 67.46±14.94                                               & 65.31                                             & 35.88                                            & 2.96                                              &  &  & 63.88±20.39                                             & 64.34±20.03                                               & 62.53                                             & 42.55                                            & 1.34                                              \\ \cline{1-3} \cline{6-10} \cline{13-17} \cline{20-24} \cline{27-31} \cline{34-38} 
\multirow{3}{*}{SAM-HQ \citep{ke2023segment}}            & ViT-B                  & SA-1B                 &  &  & 61.04±22.55                                             & 60.62±21.47                                               & 63.05                                             & 47.48                                            & 1.72                                              &  &  & 61.41±21.19                                             & 62.46±20.22                                               & 62.43                                             & 49.09                                            & 1.05                                              &  &  & 42.99±16.76                                             & 44.77±15.99                                               & 45.83                                             & 56.99                                            & 1.53                                              &  &  & 58.21±18.68                                             & 58.25±16.35                                               & 55.39                                             & 48.36                                            & 1.99                                              &  &  & 48.03±20.88                                             & 47.80±20.58                                               & 49.12                                             & 58.47                                            & 2.00                                              \\
                                                                          & ViT-L                  & SA-1B                 &  &  & 70.29±21.94                                             & 69.26±20.11                                               & 69.50                                             & 38.31                                            & 1.55                                              &  &  & 68.20±18.76                                             & 69.62±18.09                                               & 69.93                                             & 40.26                                            & 0.90                                              &  &  & 50.73±17.63                                             & 52.01±16.82                                               & 53.16                                             & 50.86                                            & 1.59                                              &  &  & 62.03±18.38                                             & 62.05±15.94                                               & 58.60                                             & 42.80                                            & 1.88                                              &  &  & 58.30±21.43                                             & 58.48±21.04                                               & 57.64                                             & 49.68                                            & 1.61                                              \\
                                                                          & ViT-H                  & SA-1B                 &  &  & 67.97±19.63                                             & 66.56±17.44                                               & 66.74                                             & 35.33                                            & 0.76                                              &  &  & 66.21±17.92                                             & 67.74±17.24                                               & 66.89                                             & 39.68                                            & 0.76                                              &  &  & 48.68±16.28                                             & 50.30±15.05                                               & 51.51                                             & 50.54                                            & 1.46                                              &  &  & 63.07±17.92                                             & 63.00±15.59                                               & 60.00                                             & 40.89                                            & 1.97                                              &  &  & 55.19±19.26                                             & 55.30±18.94                                               & 54.80                                             & 46.68                                            & 1.41                                              \\ \cline{1-3} \cline{6-10} \cline{13-17} \cline{20-24} \cline{27-31} \cline{34-38} 
\multirow{4}{*}{SAM 2 \citep{ravi2024sam}}             & Hiera-T                & SA-V                  &  &  & 76.80±19.32                                             & 77.24±17.90                                               & 81.70                                             & 30.87                                            & 0.72                                              &  &  & 76.13±15.15                                             & 77.66±14.17                                               & 77.97                                             & 26.23                                            & 0.71                                              &  &  & 66.32±16.48                                             & 67.67±14.70                                               & 70.47                                             & 33.39                                            & 1.11                                              &  &  & 66.48±15.87                                             & 66.80±13.43                                               & 63.63                                             & 34.98                                            & 1.51                                              &  &  & 45.90±18.61                                             & 45.88±18.32                                               & 47.74                                             & 55.51                                            & 1.63                                              \\
                                                                          & Hiera-S                & SA-V                  &  &  & 72.22±16.88                                             & 72.62±15.92                                               & 76.58                                             & 29.73                                            & 0.87                                              &  &  & 73.59±16.20                                             & 74.50±14.81                                               & 76.13                                             & 29.32                                            & 0.67                                              &  &  & 61.96±17.75                                             & 63.84±16.90                                               & 66.47                                             & 39.49                                            & 1.20                                              &  &  & 66.57±16.38                                             & 66.91±14.26                                               & 63.75                                             & 35.31                                            & 1.51                                              &  &  & 45.56±18.56                                             & 45.52±18.21                                               & 47.42                                             & 56.13                                            & 1.56                                              \\
                                                                          & Hiera-B+               & SA-V                  &  &  & 74.60±21.37                                             & 75.02±19.14                                               & 77.00                                             & 34.70                                            & 0.63                                              &  &  & 76.38±17.20                                             & 77.44±15.72                                               & 76.65                                             & 28.62                                            & 0.60                                              &  &  & 62.17±17.84                                             & 63.69±16.55                                               & 66.88                                             & 39.86                                            & 1.28                                              &  &  & 65.92±17.78                                             & 65.99±15.24                                               & 63.33                                             & 38.82                                            & 1.63                                              &  &  & 49.40±19.08                                             & 49.31±18.63                                               & 51.12                                             & 52.91                                            & 1.50                                              \\
                                                                          & Hiera-L                & SA-V                  &  &  & 83.08±16.58                                             & 83.78±14.15                                               & 85.26                                             & 24.71                                            & 0.57                                              &  &  & 80.77±15.25                                             & 81.45±13.05                                               & 82.05                                             & 23.24                                            & \textbf{0.57}                                     &  &  & 73.25±14.62                                             & 74.69±12.89                                               & \textit{77.22}                                    & 25.44                                            & \textbf{1.03}                                     &  &  & 74.42±13.61                                             & 74.83±11.33                                               & 72.39                                             & 25.93                                            & \underline{1.19}                                 &  &  & 50.56±19.00                                             & 50.29±18.68                                               & 51.72                                             & 51.62                                            & 1.37                                              \\ \cline{1-3} \cline{6-10} \cline{13-17} \cline{20-24} \cline{27-31} \cline{34-38} 
\multirow{4}{*}{SAM 2.1 \citep{ravi2024sam}}             & Hiera-T                & SA-V                  &  &  & 74.13±20.60                                             & 73.79±19.29                                               & 77.92                                             & 34.54                                            & 0.68                                              &  &  & 74.69±16.71                                             & 76.19±15.20                                               & 77.58                                             & 30.45                                            & 0.81                                              &  &  & 63.52±18.10                                             & 64.65±16.67                                               & 67.53                                             & 38.59                                            & 1.33                                              &  &  & 64.93±16.35                                             & 65.14±14.07                                               & 62.67                                             & 37.46                                            & 1.56                                              &  &  & 45.98±18.83                                             & 45.82±18.48                                               & 48.09                                             & 56.18                                            & 1.59                                              \\
                                                                          & Hiera-S                & SA-V                  &  &  & 72.85±19.23                                             & 72.91±17.96                                               & 80.02                                             & 32.05                                            & 0.71                                              &  &  & 75.00±16.24                                             & 76.03±15.15                                               & 79.15                                             & 28.94                                            & 0.70                                              &  &  & 66.15±17.13                                             & 67.78±15.84                                               & 71.36                                             & 35.37                                            & 1.35                                              &  &  & 66.58±16.49                                             & 66.87±14.13                                               & 63.46                                             & 36.02                                            & 1.47                                              &  &  & 45.80±18.70                                             & 45.71±18.37                                               & 47.47                                             & 56.39                                            & 1.68                                              \\
                                                                          & Hiera-B+               & SA-V                  &  &  & 72.19±22.57                                             & 72.17±21.18                                               & 73.42                                             & 38.18                                            & 0.77                                              &  &  & 74.46±18.27                                             & 75.30±16.85                                               & 76.42                                             & 31.82                                            & 0.72                                              &  &  & 57.69±19.10                                             & 59.03±17.75                                               & 63.79                                             & 46.23                                            & 1.30                                              &  &  & 65.27±17.93                                             & 65.16±15.43                                               & 62.60                                             & 40.14                                            & 1.54                                              &  &  & 48.43±18.59                                             & 48.32±18.18                                               & 50.52                                             & 52.70                                            & 1.52                                              \\
                                                                          & Hiera-L                & SA-V                  &  &  & 78.15±20.94                                             & 78.07±18.45                                               & 79.76                                             & 32.10                                            & 0.71                                              &  &  & 77.44±16.78                                             & 78.49±14.85                                               & 79.93                                             & 28.02                                            & 0.67                                              &  &  & 70.58±17.09                                             & 72.40±14.91                                               & 75.67                                             & 31.05                                            & 1.16                                              &  &  & 73.23±14.75                                             & 73.54±12.87                                               & 70.96                                             & 27.93                                            & \textbf{1.17}                                     &  &  & 48.50±18.76                                             & 48.17±18.41                                               & 49.98                                             & 53.51                                            & 1.38                                           \\
                                                                          \bottomrule
\end{tabular}
\end{sidewaystable}

%%%%%%%%%%%%%%%%%%%%%%%%%%%%%%%%%%%%%%%%%%%%%%%%%%%%%%%%%%%%

\FloatBarrier

\subsection{Finding hard instances for the first round using real-user clicks}

Using Figure~\ref{fig:mean_std} we depict a first-round real-world robustness of the methods for each instance in the datasets.
Obtained mean and standard deviation (STD) of IoU we estimated NSR for each sample.
Ideally, for robust interactive segmentation methods, all instances should have low NSR, e.g. high mean and low STD of IoU.
The instances with high NSR can be considered as hard cases for methods in the first round.

\begin{figure}[ht!]
    \centering
    \includegraphics[width=0.98\textwidth]{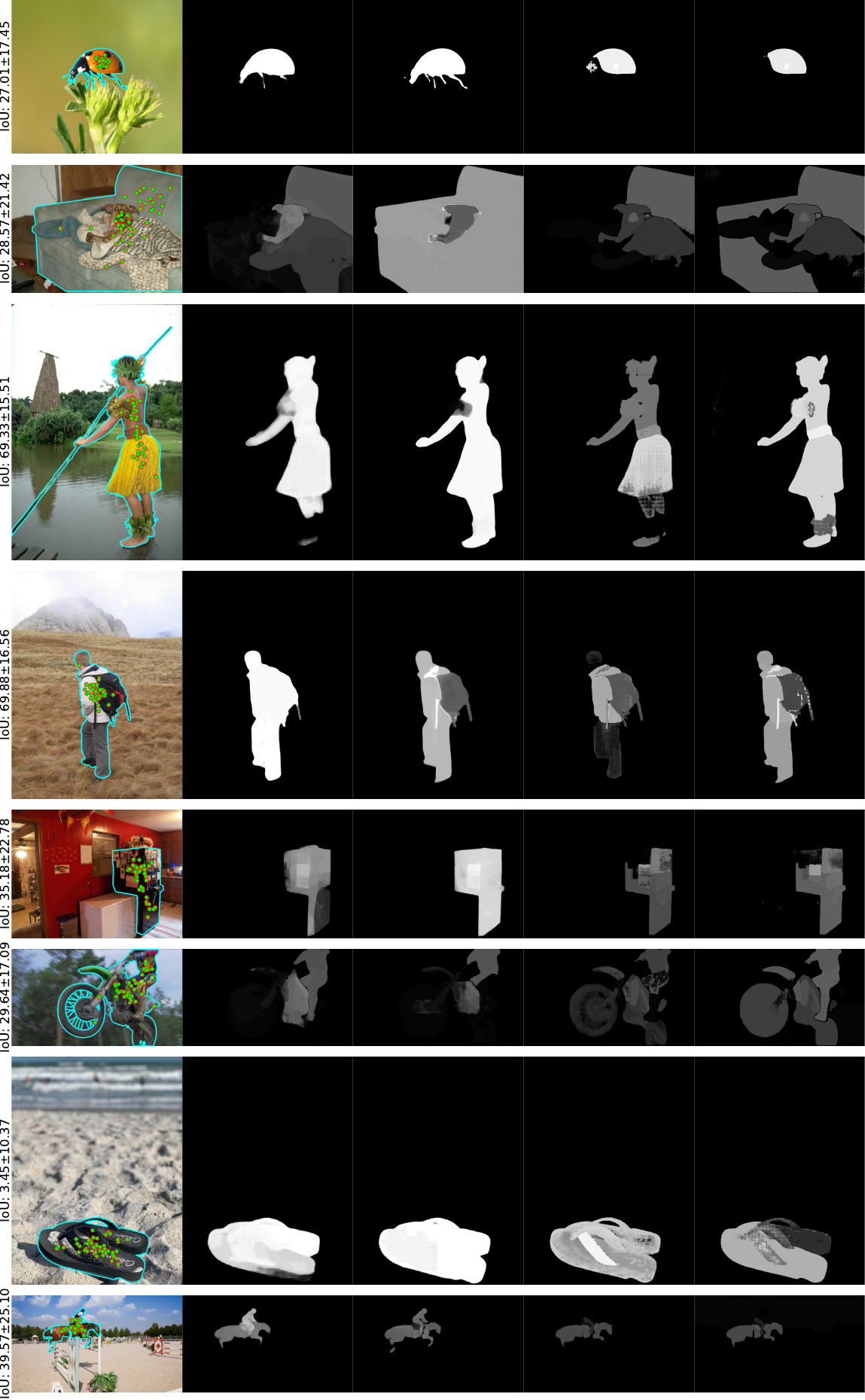}
    \caption{Samples with high User-IoU@1-NSR.
    From left to right -- the image with the target instance and real-user clicks of the first round; masks, averaged over the clicks, obtained by the methods: RITM~HRNet32-IT (C+L)~\cite{ritm}, SimpleClick ViT-H (C+L)~\cite{liu2022simpleclick}, SAM ViT-H (SA1-B)~\cite{kirillov2023segment}, SAM-HQ ViT-H (SA1-B)~\cite{ke2023segment}.
    }
    \label{fig:high_nsr}
\end{figure}

\begin{figure}[ht!]
    \centering
    \includegraphics[width=1.0\textwidth]{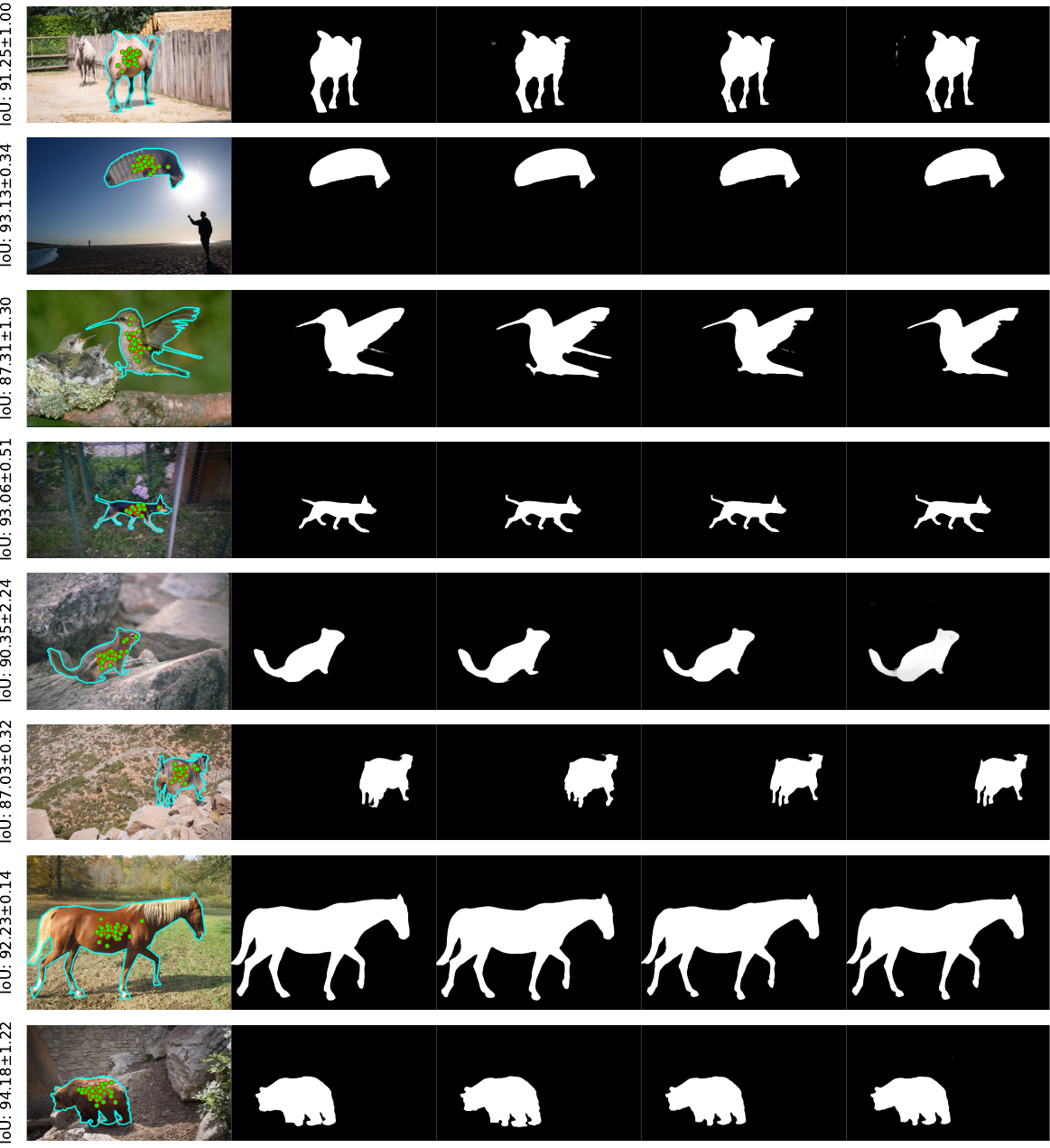}
    \caption{Samples with low User-IoU@1-NSR. 
    From left to right -- the image with the target instance and real-user clicks of the first round; masks, averaged over the clicks, obtained by the methods: RITM~HRNet32-IT (C+L)~\cite{ritm}, SimpleClick ViT-H (C+L)~\cite{liu2022simpleclick}, SAM ViT-H (SA1-B)~\cite{kirillov2023segment}, SAM-HQ ViT-H (SA1-B)~\cite{ke2023segment}.
    }
    \label{fig:low_nsr}
\end{figure}

Figures~\ref{fig:high_nsr} and~\ref{fig:low_nsr} illustrate the hard (high NSR) and simple (low NSR) cases for RITM~HRNet32-IT (C+L)~\cite{ritm}, SimpleClick ViT-H (C+L)~\cite{liu2022simpleclick}, SAM ViT-H (SA1-B)~\cite{kirillov2023segment}, SAM-HQ ViT-H (SA1-B)~\cite{ke2023segment}.
For each method, the uncertainty of prediction is depicted by the mean mask averaged over the method's predictions on the first-round clicks.
In the averaged mask, gray pixels correspond to variations of the method predictions (hard cases), black and white pixels illustrate low variation in, respectively, background and foreground predictions (simple cases).

\FloatBarrier

\newpage

\section{Datasheet for the Benchmark}

Datasheets for datasets~\cite{datasheet} facilitate communication between data creators and users in the form of the questionnaire explicating motivation, data acquisition process, and potential use cases. In this document, we provide a datasheet for the benchmark.

\subsection{Motivation}

\Q{1. For what purpose was the dataset created? Was there a specific task in mind? Was there a specific gap that needed to be filled? Please provide a description.}
\A{Interactive image segmentation aims at segmenting objects of interest given an image and sequential user input (clicks, strokes, contours), with each round allowing the user to correct prediction errors from the previous round. While numerous interactive segmentation methods have been developed, accurately evaluating these methods is crucial for identifying the best one. True evaluation requires real-user inputs. However, collecting many real-user inputs for multiple rounds is impractical, as such a dataset would need to be rebuilt for every method and interaction round due to its iterative nature.

To address this, researchers often use a simple strategy to simulate user inputs. This strategy involves generating a single click for each interaction round by selecting the largest error region from the previous round and clicking at the furthest point from the boundaries of this region (center point). However, previous works \citep{ritm, moskalenko2024tetris} have shown that assuming users click in the center of an object is overly simplistic and unrealistic.

To address the issue of unfair assessment algorithms, we started our work by collecting a large dataset of real-user clicks over multiple rounds of interactions. We trained a \textit{clickability model} to sample realistic user clicks, ensuring fairer evaluation. This model not only allows for more accurate multi-round evaluation but also provides data for first-round assessments.
}

\Q{2. Who created this dataset (e.g., which team, research group) and on behalf of which entity (e.g., company, institution, organization)?}
\A{Seven researchers at AIRI, Moscow (affiliated as of 2024) have created RClicks:
Anton Antonov, Andrey Moscalenko, Denis Shepelev, Alexander Krapukhin, Konstantin Soshin, Anton Konushin, Vlad Shakhuro.}

\Q{3. Who funded the creation of the dataset? If there is an associated grant, please provide the
name of the grantor and the grant name and number.}
\A{This research work was fully supported by the authors, including funding annotators' work.}

\Q{4. Any other comments?}
\A{No.}

\subsection{Composition}

\Q{5. What do the instances that comprise the dataset represent (e.g., documents, photos, people,
countries)? Are there multiple types of instances (e.g., movies, users, and ratings; people and
interactions between them; nodes and edges)? Please provide a description.}
\A{
RClicks dataset consists of \texttt{csv}-file with clicks and previous-round masks, obtained by some interaction methods after first round for each instance in a segmentation dataset.
Each row in the \texttt{csv}-file contains the following information in the columns:
\begin{itemize}
\item \texttt{dataset} -- a name of a segmentation dataset;
\item \texttt{image\_stem} -- a name of an image without suffix; 
\item \texttt{object\_stem} -- an encode of a target instance, if a mask contained multiple instances;
\item \texttt{model\_type} -- a name of interaction method, that was used in the previous round (in case of first round interaction -- empty);
\item \texttt{click\_type} -- a click type (\texttt{first} for the first round, or \texttt{fp} or \texttt{fn} for the subsequent rounds);
\item \texttt{full\_stem} -- a unique identifier of \texttt{image\_stem}, \texttt{object\_stem}, \texttt{model\_type} and \texttt{click\_type};
\item \texttt{device} -- a type of device where it was clicked (\texttt{pc} or \texttt{mobile});
\item \texttt{x}, \texttt{y} -- coordinates of a click;
\item \texttt{w}, \texttt{h} -- a width and a height of the image.
\end{itemize}
}

\Q{6. How many instances are there in total (of each type, if appropriate)?}
\A{There are 185\,349 clicks from the first round of iteration, 290\,195 clicks form the second (to collect these clicks we used 8144 masks from the first round). Overall — 475 544.
}

\Q{7. Does the dataset contain all possible instances or is it a sample (not necessarily random)
of instances from a larger set? If the dataset is a sample, then what is the larger set? Is
the sample representative of the larger set (e.g., geographic coverage)? If so, please describe
how this representativeness was validated/verified. If it is not representative of the larger set,
please describe why not (e.g., to cover a more diverse range of instances, because instances
were withheld or unavailable).}
\A{RClicks contains all click-annotated instances from GrabCut, Berkeley, DAVIS, COCO-MVal, TETRIS.}

\Q{8. What data does each instance consist of? “Raw” data (e.g., unprocessed text or images) or
features? In either case, please provide a description.}
\A{RClicks \texttt{csv}-file columns have a following ``raw'' types:
    \begin{itemize}
    \item strings: 
    \texttt{dataset},
    \texttt{image\_stem},
    \texttt{object\_stem},
    \texttt{model\_type},
    \texttt{click\_type},
    \texttt{full\_stem},
    \texttt{device};
    \item integers: 
    \texttt{x}, \texttt{y},
    \texttt{w}, \texttt{h}.
    \end{itemize}
Each previous-round mask in RClicks dataset -- is a single-channel image.
}

\Q{9. Is there a label or target associated with each instance? If so, please provide a description.}
\A{Yes, each instance has a corresponding mask of instance on the image, where user had to click.}

\Q{10. Is any information missing from individual instances? If so, please provide a description,
explaining why this information is missing (e.g., because it was unavailable). This does not
include intentionally removed information, but might include, e.g., redacted text.}
\A{No.}

\Q{11. Are relationships between individual instances made explicit (e.g., users’ movie ratings,
social network links)? If so, please describe how these relationships are made explicit.}
\A{No, there are neither explicit of implicit relationships between individual instances in RClicks.}

\Q{12. Are there recommended data splits (e.g., training, development/validation, testing)? If so,
please provide a description of these splits, explaining the rationale behind them.}
\A{We intend our dataset to be primarily used for benchmarking interactive segmentation methods. Hence, all instances in our dataset would be used for testing. For clickability model training there is a train/test split.}

\Q{13. Are there any errors, sources of noise, or redundancies in the dataset? If so, please provide
a description.}
\A{It can be stated with certainty that there are no erroneous values in RClicks. The sole source of noise is the fact that during the annotation process, participants were permitted to make minor "mistakes." A small subset of clicks was collected that were situated outside the chosen object mask but in close proximity to its boundaries.}

\Q{14. Is the dataset self-contained, or does it link to or otherwise rely on external resources (e.g.,
websites, tweets, other datasets)? If it links to or relies on external resources,
(a) Are there guarantees that they will exist, and remain constant, over time?\\
(b) Are there official archival versions of the complete dataset (i.e., including the external
resources as they existed at the time the dataset was created)?\\
(c) Are there any restrictions (e.g., licenses, fees) associated with any of the external resources
that might apply to a future user? Please provide descriptions of all external resources and
any restrictions associated with them, as well as links or other access points, as appropriate.}
\A{The dataset is self-contained.}

\Q{15. Does the dataset contain data that might be considered confidential (e.g., data that is
protected by legal privilege or by doctor-patient confidentiality, data that includes the
content of individuals non-public communications)? If so, please provide a description.}
\A{No, the clicks in RClicks do not cover scenarios that may be considered confidential.}

\Q{16. Does the dataset contain data that, if viewed directly, might be offensive, insulting,
threatening, or might otherwise cause anxiety? If so, please describe why.}
\A{The dataset contains no data that
might be offensive, insulting, threatening, or might cause anxiety by manually curating the set of images.}

\Q{17. Does the dataset relate to people? If not, you may skip remaining questions in this section.}
\A{No.}

\Q{21. Any other comments?}
\A{No.}

\subsection{Collection Process}

\Q{22. How was the data associated with each instance acquired? Was the data directly observable
(e.g., raw text, movie ratings), reported by subjects (e.g., survey responses), or indirectly
inferred/derived from other data (e.g., part-of-speech tags, model-based guesses for age or
language)? If data was reported by subjects or indirectly inferred/derived from other data, was
the data validated/verified? If so, please describe how.}
\A{We downloaded images and mask annotation from GrabCut, Berkeley, DAVIS, COCO-MVal, TETRIS and annotated instances from the datasets with clicks on crowd sourcing platform \href{toloka.ai}{toloka.ai}. Accordingly, the “raw” image data was directly observable by annotators, and annotations were created manually.}

\Q{23. What mechanisms or procedures were used to collect the data (e.g., hardware apparatus
or sensor, manual human curation, software program, software API)? How were these
mechanisms or procedures validated?}
\A{We selected and downloaded interactive segmentation datasets. Then with \href{toloka.ai}{toloka.ai} crowd sourcing platform annotated datasets' instances with clicks. The annotators saw image, then mask of chosen object, then they clicked at this object on image.} 

\Q{24. If the dataset is a sample from a larger set, what was the sampling strategy?}
\A{We release whole dataset.}

\Q{25. Who was involved in data collection process (e.g., students, crowd-workers, contractors)
and how were they compensated (e.g., how much were crowd-workers paid)?}
\A{We used \href{toloka.ai}{toloka.ai} crowd sourcing platform, every crowd-worker was paid 0.02\$ for 10 clicks.}

\Q{26. Over what timeframe was the data collected? Does this timeframe match the creation
timeframe of the data associated with the instances (e.g., recent crawl of old news articles)?
If not, please provide a description of the timeframe.}
\A{Data was collected from the April of 2024 to the May of 2024.}

\Q{27. Were any ethical review processes conducted (e.g., by an institutional review board)? If
so, please provide a description of these review processes, including the outcomes, as well as a
link or other access point to any supporting documentation.}
\A{No, such processes were unnecessary in our case.}

\Q{28. Does the dataset relate to people? If not, you may skip remaining questions in this section.}
\A{No.}

\Q{34. Any other comments?}
\A{No.}

\subsection{Preprocessing, Cleaning, and/or Labeling}

\Q{35. Was any preprocessing/cleaning/labeling of the data done (e.g., discretization or bucket-
ing, tokenization, part-of-speech tagging, SIFT feature extraction, removal of instances,
processing of missing values)? If so, please provide a description. If not, you may skip the
remainder of the questions in this section.}
\A{We clean data from the clicks that were done not in the object of interest.}

\Q{36. Was the “raw” data saved in addition to the preprocessed/cleaned/labeled data (e.g., to
support unanticipated future uses)? If so, please provide a link or other access point to the
“raw” data.}
\A{No.}

\Q{37. Is the software used to preprocess/clean/label the instances available? If so, please provide
a link or other access point.}
\A{We provide code scripts in supplemental materials.}

\Q{38. Any other comments?}
\A{No.}

\subsection{Uses}

\Q{39. Has the dataset been used for any tasks already? If so, please provide a description.}
\A{We have used our dataset to evaluate state-of-the-art interactive segmentation methods and train our \textit{clickability model}.}

\Q{40. Is there a repository that links to any or all papers or systems that use the dataset? If so,
please provide a link or other access point.}
\A{We do not maintain such a repository. However, citation trackers like Google Scholar and
Semantic Scholar would list all future works that cite our dataset.}

\Q{41. What (other) tasks could the dataset be used for?}
\A{We anticipate that the dataset could be used for benchmarking interactive segmentation methods and training interactive segmentation methods.}

\Q{42. Is there anything about the composition of the dataset or the way it was collected and
preprocessed/cleaned/labeled that might impact future uses? For example, is there anything
that a future user might need to know to avoid uses that could result in unfair treatment of
individuals or groups (e.g., stereotyping, quality of service issues) or other undesirable harms
(e.g., financial harms, legal risks) If so, please provide a description. Is there anything a future
user could do to mitigate these undesirable harms?}
\A{This is very difficult to anticipate. Future users should be aware that our dataset was collected both from PC and Mobile devices, choose clicks needed to their platform.}

\Q{43. Are there any tasks for which the dataset should not be used? If so, please provide a
description.}
\A{No.}

\Q{44. Any other comments?}
\A{No.}

\subsection{Distribution}

\Q{45. Will the dataset be distributed to third parties outside of the entity (e.g., company,
institution, organization) on behalf of which the dataset was created? If so, please provide a
description.}
\A{Yes, our dataset will be publicly available.}

\Q{46. How will the dataset will be distributed (e.g., tarball on website, API, GitHub) Does the
dataset have a digital object identifier (DOI)?}
\A{We will distribute our dataset at github repository. All uses of RClicks should cite this paper, explicating which version of the dataset was considered.}

\Q{47. When will the dataset be distributed?}
\A{The dataset will be publicly available starting from September 2024.}

\Q{48. Will the dataset be distributed under a copyright or other intellectual property (IP)
license, and/or under applicable terms of use (ToU)? If so, please describe this license and/or
ToU, and provide a link or other access point to, or otherwise reproduce, any relevant licensing
terms or ToU, as well as any fees associated with these restrictions.}
\A{Clicks are under \href{https://creativecommons.org/licenses/by-nc/4.0/deed.en}{CC BY-NC 4.0}, evaluation code and baseline models are under \href{https://opensource.org/licenses/MIT}{MIT}.}

\Q{49. Have any third parties imposed IP-based or other restrictions on the data associated with
the instances? If so, please describe these restrictions, and provide a link or other access point
to, or otherwise reproduce, any relevant licensing terms, as well as any fees associated with
these restrictions.}
\A{Nowadays, GrabCut and Berkeley datasets are unavailable from official sites, they can be downloaded through WebArchiveMachine or through \href{https://github.com/SamsungLabs/ritm_interactive_segmentation}{RITM repository}.}

\Q{50. Do any export controls or other regulatory restrictions apply to the dataset or to individual
instances? If so, please describe these restrictions, and provide a link or other access point to,
or otherwise reproduce, any supporting documentation.}
\A{No.}

\Q{51. Any other comments?}
\A{No.}

\subsection{Maintenance}

\Q{52. Who will be supporting/hosting/maintaining the dataset?}
\A{Our team will maintain the dataset.}

\Q{53. How can the owner/curator/manager of the dataset be contacted (e.g., email address)?}
\A{antonov@airi.net}

\Q{54. Is there an erratum? If so, please provide a link or other access point.} 
\A{There is no erratum for our initial release.}

\Q{55. Will the dataset be updated (e.g., to correct labeling errors, add new instances, delete
instances)? If so, please describe how often, by whom, and how updates will be communicated
to users (e.g., mailing list, GitHub)?}
\A{We will probably update our dataset on a non-regular basis.}

\Q{56. If the dataset relates to people, are there applicable limits on the retention of the data
associated with the instances (e.g., were individuals in question told that their data would
be retained for a fixed period of time and then deleted)? If so, please describe these limits
and explain how they will be enforced.}
\A{No.}

\Q{57. Will older versions of the dataset continue to be supported/hosted/maintained? If so,
please describe how. If not, please describe how its obsolescence will be communicated to users.}
\A{A new version release of RClicks will automatically deprecate its previous version. We
will only support and maintain the latest version at all times.}

\Q{58. If others want to extend/augment/build on/contribute to the dataset, is there a mechanism
for them to do so? If so, please provide a description. Will these contributions be verified? If so,
please describe how. If not, why not? Is there a process for communicating/distributing these
contributions to other users? If so, please provide a description.}
\A{Anyone can extend RClicks if providing high-resolution images with properly annotated masks and clicks for it. We are open to accept extensions via personal communication with potential contributors. Otherwise, our code and data licenses allow others to create independent derivative works (with proper attribution).}

% \clearpage

% \newpage
% %\hypersetup{linkcolor=blue}
% \bibliographystyle{unsrtnat}
% \bibliography{supplementary}

% \newpage

% \input{checklist}

\end{document}